\documentclass[runningheads]{llncs}

\usepackage{eccv}

\usepackage{eccvabbrv}

\usepackage{graphicx}
\usepackage{booktabs}
\usepackage{xcolor,colortbl}

\usepackage[accsupp]{axessibility}  

\usepackage{hyperref}

\usepackage{orcidlink}
\usepackage{siunitx}
\usepackage[rightcaption]{sidecap}
\usepackage{multirow}

\begin{document}

\title{Rasterized Edge Gradients: Handling Discontinuities Differentiably}


\author{Stanislav Pidhorskyi\orcidlink{0000-0003-4019-8616} \and
Tomas Simon\orcidlink{0000-0002-0972-7455} \and
Gabriel Schwartz\orcidlink{0000-0002-8781-5573} \and
He Wen \and
Yaser Sheikh \and
Jason Saragih  }

\authorrunning{S.~Pidhorskyi et al.}

\institute{Reality Labs, Meta, Pittsburgh, Pennsylvania, USA}

\maketitle

\definecolor{tabfirst}{rgb}{1, 0.7, 0.7} 
\definecolor{tabsecond}{rgb}{1, 0.85, 0.7} 
\definecolor{tabthird}{rgb}{1, 1, 0.7} 

\begin{abstract}
Computing the gradients of a rendering process is paramount for diverse applications in computer vision and graphics.
However, accurate computation of these gradients is challenging due to discontinuities and rendering approximations, particularly for surface-based representations and rasterization-based rendering.
We present a novel method for computing gradients at visibility discontinuities for rasterization-based differentiable renderers.
Our method elegantly simplifies the traditionally complex problem through a carefully designed approximation strategy, allowing for a straightforward, effective, and performant solution.
We introduce a novel concept of \textit{micro-edges}, which allows us to treat the rasterized images as outcomes of a differentiable, continuous process aligned with the inherently non-differentiable, discrete-pixel rasterization.
This technique eliminates the necessity for rendering approximations or other modifications to the forward pass, preserving the integrity of the rendered image, which makes it applicable to rasterized masks, depth, and normals images where filtering is prohibitive.
Utilizing \textit{micro-edges} simplifies gradient interpretation at discontinuities and enables handling of geometry intersections, offering an advantage over the prior art.
We showcase our method in dynamic human head scene reconstruction, demonstrating effective handling of camera images and segmentation masks.
%
\end{abstract}

\section{Introduction}
\label{sec:intro}

Significant advances have been made in recent years in modeling real 3D objects from image measurements.
Much of this advancement can be credited to improvements in inverse rendering, which enables the automatic inference of 3D scene parameters that best reconstruct the images.
While volumetric methods like NeRF~\cite{mildenhall2020nerf} simplify inverse rendering by eliminating the need for predefined scene topology, classical mesh-based representations remain widely used due to their efficiency in modeling opaque surfaces.
These representations often rely on highly performant rasterization, which requires discrete operations to reason about ordering and coverage, which poses challenges for gradient computations.
Our main contribution is a theoretical framework for approximating gradients of a rasterization process that simultaneously elucidates previous constructions, arrives at a simple, fast, and accurate formulation, and further improves gradient approximation accuracy over state-of-the-art methods.

In rasterization, a rendered pixel's value is determined by the foremost triangle covering it.
Pixels whose footprint lies entirely within a single triangle are straightforward to handle, as infinitesimal mesh motion does not change their triangle membership.
However, pixels on triangle boundaries, either between adjacent triangles, at occlusion boundaries, or at intersections between triangles, must account for how their triangle membership changes in response to infinitesimal changes in the mesh.
Existing methods employ different approximations of soft-membership to account for this, enabling gradient computation in those areas, effectively compositing contributions from member triangles via a weighted sum.
For example, ~\cite{laine2020modular} uses the anti-aliasing approach, ~\cite{jakob2022mitsuba3} averages monte-carlo samples during ray-casting, and~\cite{rhodin2015versatile} extends the influence of boundary triangles using a falloff function.
In this work, we show that all these methods exhibit approximation errors that can fail to compute correct gradients in one case or another.
In particular, none of the existing methods accurately compute the gradient at triangle intersections, which can often happen during inverse rendering, especially when the current estimate is far from the desired solution.

In contrast to existing works that directly tackle geometric discontinuities and rasterization's discrete nature, we introduce a \textit{micro-edge} formulation
which allows to interpret the rasterized image as an outcome of a continuous process that coincidentally aligns with discrete-pixel rasterization, simplifying gradient computation significantly.
Unlike some other works~\cite{laine2020modular,rhodin2015versatile}, our formulation achieves good gradient estimates without altering the rasterization forward pass, maintaining the rasterized image's integrity.
This is crucial for optimizing segmentation masks, depth maps, and normal maps where filtering or smoothing is not feasible, e.g.\ anisotropic filtering or soft rasterization would mix normals from different surfaces misrepresenting the geometry.
The simplifications offered by our \textit{micro-edge} formulation allow us to seamlessly handle discontinuities caused by geometry intersections, offering an advantage over the prior art.

In summary, our paper introduces a straightforward, accurate, and efficient method for computing gradients in rasterization, comparable to existing techniques but with greater simplicity and the ability to handle self-intersections.
We analyze our method's errors from a theoretical standpoint and compare gradients qualitatively and quantitatively with finite differences and other methods across various test cases.
We also assess runtime efficiency and accuracy by image size and showcase qualitative and quantitative findings on a synthetic blender dataset~\cite{mildenhall2021nerf}.
Our method excels in complex applications, such as detailed dynamic human head reconstructions, effectively managing the intricate details of the inner mouth region with significant occlusions and deformations.

\section{Related work}
\label{sec:related_work}

Differentiable rendering is one of the critical building blocks for many existing computer vision problems, such as 3d object reconstruction~\cite{kato2018neural,yan2016perspective,tulsiani2017multi,yang2018learning,choy20163d}, 3d object prediction~\cite{chen2019learning,beker2020monocular,chen2021dib,henzler2021unsupervised}, pose estimation~\cite{pavlakos2018learning,bogo2016keep,ge20193d,baek2019pushing} novel view synthesis~\cite{mildenhall2021nerf,barron2021mip,zhang2020nerf++,muller2022instant,kerbl20233d}, as well as newly emerging applications such as text-to-3D generative models~\cite{poole2022dreamfusion,lin2023magic3d,tsalicoglou2023textmesh,shi2023mvdream}.
Differentiable rendering employs a wide array of underlying 3D representations including explicit surfaces like mesh-based representations~\cite{loper2014opendr,deLaGorce:2011:MHP:2006854.2007005,laine2020modular}, implicit surfaces~\cite{park2019deepsdf,jiang2020sdfdiff,vicini2022differentiable,bangaru2022differentiable}, point clouds~\cite{roveri2018pointpronets,wiles2020synsin,kerbl20233d}, explicit or implicit volume representations~\cite{yan2016perspective,lombardi2019neural,liu2020neural,yu2021plenoxels,mildenhall2021nerf,niemeyer2020differentiable,barron2021mip}, and hybrid representations~\cite{chan2022efficient,muller2022instant}.
For a comprehensive introduction to differentiable rendering, we refer the reader to Zhao {\em et al.}'s excellent SIGGRAPH course notes~\cite{Zhao2020DifferentiableCourse}.

Mesh optimization often poses more challenges than implicit and volumetric methods, leading to the latter's prevalence in the field, as noted by Roessle {\em et al.}~\cite{roessle2022dense}.
Despite this, meshes excel in high-performance rendering and imply registration with predefined topologies.
Mesh rendering methods can be categorized into \textit{ray tracing-based} and \textit{rasterization-based}, with the former including frameworks like Mitsuba 3\cite{NimierDavidVicini2019Mitsuba2,Jakob2020DrJit}.
Differentiable ray tracing, aimed at direct illumination or path tracing, is computationally expensive but principled.
Key advancements by Li {\em et al.}~\cite{Li:2018:DMC} and Zhang {\em et al.}~\cite{Zhang2019DTRT} split this gradient into discontinuity integrals with Dirac delta functions and differentiable continuous parts.
Li {\em et al.} replaced Dirac delta integration with boundary line integration, while Zhang {\em et al.} applied Reynold's transport theorem for a similar decomposition.
Both methods are intricate, necessitating silhouette edge sampling --- a significant bottleneck in their application.
Loubet {\em et al.}~\cite{Loubet2019Reparameterizing} proposed a variable change to simplify discontinuity integration, further refined by Bangaru {\em et al.}~\cite{bangaru2020unbiased} using warped area fields to correct biases, yet practical application challenges persist.
%
Despite their flexibility, the computational expense of such ray tracing frameworks restrict their practicality in optimization-intensive tasks~\cite{laine2020modular}.

For \textit{rasterization-based} methods, rendering meshes differentiably is hindered by non-differentiable visibility at boundaries.
Unlike ray tracing, which uses boundary integrals for visibility gradients, rasterization's fixed-grid and z-buffer approach lacks this feature.
Solutions typically involve approximating derivatives or approximating rendering to achieve inherent differentiability.
De La Gorce {\em et al.}~\cite{deLaGorce:2011:MHP:2006854.2007005} and Loper {\em et al.}~\cite{loper2014opendr} pioneered derivative approximation in rasterization.
De La Gorce {\em et al.} distinguished between continuous regions and occlusion boundaries, introducing 'occlusion forces' for the latter.
Loper {\em et al.} simplified this by detecting discontinuities post-rasterization and employing differential filters, like the Sobel filter, for boundary approximation.

The 'Nvdiffrast' approach by Laine {\em et al.}~\cite{laine2020modular} tackles point-sampled visibility issues with innovative differentiable analytic antialiasing, which transforms sharp discontinuities into smooth transitions, enabling gradient computation.
Analytical antialiasing is achieved by estimating coverage using silhouette edges.
A limitation of the method is that if triangles containing silhouette edges do not overlap with any pixel center, they will be overlooked during the antialiasing process.
Moreover, analytical antialiasing is inherently approximate, so the obtained gradients are also approximate.
Regrettably, the approach is still quite complex, modifies the rasterized image, and requires a specialized data structure for connectivity and detailed pixel coverage computations.
On the other hand, methods like Rhodin {\em et al.}~\cite{rhodin2015versatile} and Liu {\em et al.}'s 'Soft Rasterizer'~\cite{liu2019softras} opt for modifying the rendering model.
Rhodin {\em et al.} introduce fuzzy edges to soften discontinuities, while 'Soft Rasterizer' further blurs boundaries and averages depth contributions, facilitating gradient propagation across occluded primitives.

In this work, we deliberately bypass physically based rendering, global illumination, and lighting/material models, and instead focus exclusively on rasterization\nobreakdash-based approaches for their speed.
We draw inspiration from the core idea of the edge sampling method~\cite{Li:2018:DMC} and analytical antialiasing~\cite{laine2020modular} and distill this into a much simpler yet effective method that also provides accurate gradients.
%
%
Notably, our method's simplicity allows direct handling of self-penetrating geometry, an aspect not addressed in previous research.
Other methods to handle self-penetrating geometry would require complex and computationally intensive geometry preprocessing at each optimization step, involving intersection detection and differentiable splitting of the faces.
In contrast, our approach explicitly manages interpenetrating geometry with virtually no additional overhead.

%

%
%
%

\section{Preliminaries}
\label{sec:preliminaries}

Following Li {\em et al.}~\cite{Li:2018:DMC}, given a 2D pixel filter $k$ and radiance $R$, a pixel's color can be written:
\begin{equation}
  I= \iint_{D}^{} k(x,y)R(x,y)\,dx\,dy.
  \label{eq-pixel_color}
\end{equation}
For notational convenience, as in~\cite{Li:2018:DMC}, we denote $f(x,y) = k(x,y)R(x,y)$.
For simplicity, let us first consider the case of a single pixel.
We are interested in the gradient of $\frac{\partial L (I)}{\partial \Phi}$, where the scalar function $L(I)$ of the rendered pixel $I$ defines our loss, and $\Phi$ are the scene parameters that we aim to optimize.
According to the chain rule:
\begin{equation}
  \frac{\partial L (I)}{\partial \Phi} = \frac{\partial I}{\partial \Phi} \frac{\partial L}{\partial I},
  \label{eq-chain}
\end{equation}
where $\frac{\partial L}{\partial I}$ is the incoming gradient from the loss function and is typically computed by automatic differentiation (AD).
So, our goal becomes to find how the pixel value changes with respect to the scene parameters: 
\begin{equation}
  \frac{\partial I}{\partial \Phi} = \frac{\partial}{\partial \Phi} \iint_{D}^{} f(x,y) \,dx\,dy.
  \label{eq-der1}
\end{equation}
While the function $f$ may not in fact be differentiable, its integral remains continuous and thus differentiable.
Let us assume that $f$ is partitioned into two continuous half-spaces, each represented by two continuous and differentiable functions $f_a$ and $f_b$, respectively, as follows:
\begin{equation}
 f(x,y) = \theta(\alpha(x,y)) f_a(x,y) + \theta(-\alpha(x,y)) f_b(x,y),
  \label{eq-half}
\end{equation}
where $\alpha$ specifies the dividing edge, and $\theta$ is the Heaviside step function, which selects between $f_a$ and $f_b$.
The differentiation of the integral can be broken into two parts by applying the product rule.
Let us start with the integrand $f_a$:
\begin{equation}
  \begin{aligned}
   & \frac{\partial}{\partial \Phi} \iint_{D}^{} \theta(\alpha(x,y)) f_a(x,y) \,dx\,dy \\
  =& \iint_{D}^{} \delta (\alpha(x,y))  \frac{\partial \alpha(x,y)}{\partial \Phi} f_a(x,y)  \,dx\,dy
  + \iint_{D}^{}  \frac{\partial f_a(x,y)}{\partial \Phi} \theta(\alpha(x,y)) \,dx\,dy,
  \end{aligned}
  \label{eq-product_rule}
\end{equation}
and a similar derivation can be done for $f_b$.
The second term on the right-hand side of Eqn.~(\ref{eq-product_rule}) is the integral of the derivative of the smooth function, $f_a(x,y)$.
This derivative is readily computable with AD, and in the context of differentiable rasterizers it typically involves differentiating through the interpolation step, the barycentric coordinates, and the texture sampler.
Due to its straightforward computation, we will represent it as $\Omega$ and omit it from further discussion, focusing instead on the first term on the right-hand side of Eqn.~(\ref{eq-product_rule}).
This term involves a Dirac delta function, and can be interpreted as taking the integral over the line $\alpha(x,y)=0$:
%
\begin{equation}
  \begin{aligned}
      & \iint_{D}^{} \delta (\alpha(x,y))  \frac{\partial \alpha(x,y)}{\partial \Phi} f_a(x,y)  \,dx\,dy\\
      =& \int_{\alpha(x,y)=0}^{}  \frac{\partial \alpha(x,y)}{\partial \Phi}
      \|  \nabla_{x,y}\alpha(x,y)  \| ^{-1}
      f_a(x,y)  \,dt,
  \end{aligned}
  \label{eq-integrating_dirac}
\end{equation}
where $\|  \nabla_{x,y}\alpha(x,y)  \|$  is $L^2$ length of the edge equation gradient which accounts for the Jacobian of the variable substitution, and $t$ is the parametrization of the line along which we integrate.

\section{Method}
\label{sec:method}

In this section we describe our method that we call \emph{EdgeGrad}.
First we will focus on the core assumptions and approximations that form the basis of our method.
Then we will see that due to our specifically chosen approximation, we can apply the prior method of edge sampling to rasterization.
From there, we will focus on edge detection and classification heuristics, and our discretization scheme. We will conclude by describing an extension of our method for handling geometry intersections.

\subsection{Rasterizing Edges}
\label{subsec:approxamation}

Our primary objective is to differentiate rasterization while preserving its core image formation principles.
We aim to calculate the gradients of pixel values relative to the mesh vertex positions, including gradients caused by visibility changes such as occlusion and disocclusion by primitive geometry.
We seek to do this with 
a simple, minimal, and efficient method that does not necessitate modifying the initially rasterized image.

\begin{figure}[!tb]
  \begin{minipage}{0.49\textwidth}
    \centering
    \includegraphics[width=0.9\columnwidth]{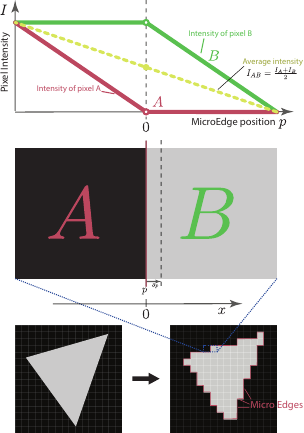}
    \caption{ \textbf{Micro-edges and Pixel Pair}.
    The figure illustrates construction of the micro-edges and the change in pixel intensities A and B with the edge position $p$ movement, highlighting the C1 discontinuity at $p=0$, absent in average intensity.}
    \label{fig:pixel_pair}
  \end{minipage}\hfill
  \begin {minipage}{0.49\textwidth}
    \centering
    \includegraphics[width=0.96\columnwidth]{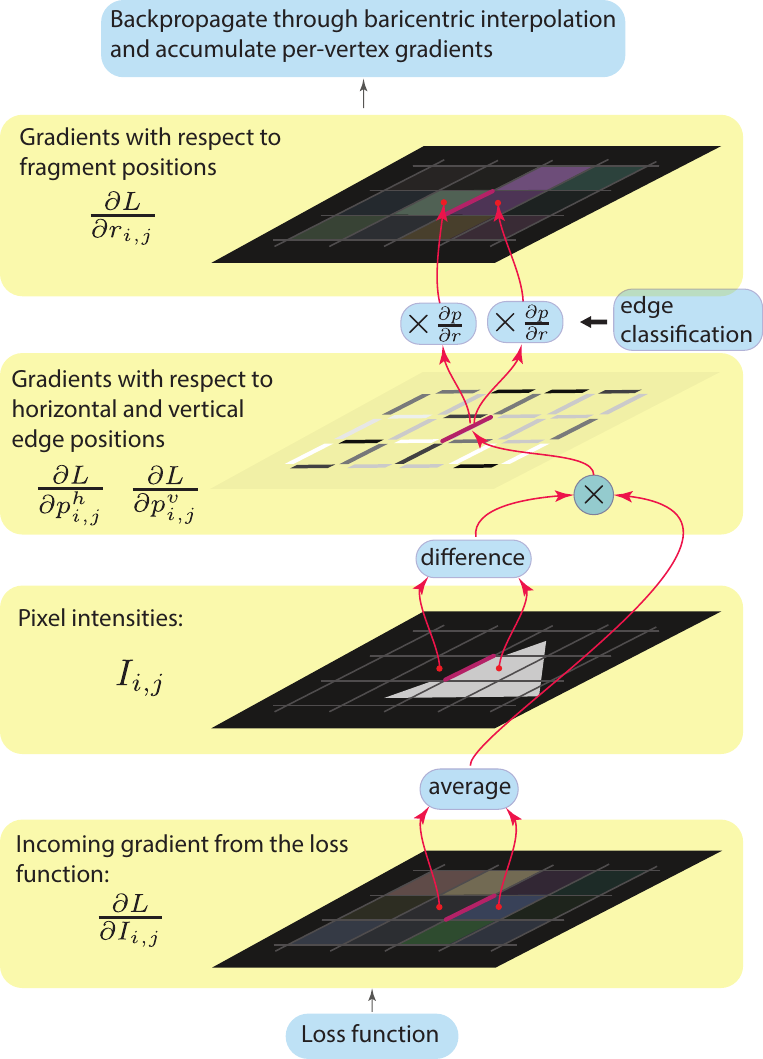}
    \caption{\textbf{Gradient computation flow}. The figure illustrates the backward propagation process, where gradients from the loss function are traced back and scattered to the fragments.
    }
    \label{fig:gradient_flow}
  \end{minipage}
\end{figure}

Our main insight is that instead of approximating rasterization using a continuous representation, we assume the rasterized image is the result of some other continuous process that coincidentally aligns with our inherently non-differentiable, discrete-pixel rasterization.\footnote{
This concept is akin to representing real numbers with floating-point arithmetic.
Although there are only 4,294,967,296 single-precision floating-point numbers
(a quarter of which lie in the interval [0,1])
, they are still employed for continuous computation.
We treat these computations as smooth and differentiable, even though they are, in essence, discontinuous.
}.
In essence, we opt for a direct approximation to compute gradients, instead of an indirect approach that first necessitates that the rendering be made continuous, for instance, through antialiasing filtering.
More specifically, we 
use a construction where
all straight edges, whether they are sides of triangles or intersections between triangles, are comprised of a collection of micro-edges.
These micro-edges are assumed to be either strictly vertical or horizontal and are always positioned precisely between two pixels.
An illustration of these constructions is shown in Figure~\ref{fig:pixel_pair}.

Our approximation leads to several key properties:
\textbf{a)} boundaries never intersect pixels, only lying exactly between them;
\textbf{b)} pixels are always either fully covered or not covered at all;
\textbf{c)} there is at most one boundary between any two neighbouring pixels;
\textbf{d)} we do not need to access the source geometry to evaluate where the edge is exactly.
With this, we can reinterpret the problem as if the rasterized image was in fact obtained by applying an antialiasing filter, albeit without visible effects; the resulting image remains identical to the pre-antialiasing source.
Essentially our construct of the micro-edges renders the antialiasing filter as an identity operation.

With this approximation, gradient estimation becomes a local operation, requiring the analysis one edge between two pixels at a time.
Note that despite the approximation that all edges reside strictly between pixels, we do not assume that they are fixed, but rather allow them to freely move within an infinitesimal range, enabling differentiation of pixel values with respect to the edge location.
This leads to the greatest advantage of our formulation;
it allows us to greatly simplify both the theory and computation, since all edges are axis-aligned.

\subsection{Pixel Pairs}
\label{subsec:pixel_pairs}

Let's analyze two adjacent pixels, A and B, separated by an edge, with their intensities represented as $I_A$ and $I_B$ respectively (refer to Figure~\ref{fig:pixel_pair} for details).
The pixel intensities can be expressed using the Heaviside step function to account for the discontinuity at the edge as in Eqn.~\eqref{eq-half}:
\begin{equation}
  \begin{aligned}
  I_A &= \iint_{D_A}^{} \theta(\alpha(x,y)) f_a(x,y) + \theta(-\alpha(x,y)) f_b(x,y) \,dx\,dy \\
  I_B &= \iint_{D_B}^{} \theta(-\alpha(x,y)) f_a(x,y)  + \theta(\alpha(x,y)) f_b(x,y) \,dx\,dy.
    \end{aligned}
  \label{eq-pixel_pair}
\end{equation}
Without loss of generality, assuming a horizontal pair we define $\alpha(x,y) = p_{AB} - x$, where $p_{AB}$ is the edge location.
Edge location $p_{AB}$ is now the scene parameter with respect to which we want to differentiate the loss function.
Since $p_{AB}$ effects both $I_A$ and $I_B$:
\begin{equation}
  \frac{\partial L}{\partial p_{AB}} = \frac{\partial L}{\partial I_A} \frac{\partial I_A}{\partial p_{AB}} +
  \frac{\partial L}{\partial I_B} \frac{\partial I_B}{\partial p_{AB}},
  \label{eq-pixel_pair_loss}
\end{equation}
where $\frac{\partial I_A}{\partial p_{AB}}$ and $\frac{\partial I_B}{\partial p_{AB}}$ derivable by applying Eqn.~\eqref{eq-integrating_dirac} to Eqn.~\eqref{eq-pixel_pair}.
Unfortunately, with the edge strictly on the pixel boundary, we face a $C1$ discontinuity at $p_{AB}=0$, making intensity values non-differentiable.
Infinitesimal movement of the edge affects $I_B$ when moving right and $I_A$ when moving left, without affecting the other.
We still can compute one-sided limits of the derivatives, e.g. $\frac{\partial I_A}{\partial p}^-$ from the negative direction and $\frac{\partial I_B}{\partial p}^+$ from the positive direction.
Applying Eqn.~\eqref{eq-integrating_dirac} and taking into account that the term $\|  \nabla_{x,y}\alpha(x,y)  \|$  becomes $1$ as we obtain:
\begin{equation}
  \frac{\partial I_B}{\partial p_{AB}}^- = \frac{\partial I_A}{\partial p_{AB}}^+ = 0; \ \frac{\partial I_B}{\partial p_{AB}}^+ = \frac{\partial I_A}{\partial p_{AB}}^- = \int_{x=0}^{}  \left [ f_a(x,y) - f_b(x,y) \right ] \,dy
  \label{eq-pixel_pair_loss2}
\end{equation}
We can circumvent the problem of $C1$ discontinuity by averaging the one-sided limits of the derivatives from the left and right:
\begin{equation}
  \frac{\partial I_B}{\partial p_{AB}} = \frac{\partial I_A}{\partial p_{AB}} = \frac{1}{2} \int_{x=0}^{}  \left [ f_a(x,y) - f_b(x,y) \right ] \,dy
  \label{eq-pixel_diff}
\end{equation}
Assuming that the pixels have constant value across all of their area (recall the rule that ``pixels are always either fully covered or not covered at all'' ) and substituting Eqn.~\eqref{eq-pixel_diff} into Eqn.~\eqref{eq-pixel_pair_loss} we deduce:
\begin{equation}
  \boxed{\frac{\partial L }{\partial p_{AB}} =  \frac{1}{2} \left ( \frac{\partial L}{\partial I_A} + \frac{\partial L}{\partial I_B} \right )   \left (I_A - I_B \right ) + \Omega},
  \label{eq-chain_new}
\end{equation}
which is one of our main results.
Here, $\frac{\partial L }{\partial p_{AB}}$ is the gradient of the loss function with respect to the position of the edge between pixels $A$ and $B$.
Additionally, we reintroduce $\Omega$ which is the second term on the right-hand side of Eqn.~(\ref{eq-product_rule})  that accounts for differentiating smooth regions of $f$.
This term was omitted in Eqn.(\ref{eq-pixel_pair_loss2}) and Eqn.~(\ref{eq-pixel_diff}) for brevity.

Now, we need to accumulate these gradients into the gradients of the vertex positions of the triangle that created those edges.
We do that first by scattering the gradients $\frac{\partial L }{\partial p_{AB}}$ to the fragments of the triangles associated with the pixels forming the edge and then by invoking the backward function of the interpolator module to gather the fragment gradients into the vertex gradients.
Next, we will take a look at the gradient scattering mechanism.

\subsection{Edge Classification and Gradient Scattering}
\label{subsec:edge_classification}

To backpropagate the gradients to the scene parameters we need to compute the gradient with respect to fragment positions, while in~\S\ref{subsec:pixel_pairs} we only derived the gradient $\frac{\partial L}{\partial p_{AB}}$ with respect to the edge position.
We scatter each edge gradient  $\frac{\partial L}{\partial p_{AB}}$ to adjacent fragments where we accumulate the fragment gradients, as illustrated in Figure~\ref{fig:gradient_flow}.
Using the chain rule, we obtain fragment gradients be multiplying $\frac{\partial L}{\partial p_{AB}}$ with $\frac{\partial p_{AB}}{\partial r}$; the later relates the movement of the edge to the movement of the fragment, where $r$ is the position of the fragment.
The computation of $\frac{\partial p_{AB}}{\partial r}$ depends on the type of the edge formed by the fragments.
We consider several cases: overhanging primitives (either one another or background); adjacent primitives; intersecting primitives.
Unlike Laine {\em et al.}~\cite{laine2020modular}, we avoid additional data structures like hash maps for connectivity.

\begin{figure}[!tb]
  \begin{minipage}{0.49\textwidth}
    \centering
    \includegraphics[width=1.0\columnwidth]{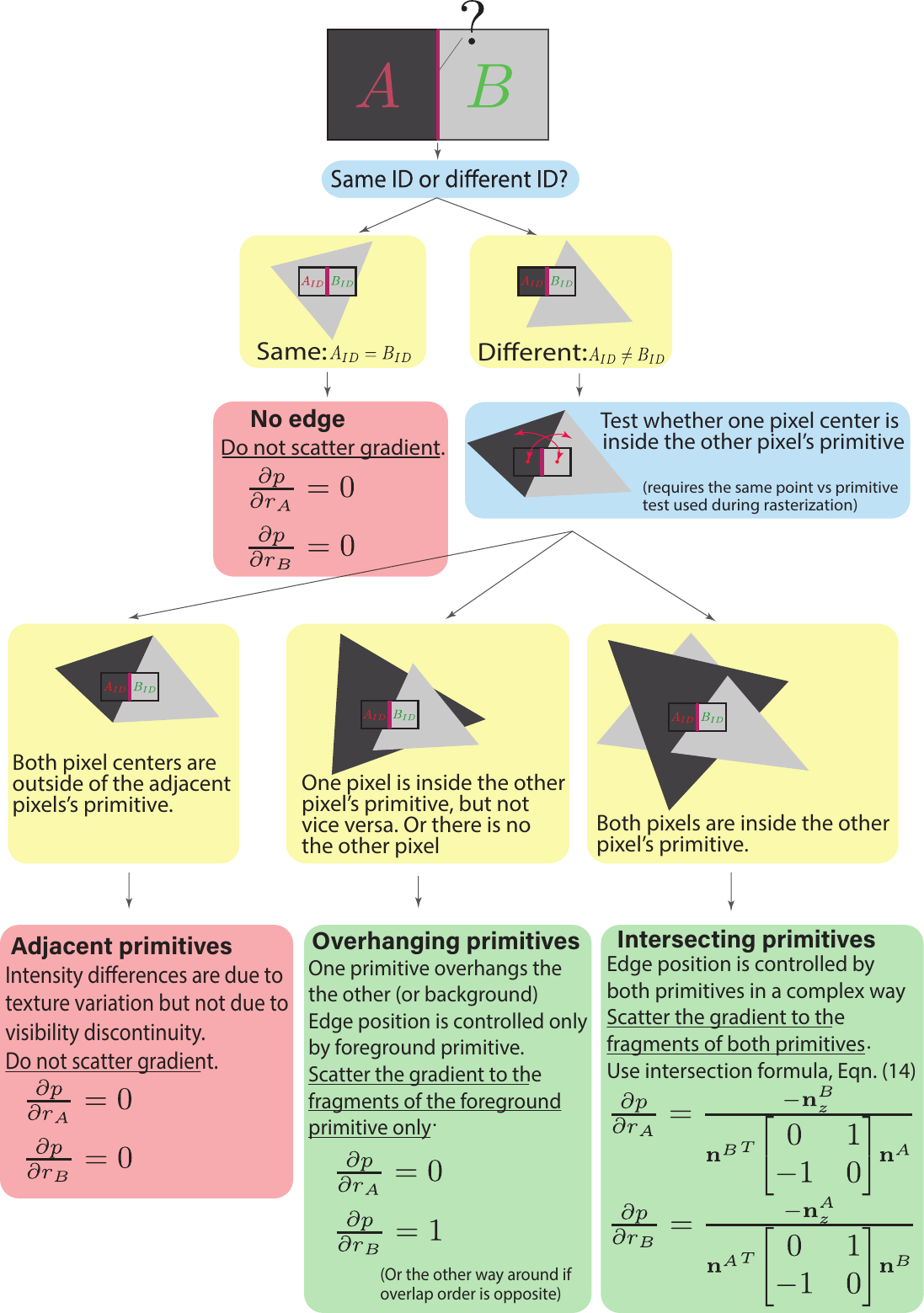}
    \caption{\textbf{Edge classification}. The accompanying figure provides a schematic representation of our edge classification process. Edges are categorized into four types: no edge, adjacent primitives, overhanging primitives, and intersecting primitives.
    }
    \label{fig:edge_classification}
  \end{minipage}\hfill
  \begin {minipage}{0.49\textwidth}
    \centering
    \includegraphics[width=1.0\columnwidth]{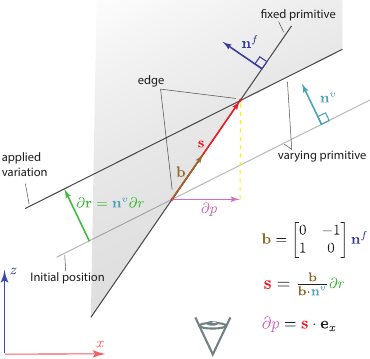}
    \caption{\textbf{Geometry intersection}. A geometrical derivation for the partial derivative $\frac{\partial p}{ \partial r}$, essential for converting edge-related gradients to fragment-related gradients in cases where edges are formed by geometry intersections.
    This shows the x-z plane, where the render camera is viewing the scene along the z axis so that edges are parallel to the $y$ axis (and perpendicular to the drawing). Projections of both intersecting primitives are transformed into two lines. }
    \label{fig:intersection}
  \end{minipage}
\end{figure}

Our approach examines each neighboring pixel pair, comparing their triangle IDs, see Figure~\ref{fig:edge_classification} for illustration.
Matching IDs indicate no edge; differing IDs prompt further tests.
A single valid ID suggests a primitive overhanging the background.
Testing a pixel's center against the triangle of the other pixel discerns overhangs and self-intersections.
Non-contained pixels in each other's triangles imply adjacency.
If both pixel centers are in the triangle of the other pixel, it implies self-intersections, and if in just one - then overhanging.
For adjacent primitives we do not scatter gradients at all, since it is likely due to primitives sharing an edge and there is no discontinuity.
Please note that even in that case $\frac{\partial L}{\partial p_{AB}}$ may be non-zero due to $\Omega$ contribution.
In the overhanging case, we scatter the gradient only to the primitive on top, as the covered primitive can not affect the position of the edge.
Thus, in such cases the covered fragment has $\frac{\partial p}{\partial r} = 0$ while the foreground fragment has $\frac{\partial p}{\partial r} = 1$ reflecting that the fragment movement is rigidly tied to the movement of the edge.
Please refer to supplementary material for more details.

\subsection{Geometry Intersections}
\label{subsec:geom_intersections}

This subsection details computing the derivative of edge location with respect to fragment position $\frac{\partial p}{\partial r}$ in scenarios where the edge arises from intersecting geometry.
This situation is inherently complex, and a general-case solution would be challenging.
However, our micro-edge formulation streamlines the problem because intersection lines projected onto the image plane comprised of micro-edges, e.i.\ either vertical or horizontal.
This allows us to analyze the problem in 2D, focusing on the x-z or y-z planes, with the z-axis perpendicular to the image plane.
We derive the derivative of the edge location by analyzing two fragments of a micro-edge that belongs to the intersection line, focusing on how the edge would move in response to infinitesimal movements of each fragment.
Without loss of generality we consider the x-z plane:
%
\begin{equation}
  \begin{aligned}
     \frac{\partial p}{\partial r} = -\mathbf{n}^f_z  \left [ {\mathbf{n}^f}^T \begin{bmatrix} 0 & 1 \\ -1 & 0 \end{bmatrix} \mathbf{n}^v \right ]^{-1},
    \end{aligned}
  \label{eq-grad_int}
\end{equation}
where $p$ is edge position, $r$ is the fragment position (in 3D clip space), $\mathbf{n}^v$  is the varying primitive's normal, and $\mathbf{n}^f$ is the fixed primitive's normal.
In this context, the $x$-axis lies on the image plane, and the $z$-axis is perpendicular to it, see Fig.~\ref{fig:intersection}.
We compute $\frac{\partial p}{\partial r}$ for both fragments that form the intersection edge by first fixing one fragment and varying the other, and then switching them.
Please refer to supplementary material for derivation details.

\subsection{Theoretical analysis of the approximation}
\label{subsec:error_analysis}
At first glance, replacing arbitrarily oriented edges with vertical and horizontal micro-edges might seem overly simplistic, raising doubts about accuracy.
It could be argued, for example, that the perimeter defined by micro-edges doesn't match the original as micro-edge size approaches zero.
However, applying the two-dimensional version of the {\em divergence theorem} from vector calculus offers insight:
\begin{equation}
\oint _{C}\mathbf {f} \cdot \mathbf {n} \,dt = \iint _{D}\left(\nabla_{\Phi} \cdot \mathbf {f} \right) \,dx\,dy,
  \label{eq-green}
\end{equation}
where $\mathbf {n}$ is normal to boundary $C$, within which the integrand function is continuous and differentiable.
We are free to pick any such boundaries that includes a discontinuity.
In our case, the term $\frac{\partial \alpha(x,y)}{\partial \Phi}$
in Eqn.~\eqref{eq-integrating_dirac}
is a normal vector to the boundary $\alpha(x,y)=0$ thus the integral is equivalent to the left-hand side of Eqn.~\eqref{eq-green}.
According to Eqn.~\eqref{eq-green}, this recasts the boundary derivative integral as an area integral.
The difference between the area integrals of micro-edges and original edges is confined to a space defined by the micro-edges size.
As pixel size diminishes, the integral approaches the exact value, and the discrepancy area vanishes, confirming the micro-edge approach converges to the same solution as the original formulation in the limit.

\section{Experimental Analysis}
\label{sec:analysis}

We implement our method in PyTorch with custom CUDA kernels, and largely follow the structure of nvdiffrast~\cite{laine2020modular}, with the exception for the antialiasing module wich we replace with our EdgeGrad module.
For all experiments we utilize Laplacian preconditioning proposed by Nicolet {\em et al.}~\cite{nicolet2021large}.
We find Laplacian preconditioning extremely helpful for stabilizing training and we use it for both, meshes and texture data.
While Laplacian preconditioning may reduce self-intersections, this happens only due to much smoother optimization process, but it does not account for them in any way.

\newcommand{\myparagraph}[1]{ \textit{#1}}

\begin{figure}[!tb]
    \centering
    \includegraphics[width=0.95\textwidth]{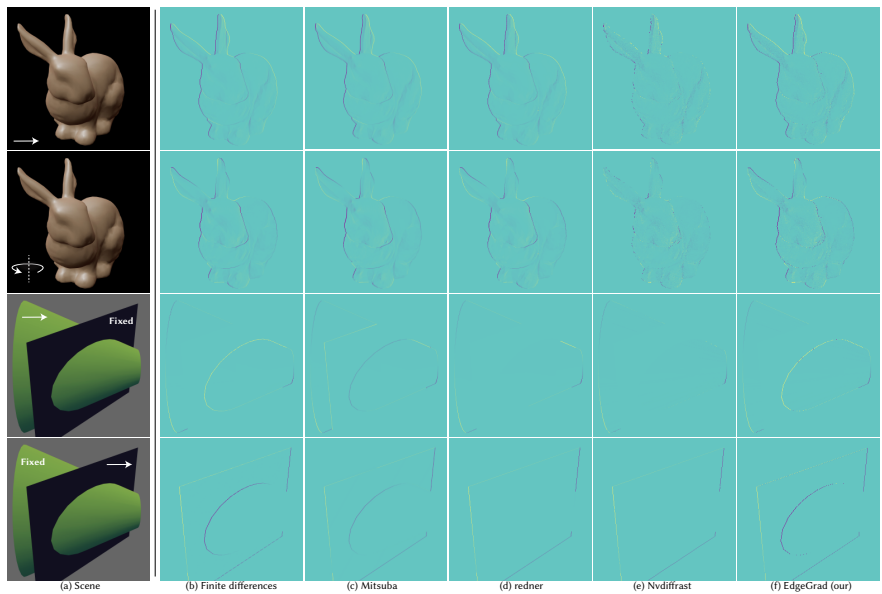}
    \caption{\textbf{Comparison of forward gradient} on several test scenes with the numerical solution using finite differences. (a) Synthetic test scene, we show gradients with respect to a parameter.
    We use (b) finite differences as a reference, and compare with (c) Mitsuba 3~\cite{NimierDavidVicini2019Mitsuba2,Jakob2020DrJit}, (d) redner~\cite{Li:2018:DMC}, (e) nvdiffrast~\cite{laine2020modular}, and (f) our edge gradient approach.
    }
    \label{fig:finite_difference}
\end{figure}

\myparagraph{Forward gradients.} We first analyze the performance of the proposed approach by evaluating the forward gradients on a variety of test cases.
Forward gradients are per-pixel derivatives with respect to some scene parameter, which is depicted with white arrows on the scene images.
%
Fig.~\ref{fig:finite_difference} shows the forward gradients computed using finite differences, raytracing methods Mitsuba~\cite{NimierDavidVicini2019Mitsuba2,jakob2022mitsuba3,Jakob2020DrJit} and redner~\cite{Li:2018:DMC}, and rasterization approaches including nvdiffrast~\cite{laine2020modular} and our approach (EdgeGrad).
Unlike other methods, our approach still matches finite differences gradients in cases of intersecting geometry.
While Mitsuba and Redner, which require multiple samples per pixel (256 in these comparisons), typically offer higher accuracy, they fall short at geometry intersections.
Nvdiffrast is noisier because it assumes triangles containing silhouette edges always overlap pixel centers, which often isn't the case, leading to omitted gradients.

\begin{figure}[!p]
  \begin{minipage}{0.49\textwidth}
           \resizebox{0.9\columnwidth}{!}{%
            \begin{tabular}{r|ccc}
                Scene
                &
                \begin{minipage}{.23\textwidth}
                    \includegraphics[width=\linewidth]{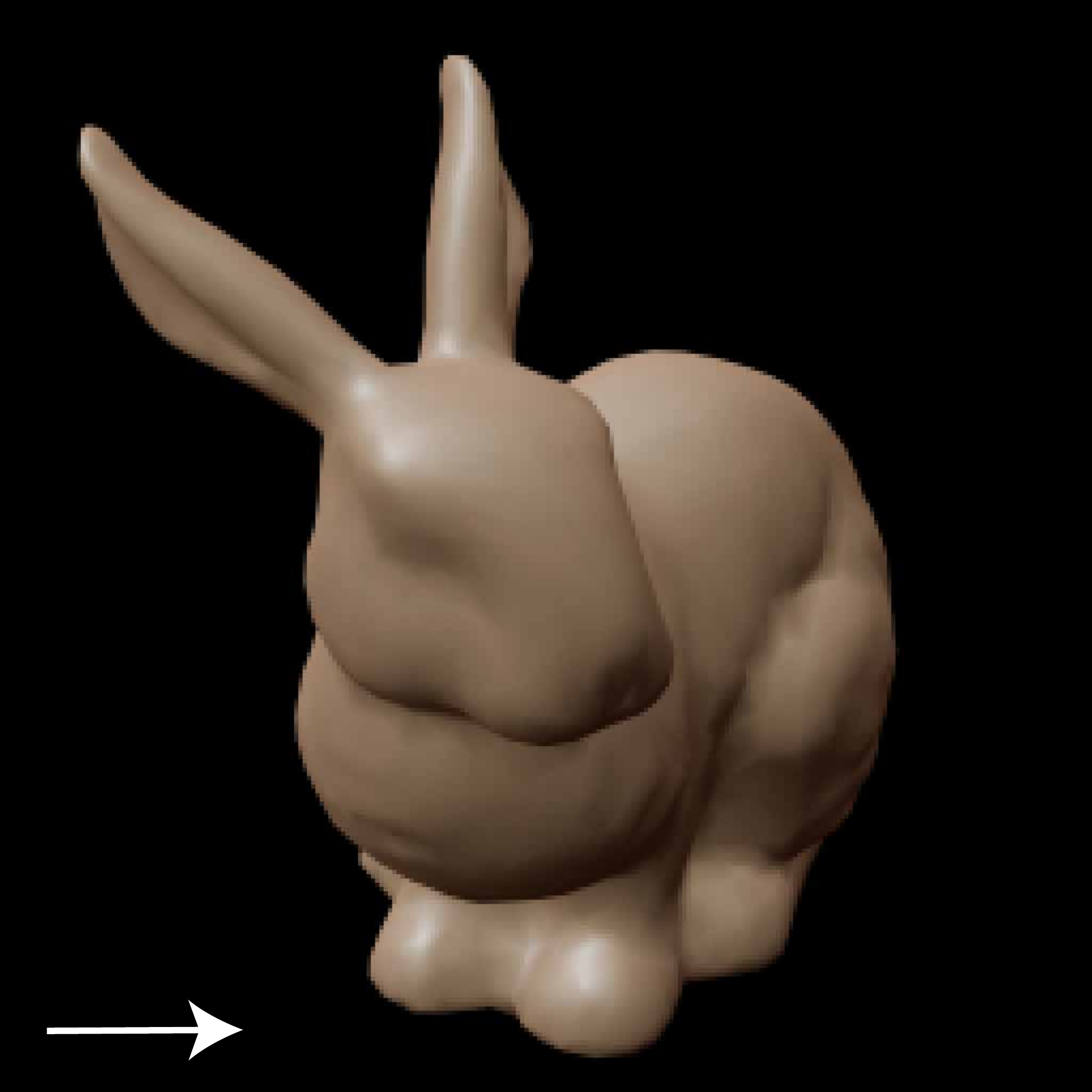}
                \end{minipage}
                &
                \begin{minipage}{.23\textwidth}
                    \includegraphics[width=\linewidth]{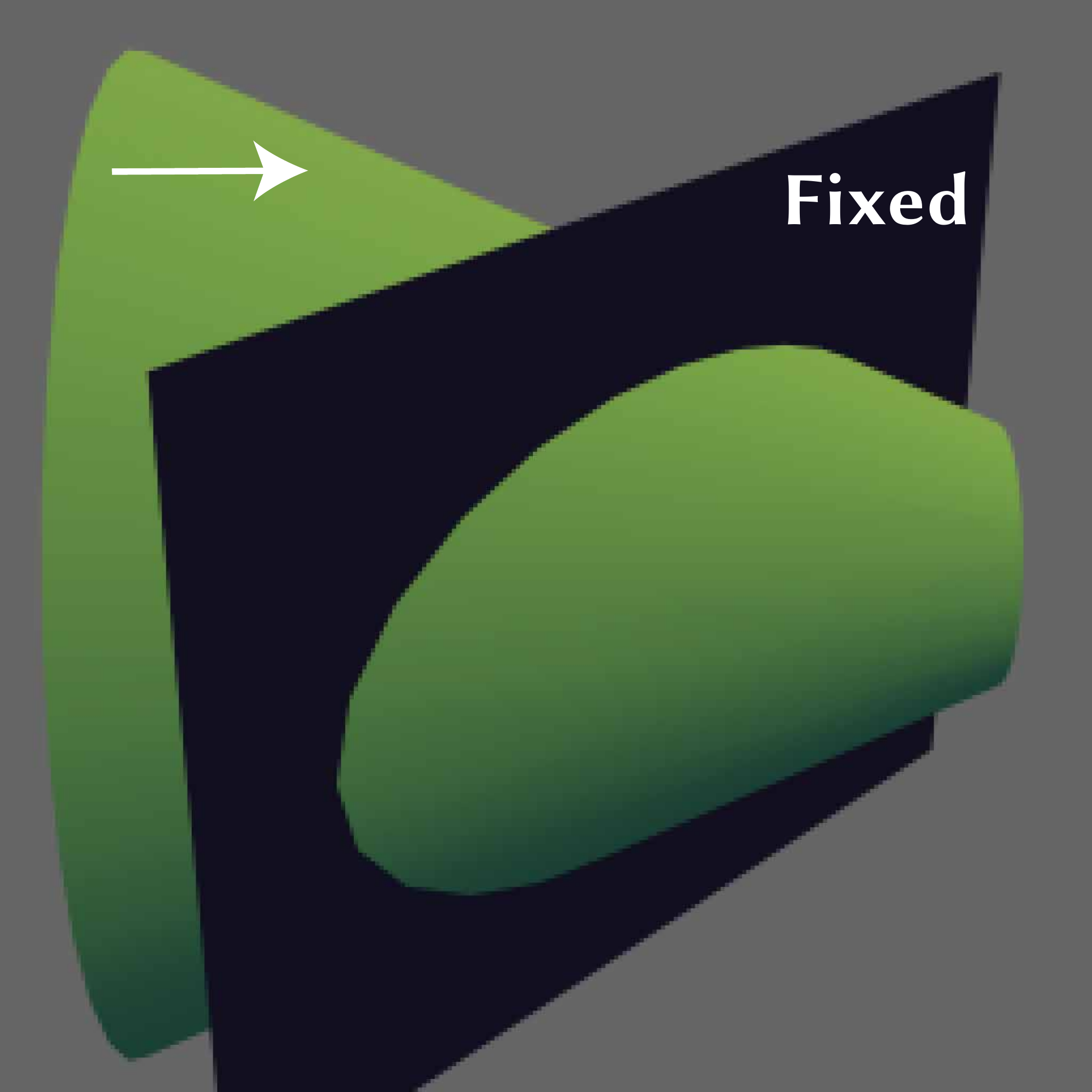}
                \end{minipage}
                &
                \begin{minipage}{.23\textwidth}
                    \includegraphics[width=\linewidth]{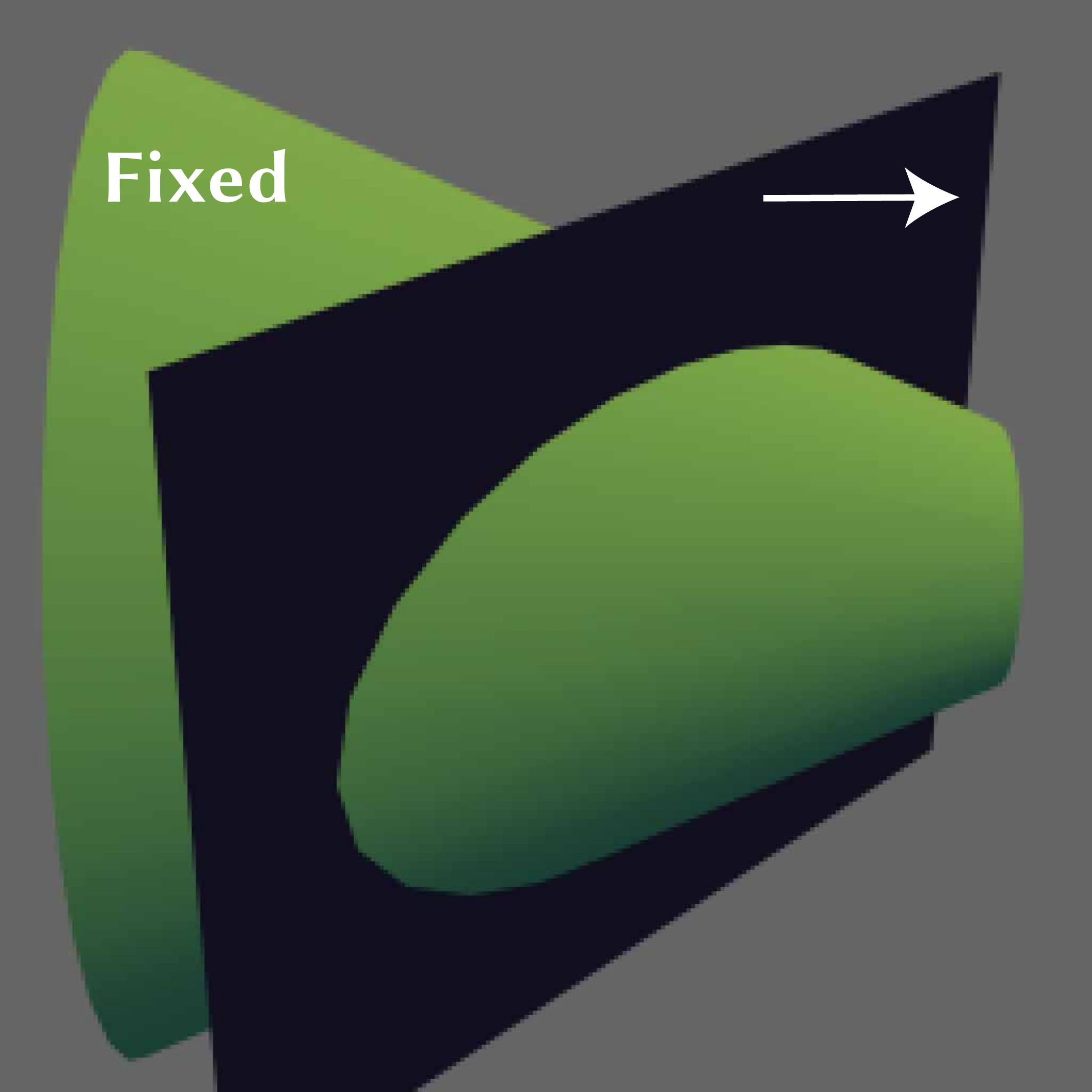}
                \end{minipage}
                \\

                \hline
                redner~\cite{Li:2018:DMC}          & \cellcolor{tabsecond} 1.20\%  & \cellcolor{tabsecond} 53.52\%
                & \cellcolor{tabthird} 37.94\%
                \\
                Mitsuba3~\cite{NimierDavidVicini2019Mitsuba2}     & \cellcolor{tabfirst} 0.67\% & 75.83\%                       & \cellcolor{tabsecond} 37.59\% \\
                Nvdiffrast~\cite{laine2020modular} & 45.91\%                      & \cellcolor{tabthird}  69.33\%   & 39.42\%                       \\
                EdgeGrad(our)                      & \cellcolor{tabthird} 6.01\%  & \cellcolor{tabfirst}  3.35\%    & \cellcolor{tabfirst} 8.35\%   \\
            \end{tabular}}
        \captionof{table}{
            \textbf{Accuracy of backward gradients. Relative error, \% ($\downarrow$)}.
            This table shows relative errors in backward gradient computations for test scenes. Second and third scenes include geometry intersections, emphasizing our method's advantage in managing these complexities.}
        \label{fig:backward_grad}
  \end{minipage}\hfill
  \begin {minipage}{0.49\textwidth}
    \centering
    \resizebox{1.0\textwidth}{!}{%
             \includegraphics[width=\textwidth]{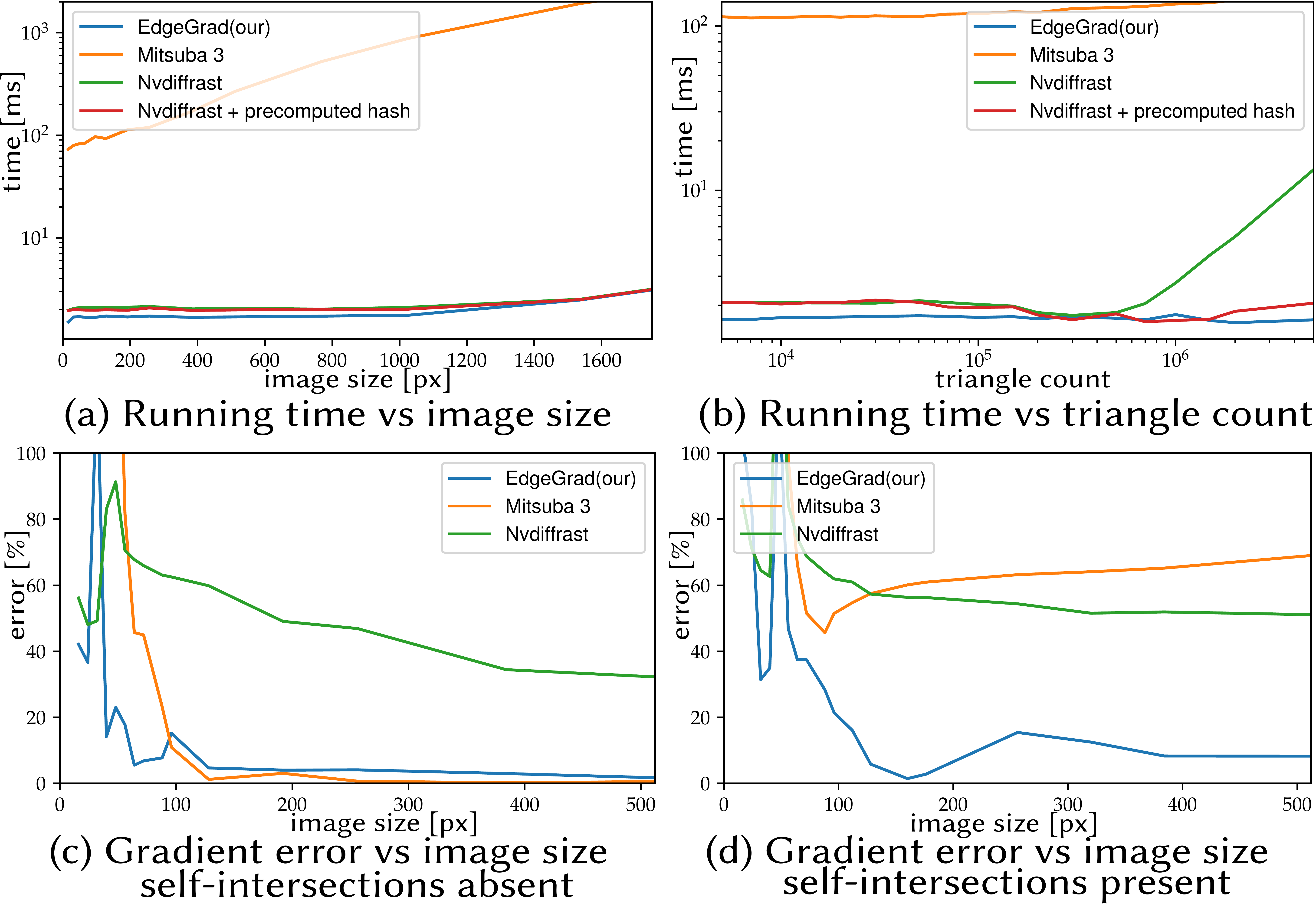}
    }
        \caption{\textbf{Runtime performance and errors}. a) Runtime [ms] by image size [px]; b) runtime [ms] by triangle count; c) gradient error [\%] by image size [px] without self-intersections; d) gradient error [\%] by image size [px] with self-intersections.
    }
    \label{fig:runtime}
  \end{minipage}
\end{figure}

\myparagraph{Backward gradients.} Tab.~\ref{fig:backward_grad} quantitatively analyzes the accuracy of the backward gradients.
We show improved results especially for complex intersecting geometry.
Compared to raytracing methods, our rasterization-based approach yields comparable performance at a fraction of the computational cost.

\myparagraph{Runtime performance end accuracy by image size.} Fig.~\ref{fig:runtime} shows running time of our method (EdgeGrad) compared to Mitsuba~\cite{NimierDavidVicini2019Mitsuba2} and nvdiffrast~\cite{laine2020modular}.
Our method significantly outperforms other approaches in running time on all variety of image sizes and triangle count.
Our method has comparable accuracy to Mitsuba~\cite{NimierDavidVicini2019Mitsuba2} (and even better for the case with geometry intersections) at large image sizes, but performs worse for smaller image size due to sub pixel triangles.

\begin{figure}[!p]
    \centering
    \includegraphics[width=1.\textwidth]{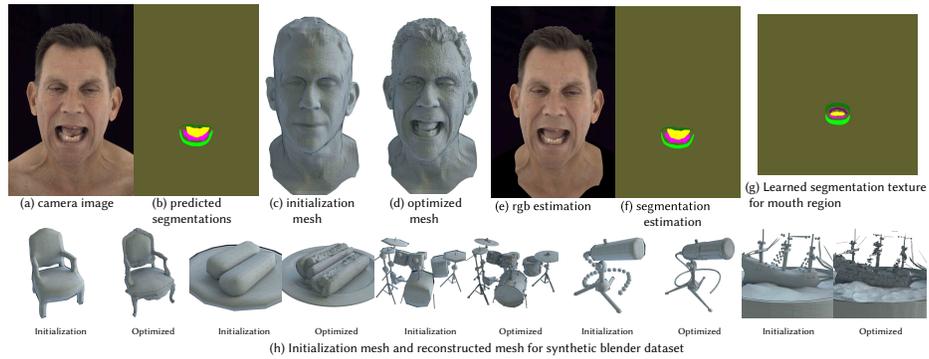}
    \caption{\textbf{Experimental setup details}. The figure shows: a) the ground truth camera image; b) predicted segmentation mask for mouth region using HRNet~\cite{wang2020deep}; c) initialization mesh from 3DMM; d) optimized avatar mesh; e) rendered image of the optimized model; f) rendered segmentation estimation image; g) learned static segmentation texture; h) initialization meshes and optimized meshes for blender dataset~\cite{mildenhall2021nerf}.
    }
    \label{fig:details}
\end{figure}

\begin{figure}[!p]
    \centering
    \includegraphics[width=1.\textwidth]{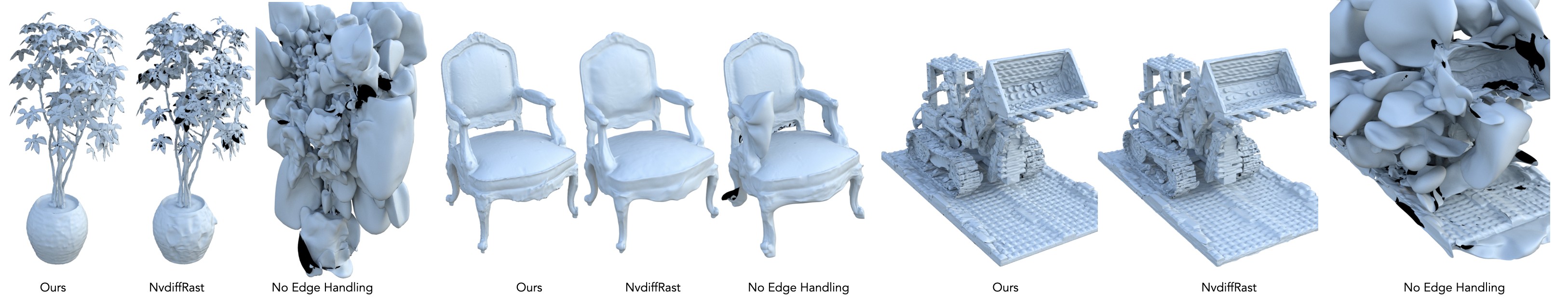}
    \caption{Reconstructions from inverse rendering compare our method with nvdiffrast~\cite{laine2020modular} and rasterization without discontinuity handling. Although nvdiffrast can match the outline of objects well, it suffers from interpenetration (see leaves of Ficus) and does not capture as much detail as ours (see the backrest of the chair and ground of lego). The method with no edge handling fails completely on these objects.
    }
    \label{fig:Comparison}
\end{figure}

\setlength{\tabcolsep}{4pt}
\begin{table}[!tbp]
    \centering
    \resizebox{0.95\textwidth}{!}{%
    \begin{tabular}{cl|cccccccc}
    \multirow{ 5}{*}{PSNR $\uparrow$ \ } & & Lego & Chair & Ship & Hotdog & Ficus & Drums & Mic & Materials \\
    \cline{2-10}
    &  Continuous only &   16.342 &   28.151 &   15.693 &   27.850 &   18.399 &   15.534 &   21.797 &   17.11 \\
    &  Nvdiffrast~\cite{laine2020modular} &  \cellcolor{tabthird} 29.44 &  \cellcolor{tabthird} 29.804 &  \cellcolor{tabthird} 26.173 &  \cellcolor{tabthird} 31.684 &  \cellcolor{tabthird} 24.964 &  \cellcolor{tabthird} 23.154 &  \cellcolor{tabthird} 29.638 &  \cellcolor{tabthird} 24.413 \\
    &  EdgeGrad -intersect.\  &  \cellcolor{tabsecond} 29.574 &  \cellcolor{tabfirst} 33.077 &  \cellcolor{tabfirst} 26.731 &  \cellcolor{tabfirst} 34.225 & \cellcolor{tabfirst} 26.781 &  \cellcolor{tabfirst} 24.08 &  \cellcolor{tabsecond} 31.224 &  \cellcolor{tabsecond} 25.221 \\
    &  EdgeGrad (our) &  \cellcolor{tabfirst} 29.667 &  \cellcolor{tabsecond} 32.981 &  \cellcolor{tabsecond} 26.65 &  \cellcolor{tabsecond} 34.081 &  \cellcolor{tabsecond} 26.559 &  \cellcolor{tabsecond} 24.014 &  \cellcolor{tabfirst} 31.346 &  \cellcolor{tabfirst} 25.287 \\
    \end{tabular}}

    \resizebox{0.95\textwidth}{!}{%
    \begin{tabular}{cl|cccccccc}
    \multirow{ 5}{*}{SSIM $\uparrow$ \ \ } & & Lego & Chair & Ship & Hotdog & Ficus & Drums & Mic & Materials \\
    \cline{2-10}
    &  Continuous only & 0.7550 &  \cellcolor{tabthird} 0.9531 &   0.6929 &   0.9374 &   0.8245 &   0.7723 &   0.9182 &   0.7693 \\
    &  Nvdiffrast~\cite{laine2020modular} &  \cellcolor{tabthird} 0.9467 &   0.9416 &  \cellcolor{tabthird} 0.8316 &  \cellcolor{tabthird} 0.9473 &  \cellcolor{tabthird} 0.9323 &  \cellcolor{tabthird} 0.8971 &  \cellcolor{tabthird} 0.9647 &  \cellcolor{tabthird} 0.8865 \\
    &  EdgeGrad -intersect.\  &  \cellcolor{tabsecond} 0.9485 &  \cellcolor{tabfirst} 0.9765 &  \cellcolor{tabfirst} 0.8478 &  \cellcolor{tabfirst} 0.9715 &  \cellcolor{tabfirst} 0.9468 &  \cellcolor{tabsecond} 0.9216 &  \cellcolor{tabsecond} 0.9756 &  \cellcolor{tabsecond} 0.8945 \\
    &  EdgeGrad (our) &  \cellcolor{tabfirst} 0.9501 &  \cellcolor{tabsecond} 0.9764 &  \cellcolor{tabsecond} 0.8472 &  \cellcolor{tabsecond} 0.9713 &  \cellcolor{tabsecond} 0.9452 &  \cellcolor{tabfirst} 0.9220 &  \cellcolor{tabfirst} 0.9767 &  \cellcolor{tabfirst} 0.8972 \\
    \end{tabular}}

    \resizebox{0.95\textwidth}{!}{%
    \begin{tabular}{cl|cccccccc}
    \multirow{ 5}{*}{LPIPS $\downarrow$ \ } & & Lego & Chair & Ship & Hotdog & Ficus & Drums & Mic & Materials \\
    \cline{2-10}
    &  Continuous only & 0.3223 &  \cellcolor{tabthird} 0.0611 &   0.3913 &   0.1109 &   0.2046 &   0.2943 &   0.1272 &   0.2511 \\
    &  Nvdiffrast~\cite{laine2020modular} &  \cellcolor{tabthird} 0.0653 &   0.0771 &  \cellcolor{tabthird} 0.2020 &  \cellcolor{tabthird} 0.0936 &  \cellcolor{tabthird} 0.0781 &  \cellcolor{tabthird} 0.1116 &  \cellcolor{tabthird} 0.0549 &  \cellcolor{tabthird} 0.1291 \\
    &  EdgeGrad -intersect.\  &  \cellcolor{tabsecond} 0.0633 &  \cellcolor{tabsecond} 0.0348 &  \cellcolor{tabsecond} 0.1899 &  \cellcolor{tabfirst} 0.0546 &  \cellcolor{tabsecond} 0.0681 &  \cellcolor{tabsecond} 0.0871 &  \cellcolor{tabsecond} 0.0439 &  \cellcolor{tabsecond} 0.1200 \\
    &  EdgeGrad (our) &  \cellcolor{tabfirst} 0.0614 &  \cellcolor{tabfirst} 0.0341 &  \cellcolor{tabfirst} 0.1850 &  \cellcolor{tabsecond} 0.0548 &  \cellcolor{tabfirst} 0.0672 &  \cellcolor{tabfirst} 0.0870 &  \cellcolor{tabfirst} 0.0399 &  \cellcolor{tabfirst} 0.1185 \\
    \end{tabular}}

    \caption{Per-scene quantitative results from the realistic synthetic blender dataset~\cite{mildenhall2021nerf}, featuring scenes with complex geometries and non-Lambertian materials rendered using Blender's Cycles pathtracer.
    }
    \label{table:suppresults2}
\end{table}
\setlength{\tabcolsep}{1.4pt}

\myparagraph{Quantitative scene reconstructions.} Tab.~\ref{table:suppresults2} shows quantitative results of scene reconstruction on blender dataset~\cite{mildenhall2021nerf}.
We follow the testing procedure described in~\cite{mildenhall2021nerf} and report PNSR, SSIM and LPIPS metrics.
We compare our method (EdgeGrad) with a "continues only" baseline (which ignores gradients from discontinuities),  nvdiffrast~\cite{laine2020modular}, and with a variant of our method that does not have a special handling for the geometry intersections.
Method optimization procedure is similar to Nicolet {\em et al.}~\cite{nicolet2021large} we only add MLP for decoding texture values in a view conditioned manner similar to~\cite{Ma2021PixelCA}.
We initialize geometry with some coarse manually created meshes shown in Fig.~\ref{fig:details}.
To avoid additional complexity we opted out automatic remeshing introduced by Nicolet {\em et al.} which would necessitate updating UV parametrization in our case.

\myparagraph{Qualitative mesh reconstruction.} Qualitatively, we compare the mesh reconstructions obtained on the public NeRF dataset~\cite{mildenhall2021nerf} using our approach, a naive approach with no special edge handling, and nvdiffrast~\cite{laine2020modular} in Fig.~\ref{fig:Comparison}.
All methods use the same shading model.
Our approach shows improved detail in the recovered meshes.

\section{Application to Dynamic Head Scenes}
\label{sec:application}

\begin{figure}[!t]
    \centering
    \includegraphics[width=\textwidth]{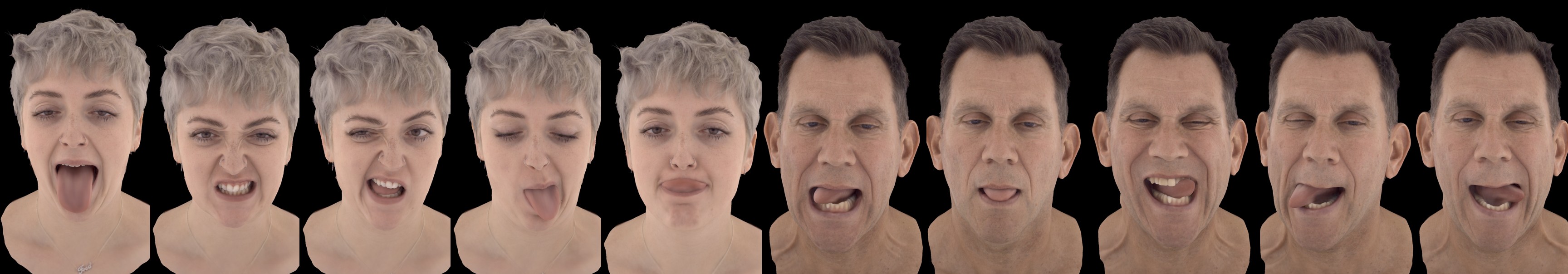}
    \caption{Example renderings of mesh-based avatars from~\cite{Ma2021PixelCA} trained using EdgeGrad, capturing complex inner mouth movements, including teeth and tongue dynamics.}
\label{fig:CA}
\end{figure}

\begin{SCfigure}[][!t]
\resizebox{0.6\textwidth}{!}{%
         \includegraphics[width=\textwidth]{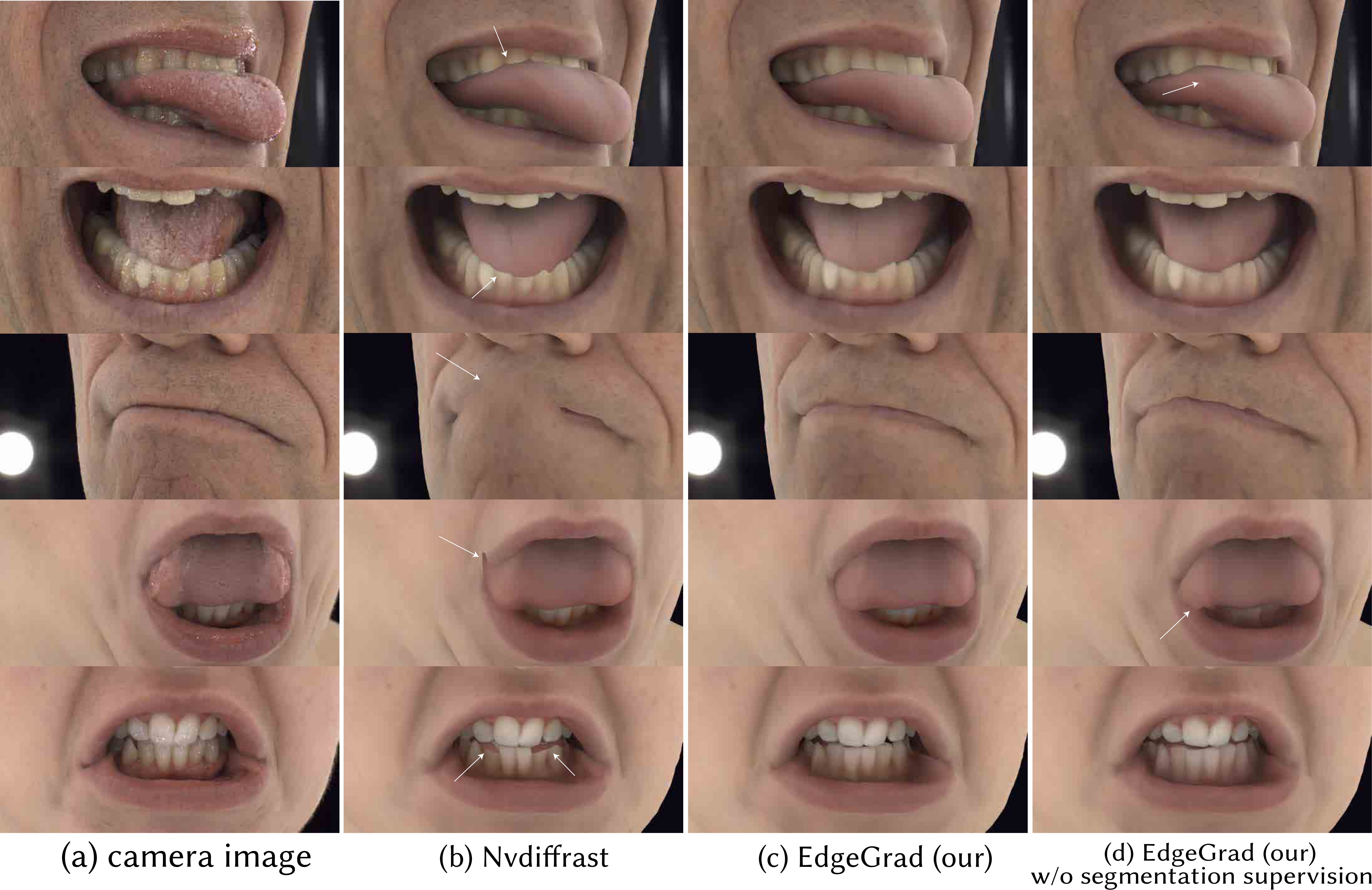}
}
    \caption{\textbf{Qualitative evaluation and ablation on dynamic head scene}. We compare: (c) our method, (a) ground truth images, (b) Nvdiffrast~\cite{laine2020modular}, and ablated version of our method without including segmentation mask. The avatar model is from~\cite{Ma2021PixelCA} trained using EdgeGrad and Nvdiffrast. We highlight artifacts with white arrows.
}
\label{fig:CA_ttl}
\end{SCfigure}

We show the importance of correct edge gradients on a head modeling application by fitting a personalized mesh-based avatar representation to facial performances captured in a multiview lightstage.
modifying the avatar creation method from ~\cite{Ma2021PixelCA}.
%
As the facial avatar representation, we use an encoder-decoder architecture to produce mesh vertices and texture-mapped appearance features~\cite{lombardi2018deep}, which are shaded using a pixel-wise MLP producing RGB appearance~\cite{Ma2021PixelCA}.
We use a coarse face tracking produced using traditional computer vision as initialization for inverse rendering.
Importantly, this coarse face template has a small number of vertices (8k) and no geometry in the interior of the mouth.

Traditionally, fitting the interior of the mouth without strong priors is a complex task due to poor observability and intricately interacting geometry with constant contact.
Prior work either relies on artists to create alignments, volumetric simulation to prevent self-intersections~\cite{ichim2017phace}, or on very specific mouth interior capture methods to build priors~\cite{wu2016modelteeth,medina2022speechtongue}.
Instead, we show that by using differentiable rasterization with correct edge gradients, we can achieve fully automatic geometric fitting of the complex geometry of the face and mouth interior including lips, teeth, and tongue using a mesh-based model of fixed topology.
We modify ~\cite{Ma2021PixelCA} in two ways.
First, we use the edge gradient rasterization method of Sec.~\ref{sec:method}.
Second, we add supervision in the form of 2D segmentation masks with separate labels for the regions of the teeth, tongue, and lips, for details see Fig.~\ref{fig:details}.
While we can achieve accurate per-frame geometry without the additional segmentation loss supervision, we find that segmentation losses help anchor the teeth and tongue geometry to a consistent region of the UV map.
%
Fig.~\ref{fig:CA} shows the quality of geometric fitting achievable with this inverse rendering method.
Note that we accurately reconstruct the complex contact geometry of the tongue sliding across teeth, teeth-to-teeth contact, as well as the tongue pushing against the lips and cheeks.
Fig.~\ref{fig:CA_ttl}demonstrates reconstruction quality of complex inner mouth details, which are challenging due to frequent self-intersections.
We also include results without 2D segmentation mask supervision.

\section{Conclusions and Limitations}
\label{sec:conclusion}
We have presented a fast method for rasterization-based differentiable rendering that has accurate gradients across visibility discontinuities such as triangle edges, occlusions and geometry intersections.
Our results demonstrate enhanced mesh reconstruction fidelity across diverse inverse rendering scenarios, comparable to implicit and volumetric methods.
Our method also unlocks new applications, accurate mouth interior fitting in dynamic facial performance capture, through fully automatic inverse rendering.
Regarding limitations, our method inherits the constraints of rasterization, including the inability to manage transparency, lack of precise antialiasing without multisampling, and the absence of physically based rendering and global illumination.
It may also inaccurately interpret adjacent primitives in finely tessellated meshes, leading to sub-pixel triangles.

\bibliographystyle{splncs04}
\bibliography{references}

\begin{thebibliography}{10}
\providecommand{\url}[1]{\texttt{#1}}
\providecommand{\urlprefix}{URL }
\providecommand{\doi}[1]{https://doi.org/#1}

\bibitem{baek2019pushing}
Baek, S., Kim, K.I., Kim, T.K.: Pushing the envelope for rgb-based dense 3d
  hand pose estimation via neural rendering. In: Proceedings of the IEEE/CVF
  Conference on Computer Vision and Pattern Recognition. pp. 1067--1076 (2019)

\bibitem{bangaru2022differentiable}
Bangaru, S.P., Gharbi, M., Luan, F., Li, T.M., Sunkavalli, K., Hasan, M., Bi,
  S., Xu, Z., Bernstein, G., Durand, F.: Differentiable rendering of neural
  sdfs through reparameterization. In: SIGGRAPH Asia 2022 Conference Papers.
  pp.~1--9 (2022)

\bibitem{bangaru2020unbiased}
Bangaru, S.P., Li, T.M., Durand, F.: Unbiased warped-area sampling for
  differentiable rendering. ACM Transactions on Graphics (TOG)  \textbf{39}(6),
   1--18 (2020)

\bibitem{barron2021mip}
Barron, J.T., Mildenhall, B., Tancik, M., Hedman, P., Martin-Brualla, R.,
  Srinivasan, P.P.: Mip-nerf: A multiscale representation for anti-aliasing
  neural radiance fields. In: Proceedings of the IEEE/CVF International
  Conference on Computer Vision. pp. 5855--5864 (2021)

\bibitem{beker2020monocular}
Beker, D., Kato, H., Morariu, M.A., Ando, T., Matsuoka, T., Kehl, W., Gaidon,
  A.: Monocular differentiable rendering for self-supervised 3d object
  detection. In: Computer Vision--ECCV 2020: 16th European Conference, Glasgow,
  UK, August 23--28, 2020, Proceedings, Part XXI 16. pp. 514--529. Springer
  (2020)

\bibitem{bogo2016keep}
Bogo, F., Kanazawa, A., Lassner, C., Gehler, P., Romero, J., Black, M.J.: Keep
  it smpl: Automatic estimation of 3d human pose and shape from a single image.
  In: Computer Vision--ECCV 2016: 14th European Conference, Amsterdam, The
  Netherlands, October 11-14, 2016, Proceedings, Part V 14. pp. 561--578.
  Springer (2016)

\bibitem{chan2022efficient}
Chan, E.R., Lin, C.Z., Chan, M.A., Nagano, K., Pan, B., De~Mello, S., Gallo,
  O., Guibas, L.J., Tremblay, J., Khamis, S., et~al.: Efficient geometry-aware
  3d generative adversarial networks. In: Proceedings of the IEEE/CVF
  Conference on Computer Vision and Pattern Recognition. pp. 16123--16133
  (2022)

\bibitem{chen2019learning}
Chen, W., Ling, H., Gao, J., Smith, E., Lehtinen, J., Jacobson, A., Fidler, S.:
  Learning to predict 3d objects with an interpolation-based differentiable
  renderer. Advances in neural information processing systems  \textbf{32}
  (2019)

\bibitem{chen2021dib}
Chen, W., Litalien, J., Gao, J., Wang, Z., Fuji~Tsang, C., Khamis, S., Litany,
  O., Fidler, S.: Dib-r++: learning to predict lighting and material with a
  hybrid differentiable renderer. Advances in Neural Information Processing
  Systems  \textbf{34},  22834--22848 (2021)

\bibitem{choy20163d}
Choy, C.B., Xu, D., Gwak, J., Chen, K., Savarese, S.: 3d-r2n2: A unified
  approach for single and multi-view 3d object reconstruction. In: Computer
  Vision--ECCV 2016: 14th European Conference, Amsterdam, The Netherlands,
  October 11-14, 2016, Proceedings, Part VIII 14. pp. 628--644. Springer (2016)

\bibitem{ge20193d}
Ge, L., Ren, Z., Li, Y., Xue, Z., Wang, Y., Cai, J., Yuan, J.: 3d hand shape
  and pose estimation from a single rgb image. In: Proceedings of the IEEE/CVF
  Conference on Computer Vision and Pattern Recognition. pp. 10833--10842
  (2019)

\bibitem{henzler2021unsupervised}
Henzler, P., Reizenstein, J., Labatut, P., Shapovalov, R., Ritschel, T.,
  Vedaldi, A., Novotny, D.: Unsupervised learning of 3d object categories from
  videos in the wild. In: Proceedings of the IEEE/CVF Conference on Computer
  Vision and Pattern Recognition. pp. 4700--4709 (2021)

\bibitem{ichim2017phace}
Ichim, A.E., Kadle\v{c}ek, P., Kavan, L., Pauly, M.: Phace: Physics-based face
  modeling and animation. ACM Trans. Graph.  \textbf{36}(4) (jul 2017)

\bibitem{jakob2022mitsuba3}
Jakob, W., Speierer, S., Roussel, N., Nimier-David, M., Vicini, D., Zeltner,
  T., Nicolet, B., Crespo, M., Leroy, V., Zhang, Z.: Mitsuba 3 renderer (2022),
  https://mitsuba-renderer.org

\bibitem{Jakob2020DrJit}
Jakob, W., Speierer, S., Roussel, N., Vicini, D.: Dr.jit: A just-in-time
  compiler for differentiable rendering. Transactions on Graphics (Proceedings
  of SIGGRAPH)  \textbf{41}(4) (Jul 2022). \doi{10.1145/3528223.3530099}

\bibitem{jiang2020sdfdiff}
Jiang, Y., Ji, D., Han, Z., Zwicker, M.: Sdfdiff: Differentiable rendering of
  signed distance fields for 3d shape optimization. In: Proceedings of the
  IEEE/CVF conference on computer vision and pattern recognition. pp.
  1251--1261 (2020)

\bibitem{kato2018neural}
Kato, H., Ushiku, Y., Harada, T.: Neural 3d mesh renderer. In: Proceedings of
  the IEEE conference on computer vision and pattern recognition. pp.
  3907--3916 (2018)

\bibitem{kerbl20233d}
Kerbl, B., Kopanas, G., Leimk{\"u}hler, T., Drettakis, G.: 3d gaussian
  splatting for real-time radiance field rendering. ACM Transactions on
  Graphics (ToG)  \textbf{42}(4),  1--14 (2023)

\bibitem{deLaGorce:2011:MHP:2006854.2007005}
de~La~Gorce, M., Fleet, D.J., Paragios, N.: Model-based 3d hand pose estimation
  from monocular video. IEEE Trans. Pattern Anal. Mach. Intell.
  \textbf{33}(9),  1793--1805 (Sep 2011). \doi{10.1109/TPAMI.2011.33},
  \url{http://dx.doi.org/10.1109/TPAMI.2011.33}

\bibitem{laine2020modular}
Laine, S., Hellsten, J., Karras, T., Seol, Y., Lehtinen, J., Aila, T.: Modular
  primitives for high-performance differentiable rendering. ACM Transactions on
  Graphics (TOG)  \textbf{39}(6),  1--14 (2020)

\bibitem{Li:2018:DMC}
Li, T.M., Aittala, M., Durand, F., Lehtinen, J.: Differentiable monte carlo ray
  tracing through edge sampling. ACM Trans. Graph. (Proc. SIGGRAPH Asia)
  \textbf{37}(6),  222:1--222:11 (2018)

\bibitem{lin2023magic3d}
Lin, C.H., Gao, J., Tang, L., Takikawa, T., Zeng, X., Huang, X., Kreis, K.,
  Fidler, S., Liu, M.Y., Lin, T.Y.: Magic3d: High-resolution text-to-3d content
  creation. In: Proceedings of the IEEE/CVF Conference on Computer Vision and
  Pattern Recognition. pp. 300--309 (2023)

\bibitem{liu2020neural}
Liu, L., Gu, J., Zaw~Lin, K., Chua, T.S., Theobalt, C.: Neural sparse voxel
  fields. Advances in Neural Information Processing Systems  \textbf{33},
  15651--15663 (2020)

\bibitem{liu2019softras}
Liu, S., Li, T., Chen, W., Li, H.: Soft rasterizer: A differentiable renderer
  for image-based 3d reasoning. The IEEE International Conference on Computer
  Vision (ICCV)  (Oct 2019)

\bibitem{lombardi2018deep}
Lombardi, S., Saragih, J., Simon, T., Sheikh, Y.: Deep appearance models for
  face rendering. ACM Transactions on Graphics (TOG)  \textbf{37}(4),  1--13
  (2018)

\bibitem{lombardi2019neural}
Lombardi, S., Simon, T., Saragih, J., Schwartz, G., Lehrmann, A., Sheikh, Y.:
  Neural volumes: Learning dynamic renderable volumes from images. arXiv
  preprint arXiv:1906.07751  (2019)

\bibitem{loper2014opendr}
Loper, M.M., Black, M.J.: Opendr: An approximate differentiable renderer. In:
  Computer Vision--ECCV 2014: 13th European Conference, Zurich, Switzerland,
  September 6-12, 2014, Proceedings, Part VII 13. pp. 154--169. Springer (2014)

\bibitem{Loubet2019Reparameterizing}
Loubet, G., Holzschuch, N., Jakob, W.: Reparameterizing discontinuous
  integrands for differentiable rendering. Transactions on Graphics
  (Proceedings of SIGGRAPH Asia)  \textbf{38}(6) (Dec 2019).
  \doi{10.1145/3355089.3356510}

\bibitem{Ma2021PixelCA}
Ma, S., Simon, T., Saragih, J.M., Wang, D., Li, Y., la~Torre, F.D., Sheikh, Y.:
  Pixel codec avatars. 2021 IEEE/CVF Conference on Computer Vision and Pattern
  Recognition (CVPR) pp. 64--73 (2021)

\bibitem{medina2022speechtongue}
Medina, S., Tomé, D., Stoll, C., Tiede, M., Munhall, K., Hauptmann, A.,
  Matthews, I.: Speech driven tongue animation. In: Proceedings of the IEEE/CVF
  Conference on Computer Vision and Pattern Recognition (CVPR). IEEE/CVF (2022)

\bibitem{mildenhall2020nerf}
Mildenhall, B., Srinivasan, P.P., Tancik, M., Barron, J.T., Ramamoorthi, R.,
  Ng, R.: Nerf: Representing scenes as neural radiance fields for view
  synthesis. In: ECCV (2020)

\bibitem{mildenhall2021nerf}
Mildenhall, B., Srinivasan, P.P., Tancik, M., Barron, J.T., Ramamoorthi, R.,
  Ng, R.: Nerf: Representing scenes as neural radiance fields for view
  synthesis. Communications of the ACM  \textbf{65}(1),  99--106 (2021)

\bibitem{muller2022instant}
M{\"u}ller, T., Evans, A., Schied, C., Keller, A.: Instant neural graphics
  primitives with a multiresolution hash encoding. arXiv preprint
  arXiv:2201.05989  (2022)

\bibitem{nicolet2021large}
Nicolet, B., Jacobson, A., Jakob, W.: Large steps in inverse rendering of
  geometry. ACM Transactions on Graphics (TOG)  \textbf{40}(6),  1--13 (2021)

\bibitem{niemeyer2020differentiable}
Niemeyer, M., Mescheder, L., Oechsle, M., Geiger, A.: Differentiable volumetric
  rendering: Learning implicit 3d representations without 3d supervision. In:
  Proceedings of the IEEE/CVF Conference on Computer Vision and Pattern
  Recognition. pp. 3504--3515 (2020)

\bibitem{NimierDavidVicini2019Mitsuba2}
Nimier-David, M., Vicini, D., Zeltner, T., Jakob, W.: Mitsuba 2: A retargetable
  forward and inverse renderer. Transactions on Graphics (Proceedings of
  SIGGRAPH Asia)  \textbf{38}(6) (Dec 2019). \doi{10.1145/3355089.3356498}

\bibitem{park2019deepsdf}
Park, J.J., Florence, P., Straub, J., Newcombe, R., Lovegrove, S.: Deepsdf:
  Learning continuous signed distance functions for shape representation. In:
  Proceedings of the IEEE/CVF conference on computer vision and pattern
  recognition. pp. 165--174 (2019)

\bibitem{pavlakos2018learning}
Pavlakos, G., Zhu, L., Zhou, X., Daniilidis, K.: Learning to estimate 3d human
  pose and shape from a single color image. In: Proceedings of the IEEE
  conference on computer vision and pattern recognition. pp. 459--468 (2018)

\bibitem{poole2022dreamfusion}
Poole, B., Jain, A., Barron, J.T., Mildenhall, B.: Dreamfusion: Text-to-3d
  using 2d diffusion. arXiv preprint arXiv:2209.14988  (2022)

\bibitem{rhodin2015versatile}
Rhodin, H., Robertini, N., Richardt, C., Seidel, H.P., Theobalt, C.: A
  versatile scene model with differentiable visibility applied to generative
  pose estimation. In: Proceedings of the IEEE International Conference on
  Computer Vision. pp. 765--773 (2015)

\bibitem{roessle2022dense}
Roessle, B., Barron, J.T., Mildenhall, B., Srinivasan, P.P., Nie{\ss}ner, M.:
  Dense depth priors for neural radiance fields from sparse input views. In:
  Proceedings of the IEEE/CVF Conference on Computer Vision and Pattern
  Recognition. pp. 12892--12901 (2022)

\bibitem{roveri2018pointpronets}
Roveri, R., {\"O}ztireli, A.C., Pandele, I., Gross, M.: Pointpronets:
  Consolidation of point clouds with convolutional neural networks. In:
  Computer Graphics Forum. vol.~37, pp. 87--99. Wiley Online Library (2018)

\bibitem{shi2023mvdream}
Shi, Y., Wang, P., Ye, J., Long, M., Li, K., Yang, X.: Mvdream: Multi-view
  diffusion for 3d generation. arXiv preprint arXiv:2308.16512  (2023)

\bibitem{tsalicoglou2023textmesh}
Tsalicoglou, C., Manhardt, F., Tonioni, A., Niemeyer, M., Tombari, F.:
  Textmesh: Generation of realistic 3d meshes from text prompts. arXiv preprint
  arXiv:2304.12439  (2023)

\bibitem{tulsiani2017multi}
Tulsiani, S., Zhou, T., Efros, A.A., Malik, J.: Multi-view supervision for
  single-view reconstruction via differentiable ray consistency. In:
  Proceedings of the IEEE conference on computer vision and pattern
  recognition. pp. 2626--2634 (2017)

\bibitem{vicini2022differentiable}
Vicini, D., Speierer, S., Jakob, W.: Differentiable signed distance function
  rendering. ACM Transactions on Graphics (TOG)  \textbf{41}(4),  1--18 (2022)

\bibitem{wang2020deep}
Wang, J., Sun, K., Cheng, T., Jiang, B., Deng, C., Zhao, Y., Liu, D., Mu, Y.,
  Tan, M., Wang, X., et~al.: Deep high-resolution representation learning for
  visual recognition. IEEE transactions on pattern analysis and machine
  intelligence  \textbf{43}(10),  3349--3364 (2020)

\bibitem{wiles2020synsin}
Wiles, O., Gkioxari, G., Szeliski, R., Johnson, J.: Synsin: End-to-end view
  synthesis from a single image. In: Proceedings of the IEEE/CVF Conference on
  Computer Vision and Pattern Recognition. pp. 7467--7477 (2020)

\bibitem{wu2016modelteeth}
Wu, C., Bradley, D., Garrido, P., Zollh\"{o}fer, M., Theobalt, C., Gross, M.,
  Beeler, T.: Model-based teeth reconstruction. ACM Trans. Graph.
  \textbf{35}(6) (dec 2016)

\bibitem{yan2016perspective}
Yan, X., Yang, J., Yumer, E., Guo, Y., Lee, H.: Perspective transformer nets:
  Learning single-view 3d object reconstruction without 3d supervision.
  Advances in neural information processing systems  \textbf{29} (2016)

\bibitem{yang2018learning}
Yang, G., Cui, Y., Belongie, S., Hariharan, B.: Learning single-view 3d
  reconstruction with limited pose supervision. In: Proceedings of the European
  Conference on Computer Vision (ECCV). pp. 86--101 (2018)

\bibitem{yu2021plenoxels}
Yu, A., Fridovich-Keil, S., Tancik, M., Chen, Q., Recht, B., Kanazawa, A.:
  Plenoxels: Radiance fields without neural networks (2021)

\bibitem{Zhang2019DTRT}
Zhang, C., Wu, L., Zheng, C., Gkioulekas, I., Ramamoorthi, R., Zhao, S.: A
  differential theory of radiative transfer. ACM Trans. Graph.  \textbf{38}(6),
   227:1--227:16 (2019)

\bibitem{zhang2020nerf++}
Zhang, K., Riegler, G., Snavely, N., Koltun, V.: Nerf++: Analyzing and
  improving neural radiance fields. arXiv preprint arXiv:2010.07492  (2020)

\bibitem{Zhao2020DifferentiableCourse}
Zhao, S., Jakob, W., Li, T.M.: Physics-based differentiable rendering: From
  theory to implementation. In: ACM SIGGRAPH 2020 Courses. SIGGRAPH '20,
  Association for Computing Machinery, New York, NY, USA (2020).
  \doi{10.1145/3388769.3407454}, \url{https://doi.org/10.1145/3388769.3407454}

\end{thebibliography}

\clearpage

\appendix
\setcounter{section}{-1}

{
\centering
\Large
\textbf{Rasterized Edge Gradients: Handling Discontinuities Differentiably} \\
\vspace{0.5em}Supplementary Material \\
\vspace{1.0em}
\appendix
}


\setcounter{page}{1}

%
%
%

\section{Additional details}\label{sec:additional-details}
\setcounter{subsection}{-1}
\label{sec:supplementary}

In this supplementary material, we provide further insights, detailed derivations, and additional qualitative results to complement our work.
We start by discussing the rationale behind employing one-sided derivatives to address C1 discontinuities in derivatives, initially introduced in ~\S\ref{subsec:pixel_pairs} of the main paper, detailed here in ~\S\ref{subsec:sup_pixel_pairs}.
We then explain in more detail edge classification process and gradient scattering, introduced in \S\ref{subsec:edge_classification} of the main paper and further elaborated here in \S\ref{subsec:sup_edge_class_splatting_detail}.
Following that, we offer a detailed derivation of Eqn.\eqref{eq-grad_int} introduced \S\ref{subsec:geom_intersections}, in \S\ref{subsec:sup_geom_intersections_detailed} of this supplement.
Lastly, we provide additional qualitative evaluations including toy example, Blender dataset from Mildenhall {\em et al.}~\cite{mildenhall2021nerf} and demonstrate our method's application to dynamic head scene mesh fitting in \S\ref{sec:sup_results}.

\subsection{Pixel pairs}
\label{subsec:sup_pixel_pairs}
As discussed in Section~\ref{subsec:pixel_pairs}, placing an edge precisely at the boundary between two pixels complicates matters due to C1 discontinuity.
Before addressing these complexities, let's first revisit Eqn.~\eqref{eq-pixel_pair} and further expand on the derivations:

\begin{equation}
  \begin{aligned}
  I_A &= \iint\limits_{D_A}^{} \theta(\alpha(x,y)) f_a(x,y) + \theta(-\alpha(x,y)) f_b(x,y) \,dx\,dy \\
  I_B &= \iint\limits_{D_B}^{} \theta(-\alpha(x,y)) f_a(x,y)  + \theta(\alpha(x,y)) f_b(x,y) \,dx\,dy,
    \end{aligned}
  \tag{7}\label{eq:sup_pixel_pair}
\end{equation}

where $I_A$ and $I_B$ are intensities of two adjacent pixels, A and B, separated by an edge defined by $\alpha(x,y)$.
Without loss of generality, we consider a horizontal pair, thus we define $\alpha(x,y) = p_{AB} - x$, where $p_{AB}$ is the edge location:

\begin{equation}
  \begin{aligned}
  I_A &= \iint\limits_{D_A}^{} \theta(p_{AB} - x) f_a(x,y) + \theta(x - p_{AB}) f_b(x,y) \,dx\,dy \\
  I_B &= \iint\limits_{D_B}^{} \theta(x - p_{AB}) f_a(x,y)  + \theta(p_{AB} - x) f_b(x,y) \,dx\,dy,
    \end{aligned}
  \label{eq:sup_pixel_pair_2}
\end{equation}

As was mentioned in Section~\ref{subsec:pixel_pairs}, edge location $p_{AB}$ is now the scene parameter with respect to which we want to differentiate the loss function, which in turn would allow us to optimize the edge location $p_{AB}$.
Since $p_{AB}$ effects both $I_A$ and $I_B$:
\begin{equation}
  \frac{\partial L}{\partial p_{AB}} = \frac{\partial L}{\partial I_A} \frac{\partial I_A}{\partial p_{AB}} +
  \frac{\partial L}{\partial I_B} \frac{\partial I_B}{\partial p_{AB}},
  \tag{8}\label{eq-pixel_pair_loss_2}
\end{equation}

where $\frac{\partial L}{\partial I_B}$ and $\frac{\partial L}{\partial p_{AB}}$ are the known gradients, typically obtained by backpropagation the loss through the loss function, while $\frac{\partial I_B}{\partial p_{AB}}$ and $\frac{\partial I_A}{\partial p_{AB}}$ are the unknown derivatives that we want to compute.

Differentiating Eqn.~\eqref{eq:sup_pixel_pair_2} by $p_{AB}$ and applying chain rule we obtain:

\begin{equation}
  \begin{aligned}
  \frac{\partial I_A}{\partial p_{AB}} &= \overbrace{\iint\limits_{D_A}^{} \frac{\partial \theta(p_{AB} - x)}{\partial p_{AB}} f_a(x,y) + \frac{\partial \theta(x - p_{AB})}{\partial p_{AB}} f_b(x,y) \,dx\,dy}^\text{boundary component} \\
  &+ \overbrace{\iint\limits_{D_A}^{} \theta(p_{AB} - x) \frac{\partial f_a(x,y)}{\partial p_{AB}} + \theta(x - p_{AB}) \frac{\partial f_b(x,y)}{\partial p_{AB}} \,dx\,dy}^\text{smooth component} \\
  \frac{\partial I_B}{\partial p_{AB}} &= \overbrace{\iint\limits_{D_B}^{} \frac{\partial \theta(x - p_{AB})}{\partial p_{AB}} f_a(x,y)  + \frac{\partial \theta(p_{AB} - x)}{\partial p_{AB}} f_b(x,y) \,dx\,dy}^\text{boundary component} \\
      &+ \overbrace{\iint\limits_{D_B}^{} \theta(x - p_{AB}) \frac{\partial f_a(x,y)}{\partial p_{AB}}  + \theta(p_{AB} - x)\frac{\partial  f_b(x,y)}{\partial p_{AB}} \,dx\,dy.}^\text{smooth component}
    \end{aligned}
  \label{eq:sup_pixel_pair_3}
\end{equation}

As discussed in Section~\ref{subsec:pixel_pairs}, after applying the chain rule, the integral separates into two.
The second integral in both equations in of Eqn.~\ref{eq:sup_pixel_pair_3} is an integral of the derivative of the smooth function, $f_a(x,y)$.
This derivative is readily computable with AD, thus we ignore it for brevity.
We replace it with $\Omega$ symbol that shows that there is additional smooth component that being omitted.

Derivative of the Heaviside function is the Dirac delta function: $\delta(x) = \frac{d}{x}\theta(x)$.
Thus, we can rewrite Eqn.~\ref{eq:sup_pixel_pair_3} as:

\begin{equation}
  \begin{aligned}
  \frac{\partial I_A}{\partial p_{AB}} &= \iint\limits_{D_A}^{} \delta(p_{AB} - x) f_a(x,y) - \delta(x - p_{AB}) f_b(x,y) \,dx\,dy + \Omega_A\\
  \frac{\partial I_B}{\partial p_{AB}} &= \iint\limits_{D_B}^{} \delta(x - p_{AB}) f_a(x,y) - \delta(p_{AB} - x) f_b(x,y) \,dx\,dy + \Omega_B\\
    \end{aligned}
  \label{eq:sup_pixel_pair_4}
\end{equation}

Unfortunately, Eqn.~\ref{eq:sup_pixel_pair_4} has a problem preventing us from simply computing the integrals.
Regions $D_A$ and $D_B$ which correspond to pixels $A$ and $B$ are disjoint.
The boundary defined by $\delta(p_{AB} - x)$ lies between $D_A$ and $D_B$, thus it is not clear if we should include it to the integral for $\frac{\partial I_A}{\partial p_{AB}}$ or $\frac{\partial I_B}{\partial p_{AB}}$.
We can not include it to both.
This ambiguity is a direct consequence of $I_A$ and $I_B$ having a C1 discontinuity at $p_{AB} = 0$.

Indeed, if we take a closer look at Eqn.\ref{eq:sup_pixel_pair_2} then we will see that both $I_A$ and $I_B$ are piece-wise
linear with respect to $p_{AB}$.
That follows from the micro-edge formulation according to which pixels have constant value across all of their area (both $f_a$ and $f_b$ are constant within $D_A$ and $D_B$ respectively)
The breakpoint happens to be at the boundary of two pixels, at $p_{AB} = 0$, but that's also where we want to compute the derivative.
Because of this breakpoint, neither $I_A$ or $I_B$ are differentiable at $p_{AB} = 0$.
However, subderivative still exists, which we can introduce using one-sided limits:

\begin{equation}
  \begin{aligned}
  \frac{\partial I_A}{\partial p_{AB}}^- &= \lim _{x\to 0^{-}}{\frac {I_A(x)-I_A(0)}{x}}, \quad
  \frac{\partial I_A}{\partial p_{AB}}^+ &= \lim _{x\to 0^{+}}{\frac {I_A(x)-I_A(0)}{x}} \\
  \frac{\partial I_B}{\partial p_{AB}}^- &= \lim _{x\to 0^{-}}{\frac {I_B(x)-I_B(0)}{x}}, \quad
  \frac{\partial I_B}{\partial p_{AB}}^+ &= \lim _{x\to 0^{+}}{\frac {I_B(x)-I_B(0)}{x}},
    \end{aligned}
  \label{eq:sup_pixel_pair_5}
\end{equation}

where $x\to 0^{-}$ means that x is increasing and approaching $0$ from the left, and $x\to 0^{+}$ means that x is decreasing and approaching $0$ from the right.
In order to approximate $\frac{\partial I_A}{\partial p_{AB}}$ and $\frac{\partial I_B}{\partial p_{AB}}$ we can average derivatives from the left and from the right.
That is equivalent to assuming that the Dirac delta in Eqn.~\ref{eq:sup_pixel_pair_4} partially belongs to both, $D_A$ and $D_B$, which is exactly what we want.
Thus, averaging the subderivatives, and taking into account that $\int _{-\infty }^{\infty }\delta (x)\,dx=1$ we arrive at:

\begin{equation}
  \begin{aligned}
  \frac{\partial I_A}{\partial p_{AB}} &= \frac{1}{2} \int\limits_{D_A \cup x=0}^{} \left [ f_a(x,y) - f_b(x,y) \right ] \,dy + \Omega_A\\
  \frac{\partial I_B}{\partial p_{AB}} &= \frac{1}{2} \int\limits_{D_B \cup x=0}^{} \left [ f_a(x,y) - f_b(x,y) \right ] \,dy + \Omega_B.
    \end{aligned}
  \label{eq:sup_pixel_pair_6}
\end{equation}

Using the mentioned property of micro-edge formulation, that both $f_a$ and $f_b$ are constant within $D_A$ and $D_B$ respectively:

\begin{equation}
  \begin{aligned}
  \frac{\partial I_A}{\partial p_{AB}} &= \frac{1}{2}  \left ( I_A - I_B \right ) + \Omega_A\\
  \frac{\partial I_B}{\partial p_{AB}} &= \frac{1}{2} \left ( I_A - I_B \right ) + \Omega_B.
    \end{aligned}
  \label{eq:sup_pixel_pair_7}
\end{equation}

Now we can plug this result into Eqn.~\eqref{eq-pixel_pair_loss_2} and get:

\begin{equation}
  \boxed{\frac{\partial L }{\partial p_{AB}} =  \frac{1}{2} \left ( \frac{\partial L}{\partial I_A} + \frac{\partial L}{\partial I_B} \right )   \left (I_A - I_B \right ) + \Omega},
  \tag{11}\label{eq-sup_chain_new}
\end{equation}

Notably, we can arrive to the same result from slightly different perspective.
Instead of averaging subderivatives, we could circumvent the C1 discontinuity by considering the pixel pair as a single unit with average pixel value $I_{AB} = \frac{I_A + I_B}{2}$, thus eliminating the discontinuity problem:
\begin{equation}
  \begin{aligned}
  I_{AB} &= \frac{1}{2} \iint_{D_{AB}}^{} \left [ \theta(\alpha(x,y)) f_a(x,y) + \theta(-\alpha(x,y)) f_b(x,y) \right ] \,dx\,dy.
    \end{aligned}
  \label{eq-sup_pixel_pair_unit}
\end{equation}

Similarly, we assume a horizontal pair: $\alpha(x,y) = p_{AB} - x$:

\begin{equation}
  \begin{aligned}
  I_{AB} &= \frac{1}{2} \iint_{D_{AB}}^{} \left [ \theta(p_{AB} - x) f_a(x,y) + \theta(x - p_{AB}) f_b(x,y) \right ] \,dx\,dy.
    \end{aligned}
  \label{eq-sup_pixel_pair_unit_2}
\end{equation}

Then the derivative with respect to the edge location $p_{AB}$:

\begin{equation}
  \begin{aligned}
 \frac{\partial I_{AB}}{\partial p_{AB}}  &= \frac{1}{2} \iint_{D_{AB}}^{} \left [ \frac{\partial  \theta(p_{AB} - x) }{\partial p_{AB}} f_a(x,y) + \frac{\partial   \theta(x - p_{AB}) }{\partial p_{AB}} f_b(x,y) \right ] \,dx\,dy + \Omega.
    \end{aligned}
  \label{eq-sup_pixel_pair_unit_3}
\end{equation}

Using identity $\delta(x) = \frac{d}{x}\theta(x)$:

\begin{equation}
  \begin{aligned}
 \frac{\partial I_{AB}}{\partial p_{AB}}  &= \frac{1}{2} \iint_{D_{AB}}^{} \left [ \delta(p_{AB} - x)f_a(x,y)-  \delta(x - p_{AB}) f_b(x,y) \right ] \,dx\,dy + \Omega.
    \end{aligned}
  \label{eq-sup_pixel_pair_unit_4}
\end{equation}

Taking into account that $\int _{-\infty }^{\infty }\delta (x)\,dx=1$ and combining into one integral we arrive at:

\begin{equation}
  \begin{aligned}
 \frac{\partial I_{AB}}{\partial p_{AB}}  &= \frac{1}{2} \int\limits_{D_A \cup x=0}^{}  \left [ f_a(x,y) - f_b(x,y) \right ] \,dy + \Omega.
    \end{aligned}
  \label{eq-sup_pixel_pair_unit_5}
\end{equation}

Similarly, taking into account that $f_a$ and $f_b$ are constant within $D_A$ and $D_B$ respectively:

\begin{equation}
  \begin{aligned}
       \frac{\partial I_{AB}}{\partial p} = \frac{1}{2} \left ( I_B - I_A \right ).
    \end{aligned}
  \label{eq-pixel_pair_diff_integrated}
\end{equation}

Now that we know the derivative of the average intensity, we need to apply it to Eqn.~\eqref{eq-pixel_pair_loss_2}.
Since we only know average intensity, we assume that individual derivatives $\frac{\partial I_A}{\partial p_{AB}}$ and $\frac{\partial I_B}{\partial p_{AB}}$ are close enough to the average  $\frac{\partial I_{AB}}{\partial p}$, thus:

\begin{equation}
  \frac{\partial L}{\partial p_{AB}} = \frac{\partial L}{\partial I_A} \frac{\partial I_A}{\partial p_{AB}} +
  \frac{\partial L}{\partial I_B} \frac{\partial I_B}{\partial p_{AB}} \approx
  \left( \frac{\partial L}{\partial I_A}  +
  \frac{\partial L}{\partial I_B} \right ) \frac{\partial I_{AB}}{\partial p_{AB}}.
  \tag{8}\label{eq-pixel_pair_loss_22}
\end{equation}

And finally, if we plug Eqn.~\eqref{eq-pixel_pair_diff_integrated} into Eqn.~\eqref{eq-pixel_pair_loss_22}:

\begin{equation}
  \boxed{\frac{\partial L }{\partial p} =  \frac{1}{2} \left ( \frac{\partial L}{\partial I_A} + \frac{\partial L}{\partial I_B} \right )   \left (I_B - I_A \right )},
  \label{eq-chain_new_2}
\end{equation}

which coincides with our previous result and resolves the ambiguity of the Dirac delta function integration.
In summary, our approximation method allows to circumvent the non-existent derivative with the average of two one-sided limits, which leads to the solution described in Section~\ref{subsec:pixel_pairs}.

\subsection{Edge classification and gradient scattering}
\label{subsec:sup_edge_class_splatting_detail}

We have devised a method to compute the gradient with respect to the edge location, $\frac{\partial L}{\partial p} $.
However, our ultimate goal is to derive the gradient concerning vertex positions, which presents a different challenge.
 As outlined in Section~\ref{subsec:edge_classification}, we utilize Automatic Differentiation (AD) and primarily focus on gradients relative to fragment locations.
These gradients are then back-propagated via AD to the vertex positions.

It's important to note that gradients with respect to fragment positions are distinct from gradients with respect to edge position.
For the later one, we have calculated
$\frac{\partial L}{\partial p} $ for vertical and horizontal edges, reside between pixels, hence there is no direct mapping from edges to fragments.

The computation process is more intuitive when approached from an edge-centric perspective, see Figure~\ref{fig:gradient_flow} of the main paper.
We iterate over every edge in the image.
For each edge, we take the average of the incoming gradient from the pixels adjacent to the edge, then we multiply it with the difference of the pixel pair intensity, and then scatter this computed gradient onto the gradient image relative to fragment locations.

Before scattering, we transform our edge-related gradients $\frac{\partial L}{\partial p} $ into fragment-related gradients $\frac{\partial L}{\partial r}$ by multiplying with $\frac{\partial p}{\partial r} $.
These derivatives of edge location with respect to the fragment locations $\frac{\partial p}{\partial r} $ greatly dependent on the edge type, as detailed in Figure~\ref{fig:edge_classification} of the main paper.

In scenarios without an edge, or where the edge doesn't create a discontinuity (adjacent primitives), the value of $\frac{\partial p}{\partial r}$ is zero, so no gradient is scattered.
The simplest case for gradient propagation is the overhanging scenario, occurring when a front primitive overlaps another or the background.
Since for the background primitive's infinitesimal movement doesn't affect the edge position, its partial derivative is zero.
Conversely, for the foreground primitive, this derivative is $1$ since the infinitesimal translation of the foreground primitive directly moves the edge with the same direction and velocity.

The most complex case is when the edge is formed with self-intersecting geometry.
In this scenario, the fragment translation does not match the translation of the edge and in fact, they can even have opposite directions.
Therefore, the partial derivative $\frac{\partial p}{\partial r}$ might have positive or negative values for both primitives.
Since both primitives contribute to the edge displacement, we use the product rule for differentiation and first fix one primitive and differentiate with respect to the movement of the other, and then do it vice versa.
For detailed computations of these partial derivatives, refer to \S\ref{subsec:sup_geom_intersections_detailed}.


\subsection{Geometry intersections}
\label{subsec:sup_geom_intersections_detailed}

In this subsection we explore the computation of the derivatives of edge location with respect to fragment locations $\frac{\partial p}{\partial r} $ when the edge is formed by intersecting geometry rather than from the edges of primitives.

This situation is inherently complex, and a general-case solution would be challenging.
However, our specific simplifications, namely, the assumption that all edges are either vertical or horizontal, significantly streamline the problem.
In our framework, intersections are only feasible between two vertical or two horizontal edges.
Consequently, we can reduce our analysis from 3D to 2D by focusing on either the x-z or y-z planes, where the z-axis is perpendicular to the image plane and the x and y axes lie within it.

Without loss of generality, we will only consider the x-z plane; the same logic applies to the y-z plane.
Recall that the x-z plane is the plane for vertical edges and the y-z plane is the plane for horizontal edges.

Please see Figure~\ref{fig:intersection} of the main paper while following the derivation below.

We note two observations: translation of the primitives along the primitive plane does not cause the edge to move.
Local rotation (around the edge) of the primitive does not move the edge either.
So in order to move the edge we need to translate the primitive along its normal.

As we mentioned previously, we first consider one primitive fixed and vary the position of the other, and then switch the primitives roles.

For brevity, we will only consider one primitive fixed and the other varying.

Let the normal of the fixed primitive be $\mathbf{n^f}$ and the normal of the varying primitive $\mathbf{n^v}$.
We compute a unit length, in-plane vector of the fixed primitive, $\mathbf{b}$,  by rotating the primitive's normal by \ang{90}:

\begin{equation}
 \mathbf{b} = \begin{bmatrix} 0 & -1 \\ 1 & 0 \end{bmatrix} \mathbf{n}^f.
  \label{eq-b_vec}
\end{equation}

The translation of the varying primitive is $\partial \mathbf{r}$.
Let the $\partial r$ be the projection of $\partial \mathbf{r}$ onto the normal of the varying primitive:

\begin{equation}
 {\partial \mathbf{r}} = {\partial r} \mathbf{n}^v.
  \label{eq-r}
\end{equation}

Let $\mathbf{s}$ be the displacement vector of the edge in 3D. It is obvious, that the edge can only move along the line of the fixed primitive, thus $\mathbf{s}$ and $\mathbf{b}$ are collinear.
Moreover, ${\partial \mathbf{r}}$ is the projection of $\mathbf{s}$ onto $\mathbf{n}^v$:

\begin{equation}
 {\partial \mathbf{r}} = \left ( \mathbf{s} \cdot \mathbf{n}^v \right ) \mathbf{n}^v.
  \label{eq-projection}
\end{equation}

Thus:

\begin{equation}
\mathbf{s} = \frac{\mathbf{b}}{\mathbf{b} \cdot \mathbf{n}^v } {\partial r}
  \label{eq-s}
\end{equation}

The change in the edge position, $\partial p$ is projection of $\mathbf{s}$ onto the z axis: ${\partial p} = \mathbf{s} \cdot \mathbf{e}_x$.
So, we can rewrite the~\eqref{eq-s} as:

\begin{equation}
\frac{\partial p}{\partial r} = \frac{\mathbf{b}}{\mathbf{b} \cdot \mathbf{n}^v } \cdot \mathbf{e}_x,
  \label{eq-prs}
\end{equation}

or by substituting the~\eqref{eq-b_vec}

\begin{equation}
     \frac{\partial p}{\partial r} = \frac{-\mathbf{n}^f_z }{{\mathbf{n}^f}^T \begin{bmatrix} 0 & 1 \\ -1 & 0 \end{bmatrix} \mathbf{n}^v },
  \label{eq-prd}
\end{equation}

which coincides with the equation introduced in~\S\ref{subsec:geom_intersections} of the main paper.

\section{Implementation detail}
\label{sec:sup_implementation_details}

We implement our method as a collection of CUDA extension modules for PyTorch, and largely follow the structure of nvdiffrast~\cite{laine2020modular}, with the exception for the antialiasing module which we replace with our EdgeGrad module.

We implement the following modules:
\begin{enumerate}
  \item Vertex transform module.
  \item Rasterization module.
  \item Barycentrics module.
  \item Interpolation module.
  \item EdgeGrad module.
\end{enumerate}

\begin{figure}[t!]
  \begin{minipage}{1.0\textwidth}
    \centering
    \includegraphics[width=1.0\columnwidth]{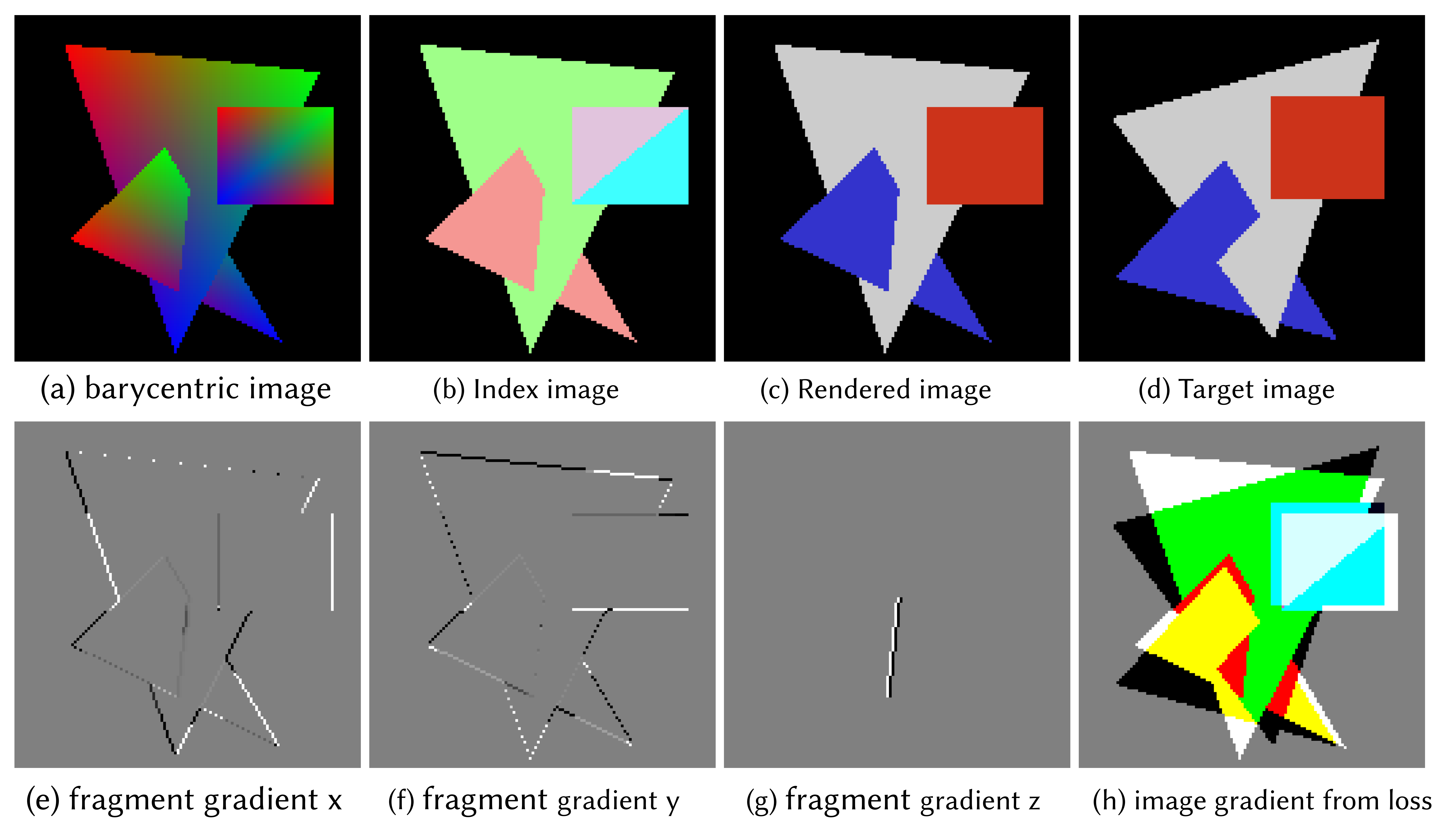}
    \caption{\textbf{A toy example}. The figure shows different intermidiate tensors demonstrating how our method works: a) barycentric image; b) index image; c) rendered image, here it was obtained by interpolating flat vertex colors; d) target image, $l2$ loss is computed between the target and rendered images; e) fragment gradients in $x$ direction, $\frac{\partial L } {\partial  \mathbf{r}_x}$
    ; f) fragment gradients in $y$ direction, $\frac{\partial L } {\partial  \mathbf{r}_y}$
    ; g) fragment gradients in $z$ direction, $\frac{\partial L } {\partial  \mathbf{r}_z}$
    ; h) image gradient from the loss, $\frac{\partial L } {\partial I_{i, j}}$
    }.
    \label{fig:toy_example}
  \end{minipage}\vspace{-1.5em}
\end{figure}

\begin{figure}[t]
  \begin{minipage}{1.0\textwidth}
    \centering
    \includegraphics[width=1.0\columnwidth]{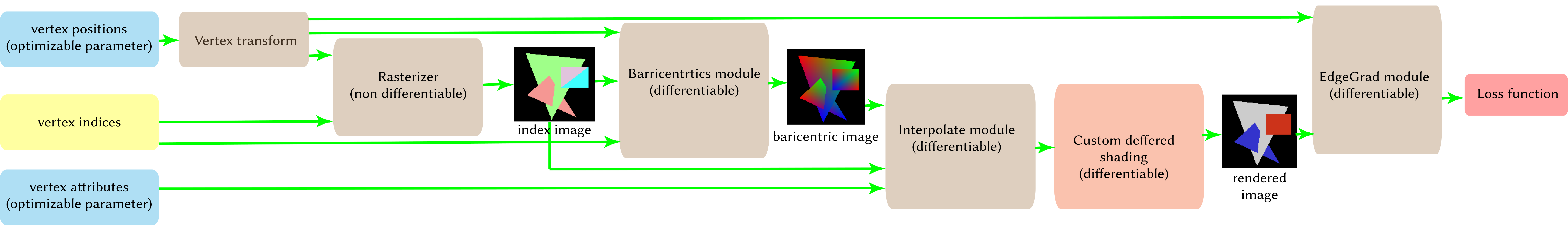}
    \caption{\textbf{Components}. The figure shows components of our differentiable rasterization system.}
    \label{fig:drtk}
  \end{minipage}\vspace{-1.5em}
\end{figure}

We implement module for vertex transformation, which transforms the vertices to the camera frame and then projects them onto the image plane.

Our rasterization module, which rasterizes the triangles to pixels and simultaneously identifies the closest primitive using z-buffer.
The output of the rasterization module are \textit{index image} and \textit{depth image}.
Index image records for each pixel the triangle ID that was rasterized to it.
Depth image records the distance to the fragment along the principal axis.

We implement barycentrics module which takes in geometry data and the index image and returns barycentrics image.

Our interpolation module interpolates vertex attributes within the rasterized fragments.
It takes vertex attributes, barycentrics image and the index image and returns images of the interpolated attributes.

Vertex transform, Barycentrics, and Interpolation modules implement backward pass, and the backward pass is a straightforward differentiation of the forward pass.
Rasterization module does not have a backward pass since that role is taken by EdgeGrad module.

In order to make the best use of the AD system, we implement EdgeGrad module as a module with both forward and backward passes.
Forward pass takes in rendered image and vertex positions.
However, the forward pass is not doing anything, we pass the rendered image through it unchanged.
The purpose of it is that the backward pass will be called by AD during the backpropagation and will receive gradients from the loss function: $\frac{\partial L}{\partial I_{i,j}}$ and will also return gradients for the vertex positions.
It is the backward pass where all computation happens, unlike the forward pass which is the identity operation.
Please see Fig.~\ref{fig:toy_example} for more details.

Given the modular structure, we can e.g.\ interpolate vertex colors, uv coordinates, etc.
After that, we can do deferred shading using various methods of our choice.
We can render not only rgb image, but also different axillary images, such as masks, segmentations, normals.
All of them can be passed though the EdgeGrad module to enable propagation of gradients from the boundary discontinuities.

For all experiments we utilize Laplacian preconditioning proposed by Nicolet {\em et al.}~\cite{nicolet2021large}.
We find Laplacian preconditioning extremely helpful for stabilizing training and we use it for both, meshes and texture data.
We use $\lambda = 8 .. 32$, most of the time $\lambda = 16$.

In all baselines, we use $l2$ loss as a photometric loss.
For all experiments on blender datasets, we add Laplacian regularization loss with a very small weight of $4e-08$, which is necessary because Laplacian preconditioning still allows accumulation of small artifacts over time.

Additionally, for all experiments we render an image $I_{triangle\_side}$ where each pixel is $1$ if the wrong side of triangle is visible and $0$ otherwise.
All meshes are watertight, thus the internal side of triangle is visible only if teh mesh has severe self-intersections.
Such rendered binary mask is not differentiable, but we pass it through EdgeGrad module which allows us to add a loss which we call \textit{triangle back face loss}: $L_{back\_face} = \|I_{triangle\_side}\|^2$. We use weight $10.0$ for $L_{back\_face}$ loss.

\subsection{Experiments on Blender~\cite{mildenhall2021nerf} dataset}
\label{sec:sup_implementation_details_blender}

For all experiments with the blender dataset we follow the geometry growing procedure proposed by Nicolet {\em et al.}~\cite{nicolet2021large}, except that we do not do re-meshing (for simplicity).
We start with manually created, rough, low polygon meshes with from $300$ to $1000$ vertices.
We also equip the initialization mesh with uv parametrization which is not optimized and stays the same.
We grow the mesh using loop subdivision during optimization up until it reaches 200k vertices.
The schedule has the following number of iteration per step $[1000, 2000, 4000, 8000]$, the last step has as many iterations as needed to achive the highest PSNR score on validation set.
We add view conditioned MLP for decoding texture values, which is similar to~\cite{Ma2021PixelCA}.
For each rendered pixel, we interpolate uv coordinates, and then in deferred shading step using the uv coordinates of the fragment, we sample from a neural texture.
The neural texture has 8 channels, and it has sizes $1024\times1024$.
The feature sampled from the neural texture is concatenated with the per-pixel view vector.
The resulting 11 channels feature vector is passed to the MLP which has in total 4 layers, with ReLU activations except for the last layer which just returns RGB value.
All intermediate layers have 32 channels.
We optimize vertex locations as well as the neural texture.
We use Adam optimizer with learning rate schedule $[0.0001, 0.001, 0.0005, 0.00005, 0.000025]$ for the geometry (we switch to the next learning rate each time we do geometry subdivision) and fixed learning rate of $0.002$ for neural texture and MLP.

\subsection{Application to dynamic head reconstruction}
\label{sec:sup_implementation_details_ca}

Please follow Ma  {\em et al.}~\cite{Ma2021PixelCA} for all details on how the avatar model is trained.
This model is based on encoder-decoder VAE architecture.
The encoder produces an expression code of length 256.
The decoder network has two branches, one of which produces
 $512\times512\times3$ displacement for geometry, and the other produces $512\times512\times4$ dynamic neural texture.
Only the neural texture branch is conditioned on a view vector additionally to the expression code.
Similar to the blender dataset experiments, there is also a static (not produced by a network) $1024\times1024\times4$ neural texture.
At deferred shading step, we sample both neural textures and concatenate results, thus obtaining 8-channel feature which is then pass to an MLP.
The MLP is applied separatly to each rendered pixel, it has 4 layers, with ReLU activations except for the last layer which just returns RGB value.
All intermediate layers have 8 channels.

We add supervision in the form of 2D segmentation masks with separate labels for the regions of the teeth, tongue, and lips, for details see Fig.~\ref{fig:details} of teh main paper.
The 2D segmentation masks are obtained from a detector network: HRNet~\cite{wang2020deep}, which is applied to the camera image, and it is trained using our in-house dataset.
Additionally, to the neural texture, we add segmentation texture which can do one-hot encoding of the label.
This segmentation texture is optimizable, and we initialize it to background label.
During training, we optimize $l2$ loss between the rendered segmentation texture and segmentation prediction from the segmentation detector.
In a such way, we learn content of the segmentation texture.
We also apply Laplacian preconditioning to the segmentation texture.

\section{Results}
\label{sec:sup_results}

We follow the experimental settings described for synthetic blender dataset in~\cite{mildenhall2020nerf}. For the mesh fitting, we use 100 training views and terminate the optimization when the highest score is achieved on the validation set of another 100 views. Subsequent evaluation employs the remaining 100 views of the test set. All images used in training, validation, and testing are 800 × 800 pixels. Our method, EdgeGrad, is compared against a variant without intersection handling and another without edge gradient handling at all. We also benchmark against nvdiffrastr, one of the methods most similar to ours.

The quantitative results are detailed in Table~\ref{table:suppresults2}.
As indicated, intersection handling has a nominal impact on PSNR and SSIM. However, it significantly enhances the LPIPS score. We observe that intersection handling effectively reduces topological artifacts. Although these artifacts are typically small, they significantly impact perceptual quality. We hypothesize that such artifacts minimally affect non-perceptual losses like PSNR and have limited impact on SSIM, but significantly deteriorate the LPIPS score.

\begin{figure*}
\centering\footnotesize
\begingroup
\renewcommand{\arraystretch}{0.}
\setlength{\tabcolsep}{0pt}
\begin{tabular}{ccc}
Initialization mesh & Nvdiffrast~\cite{laine2020modular} & EdgeGrad (our) \\
\midrule
\includegraphics[width=0.28\linewidth]{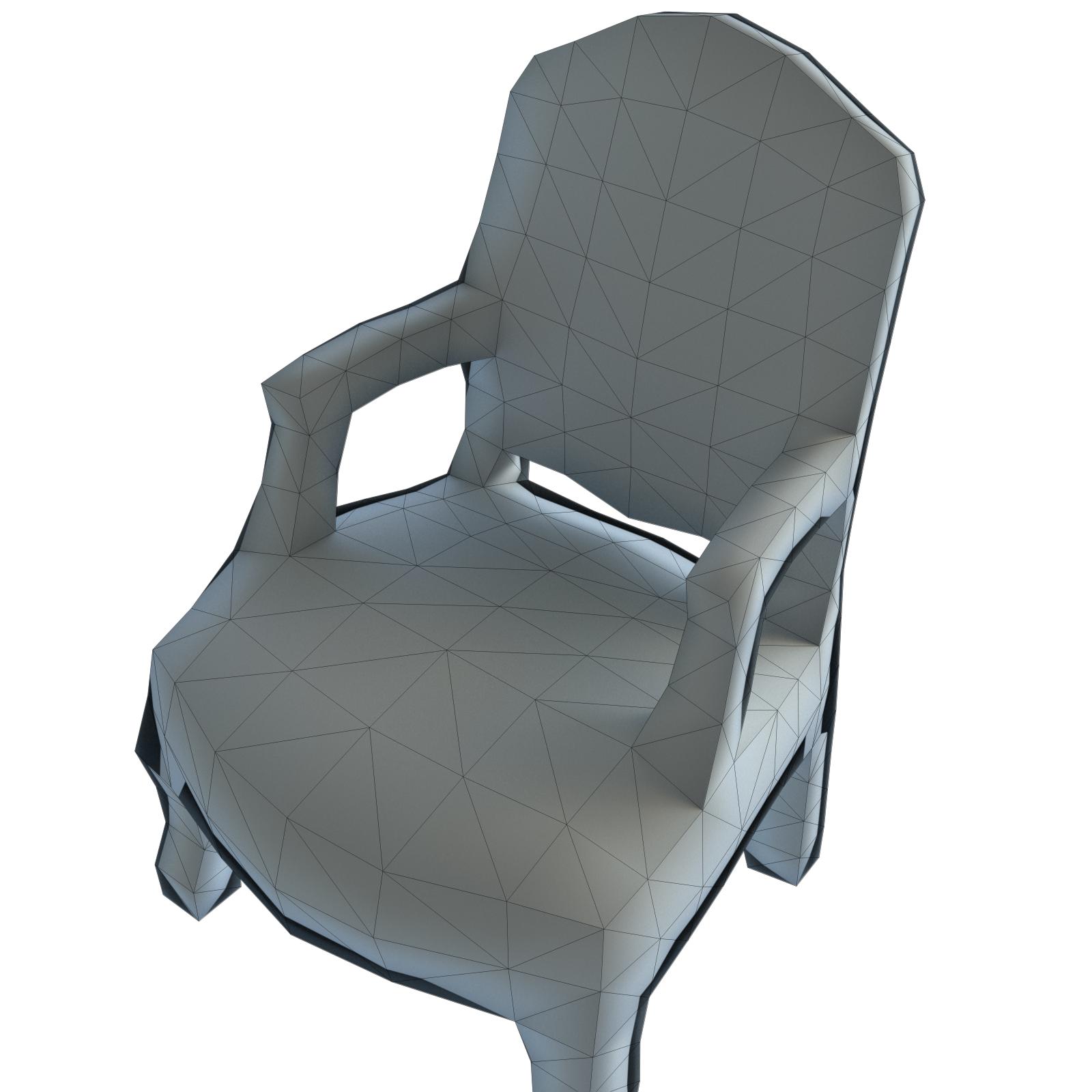} &
\includegraphics[width=0.28\textwidth]{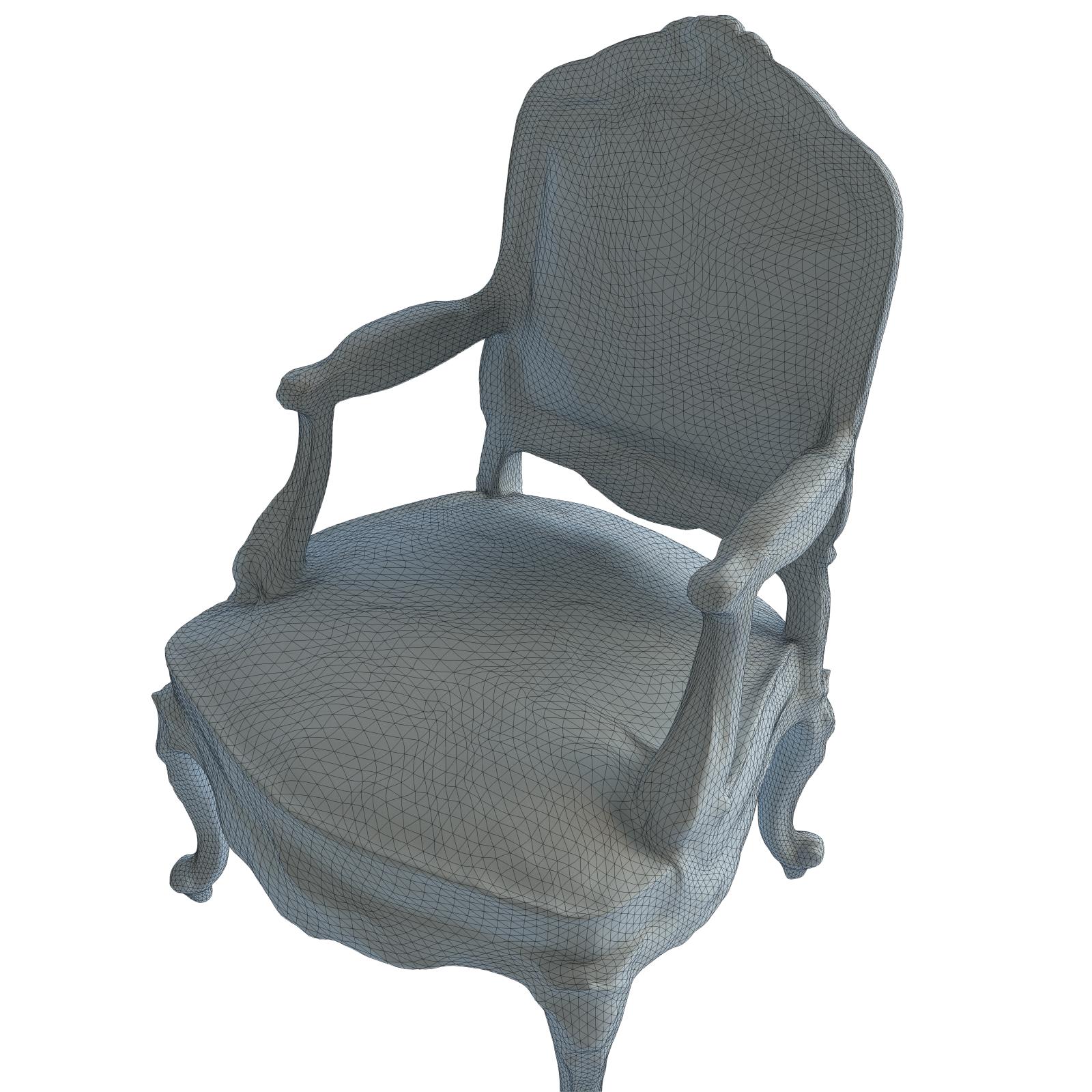} &
\includegraphics[width=0.28\textwidth]{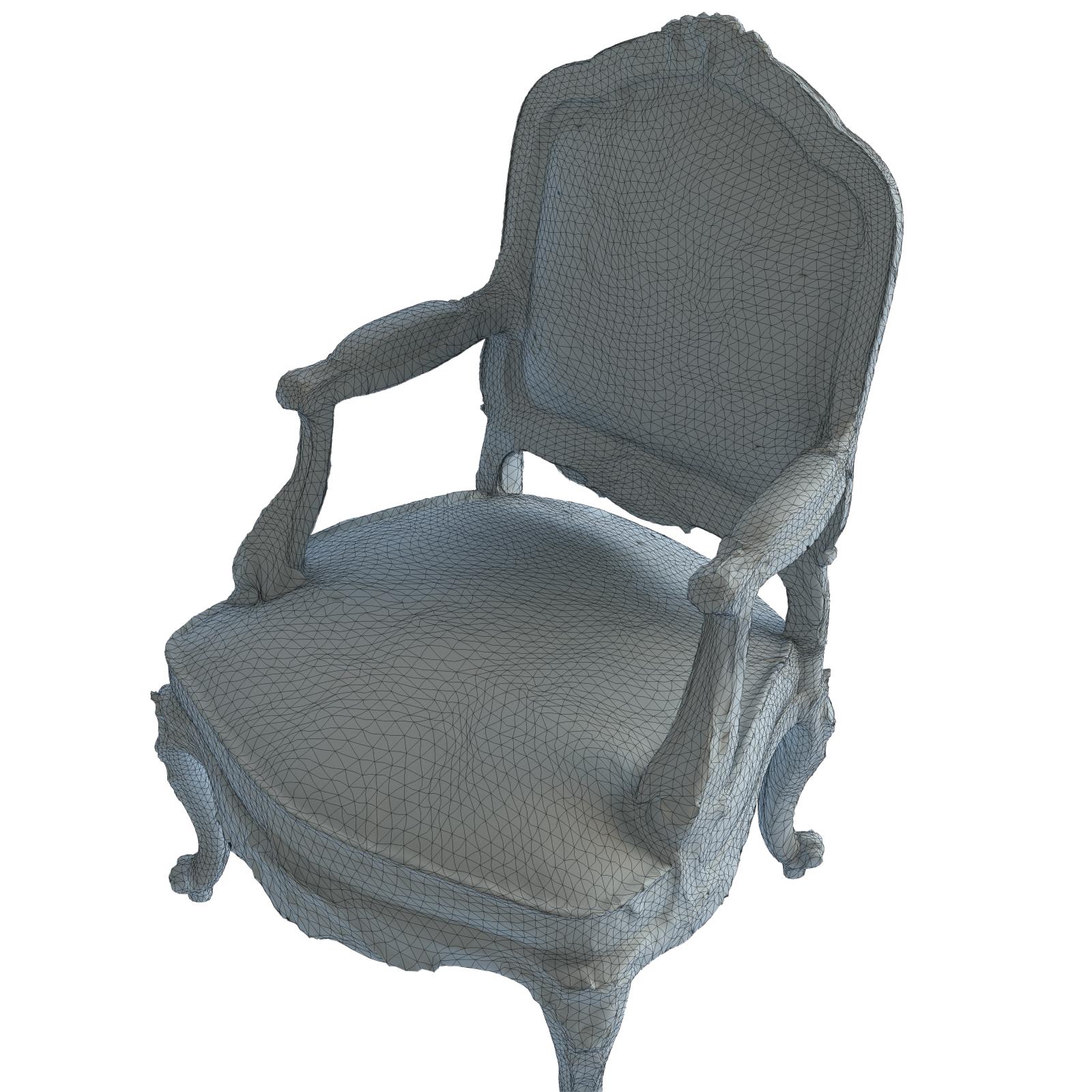} \\
\includegraphics[width=0.28\linewidth]{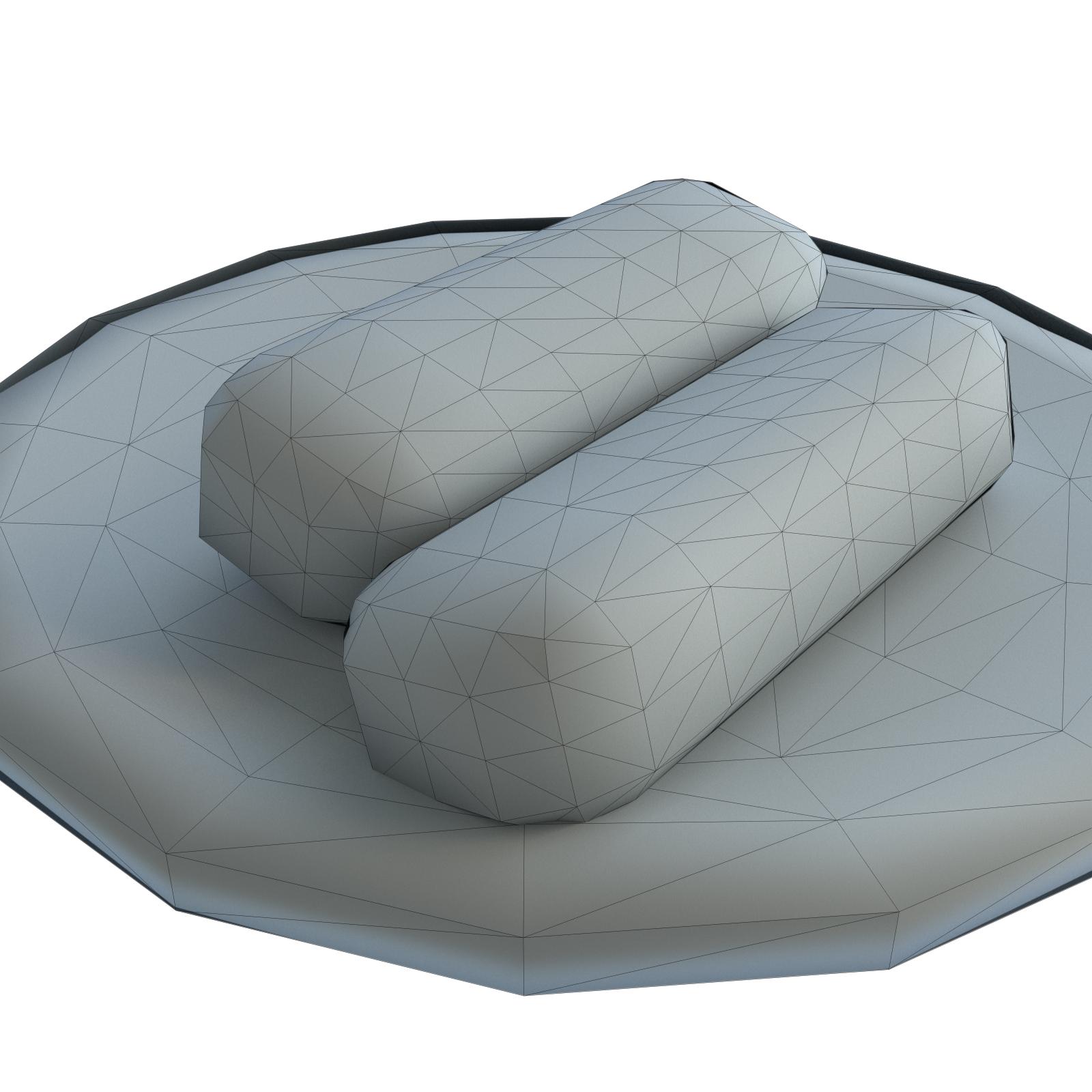} &
\includegraphics[width=0.28\textwidth]{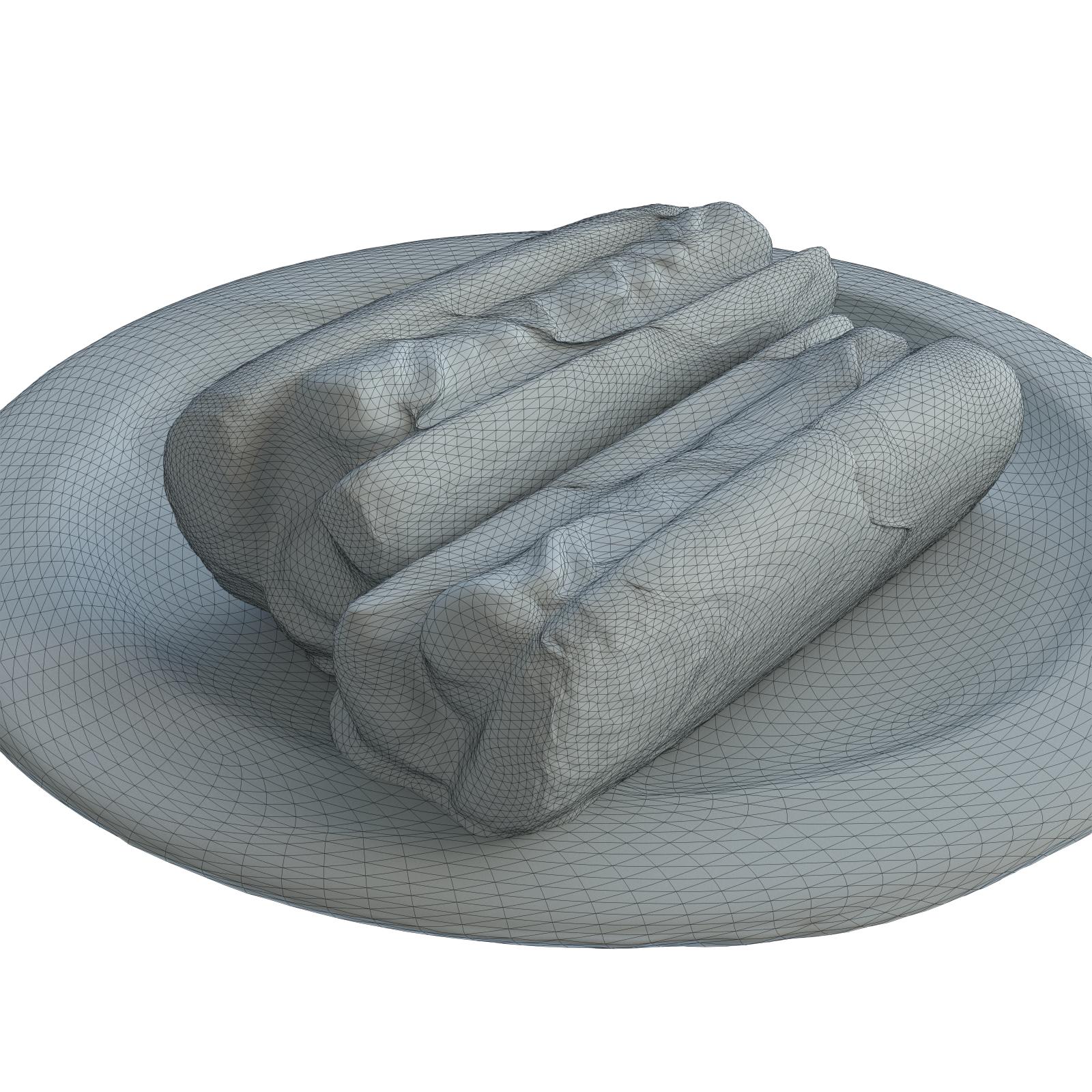} &
\includegraphics[width=0.28\textwidth]{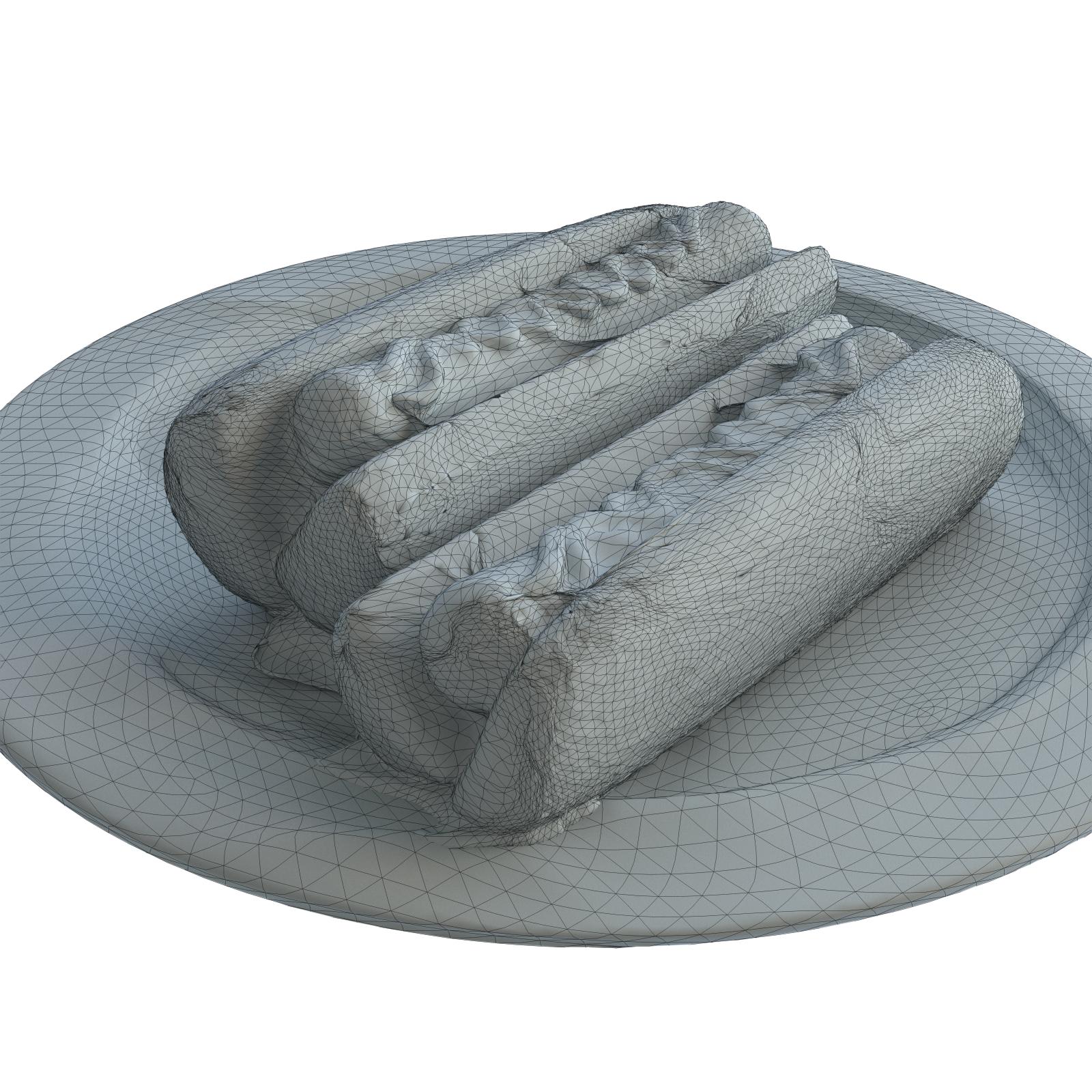} \\
\includegraphics[width=0.28\linewidth]{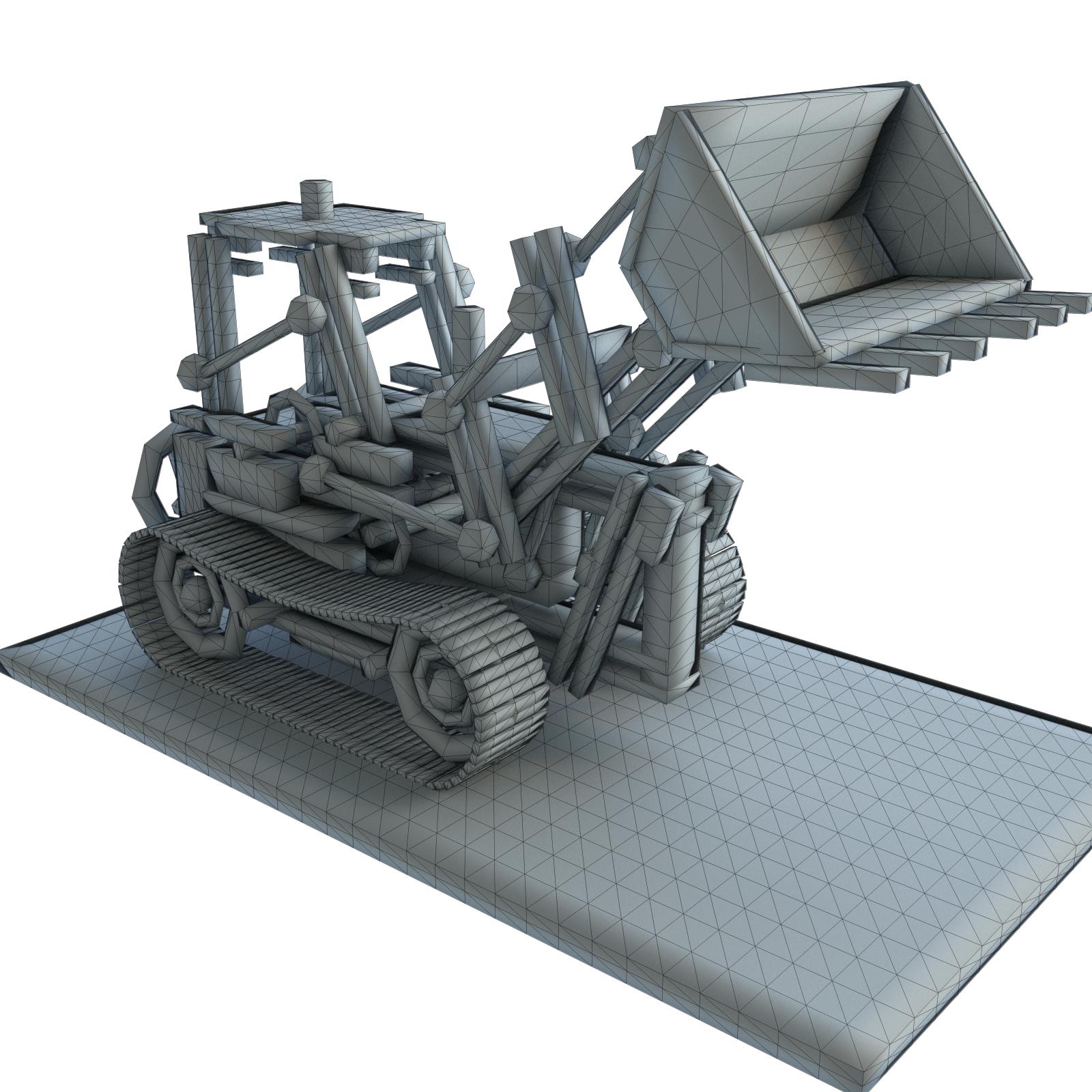} &
\includegraphics[width=0.28\textwidth]{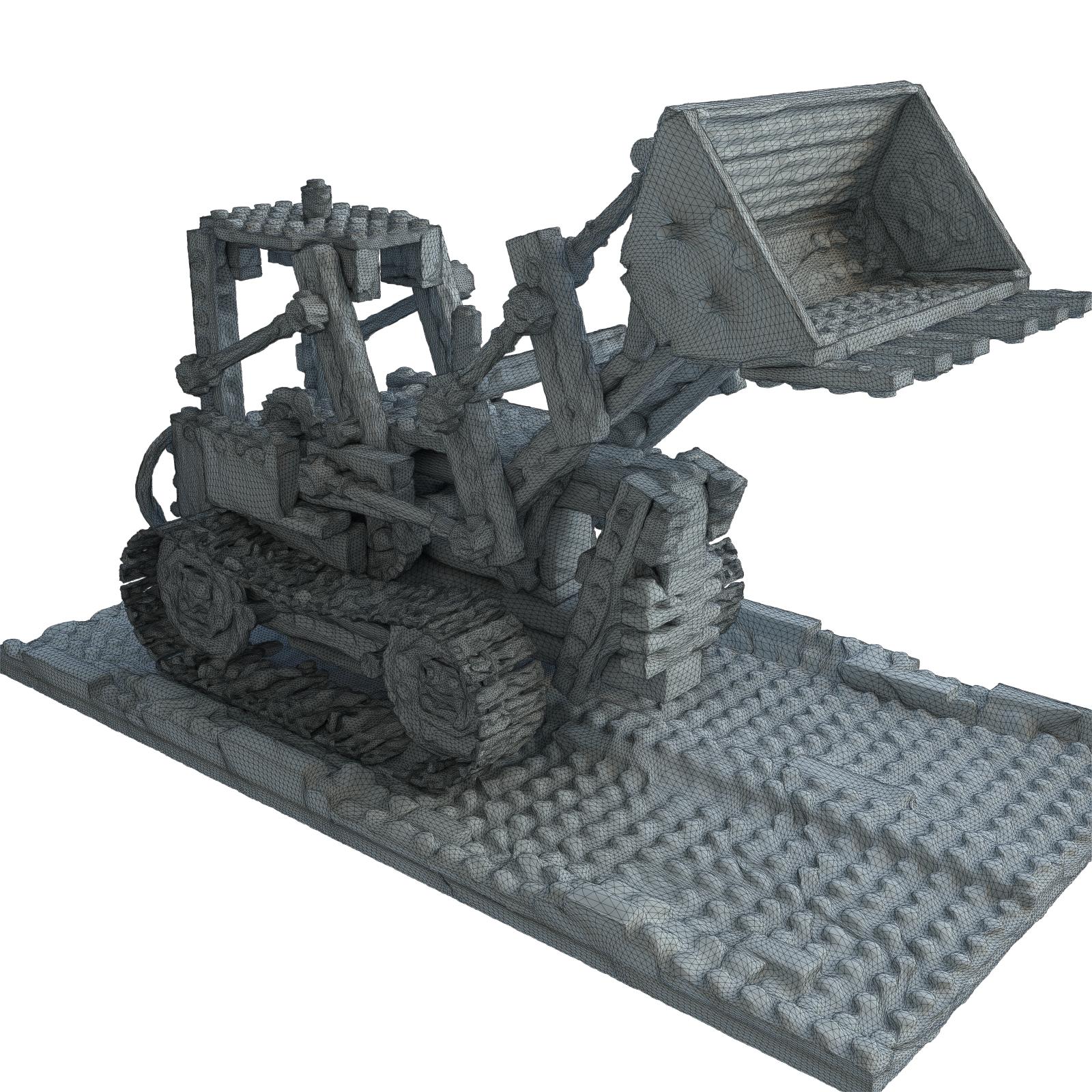} &
\includegraphics[width=0.28\textwidth]{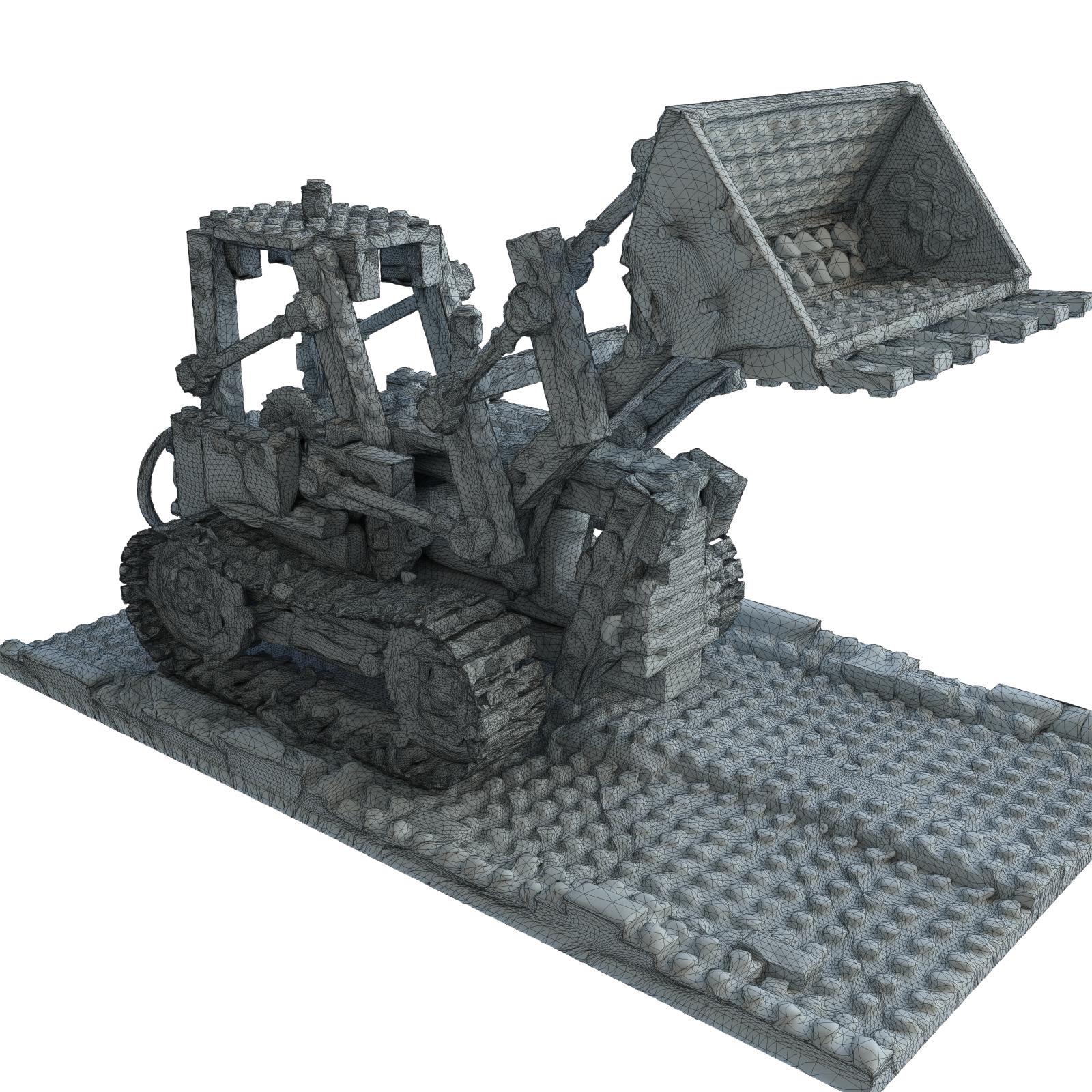} \\
\includegraphics[width=0.28\linewidth]{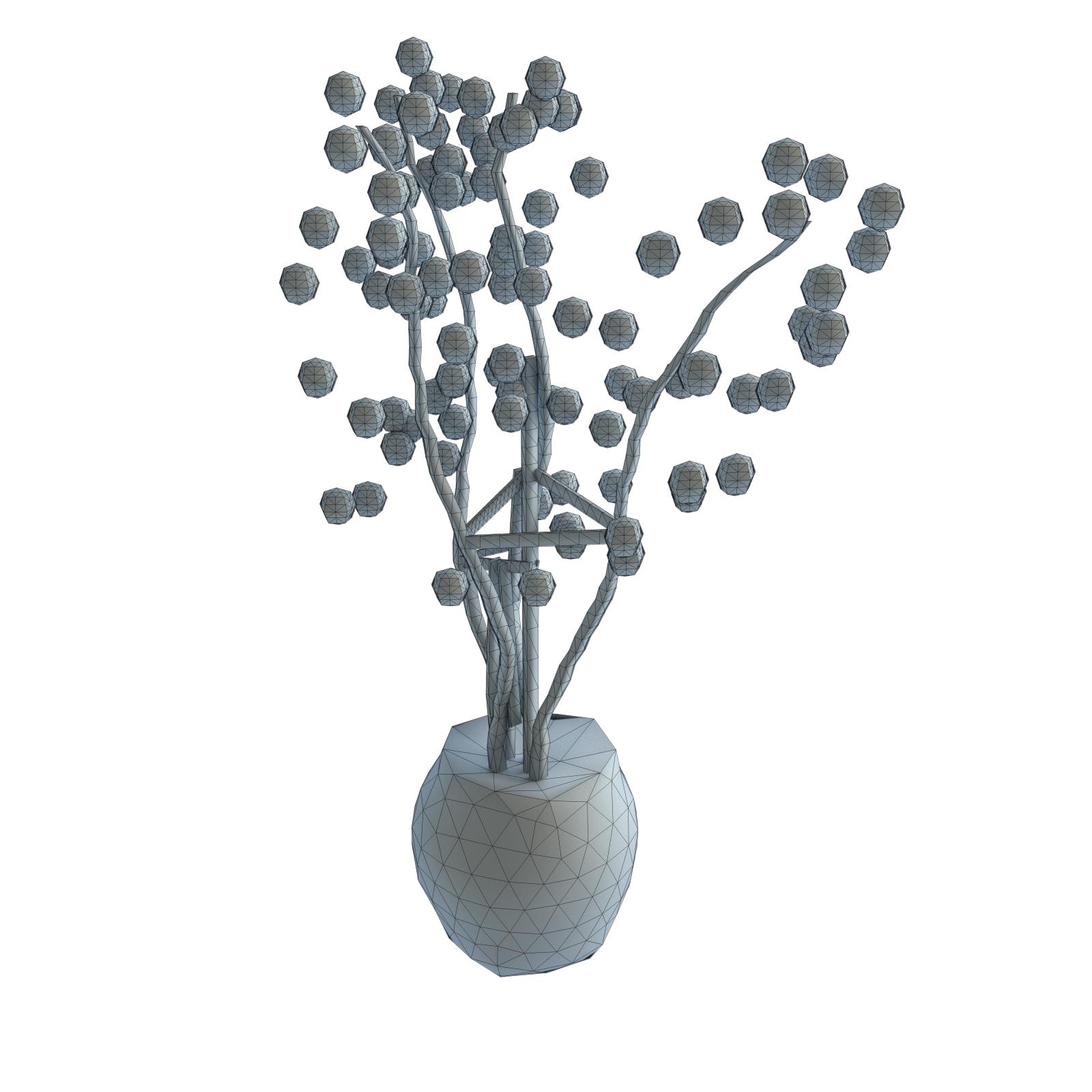} &
\includegraphics[width=0.28\textwidth]{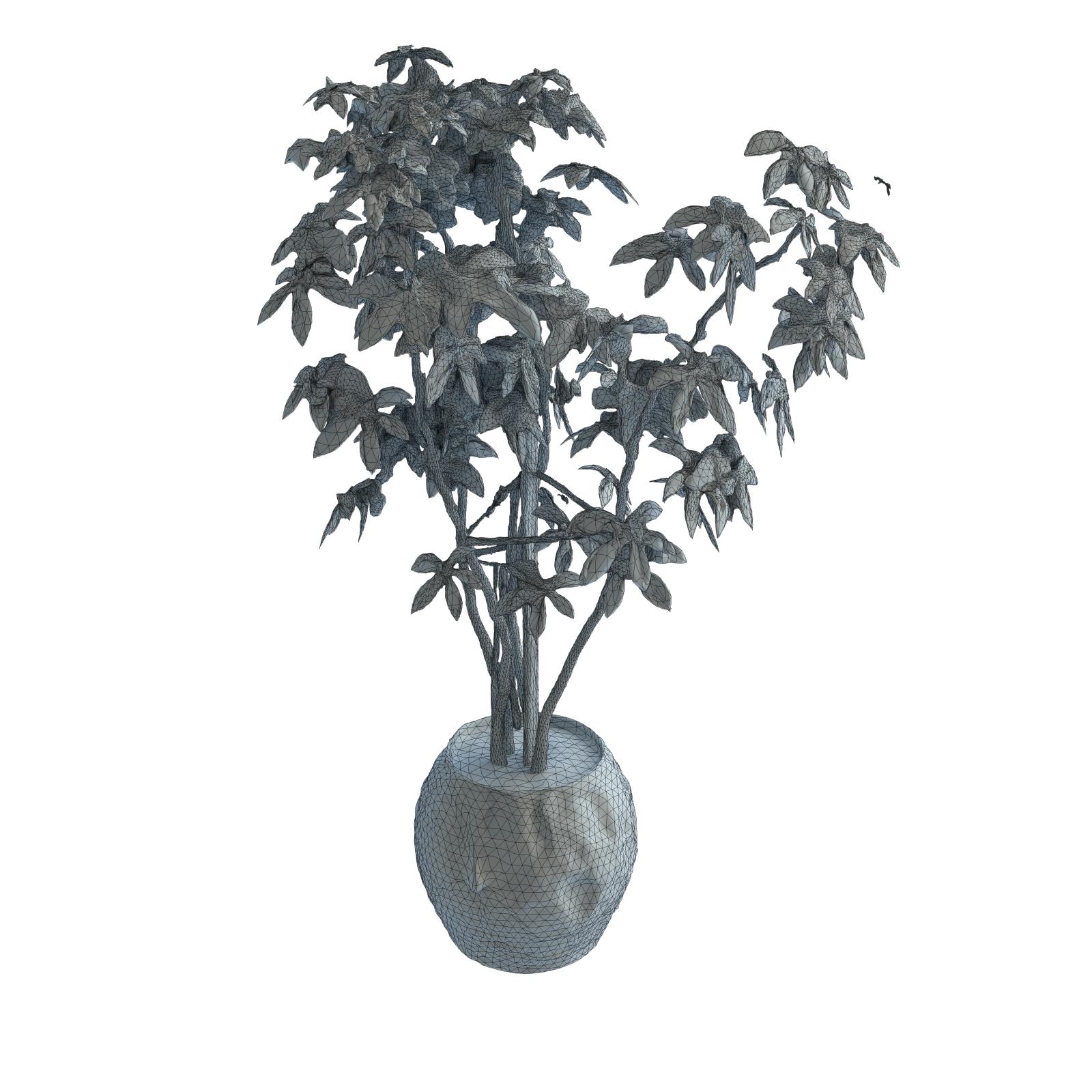} &
\includegraphics[width=0.28\textwidth]{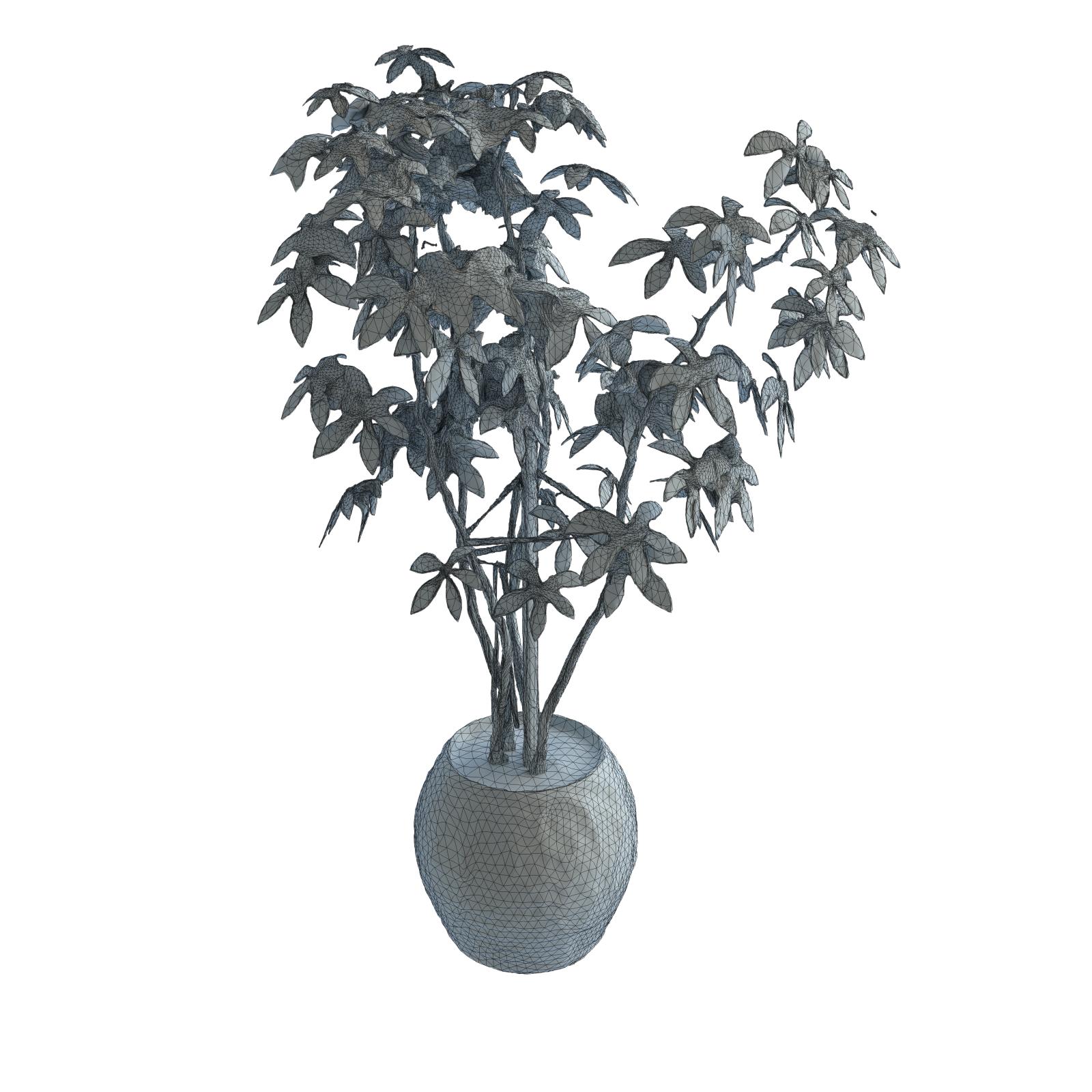} \\
\includegraphics[width=0.28\linewidth]{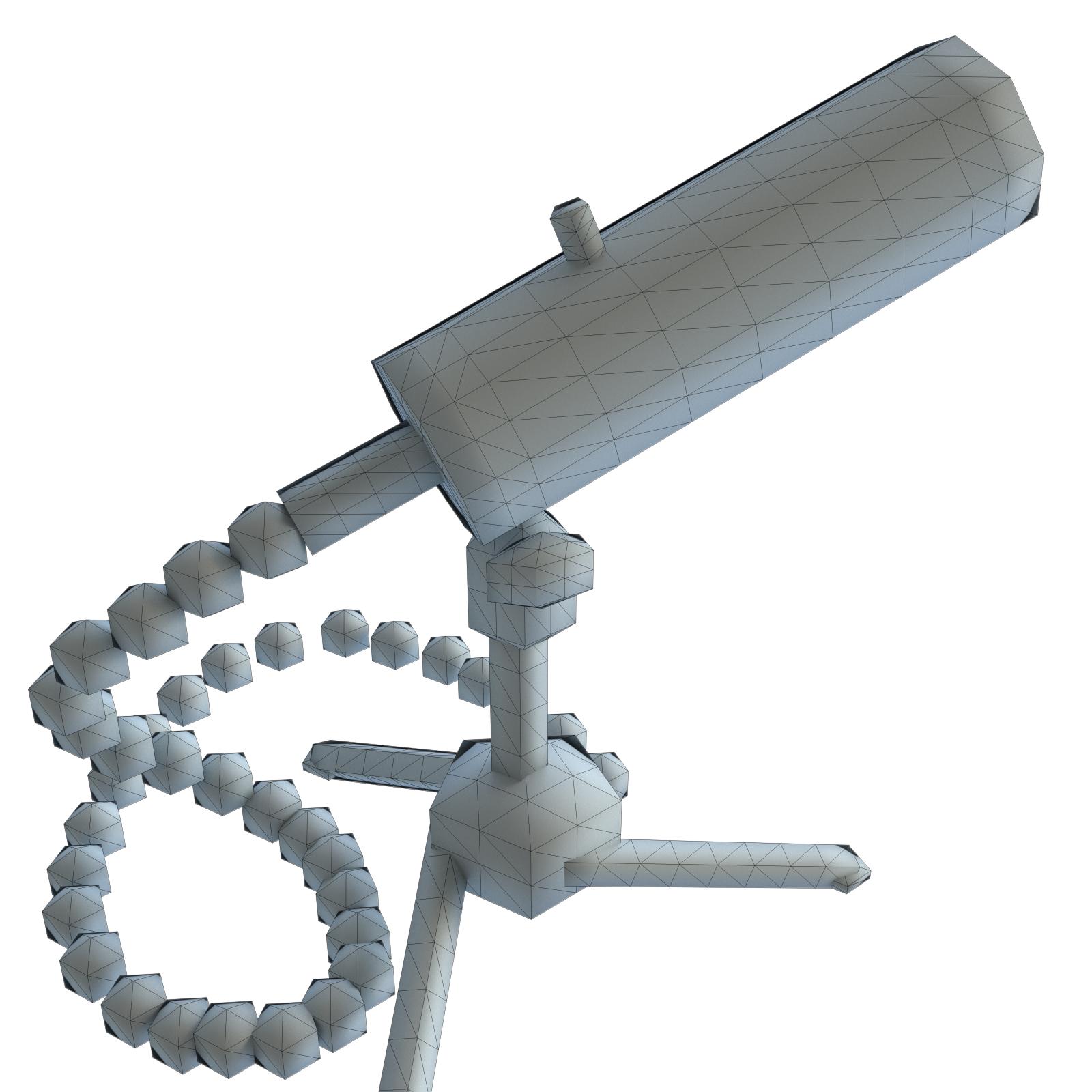} &
\includegraphics[width=0.28\textwidth]{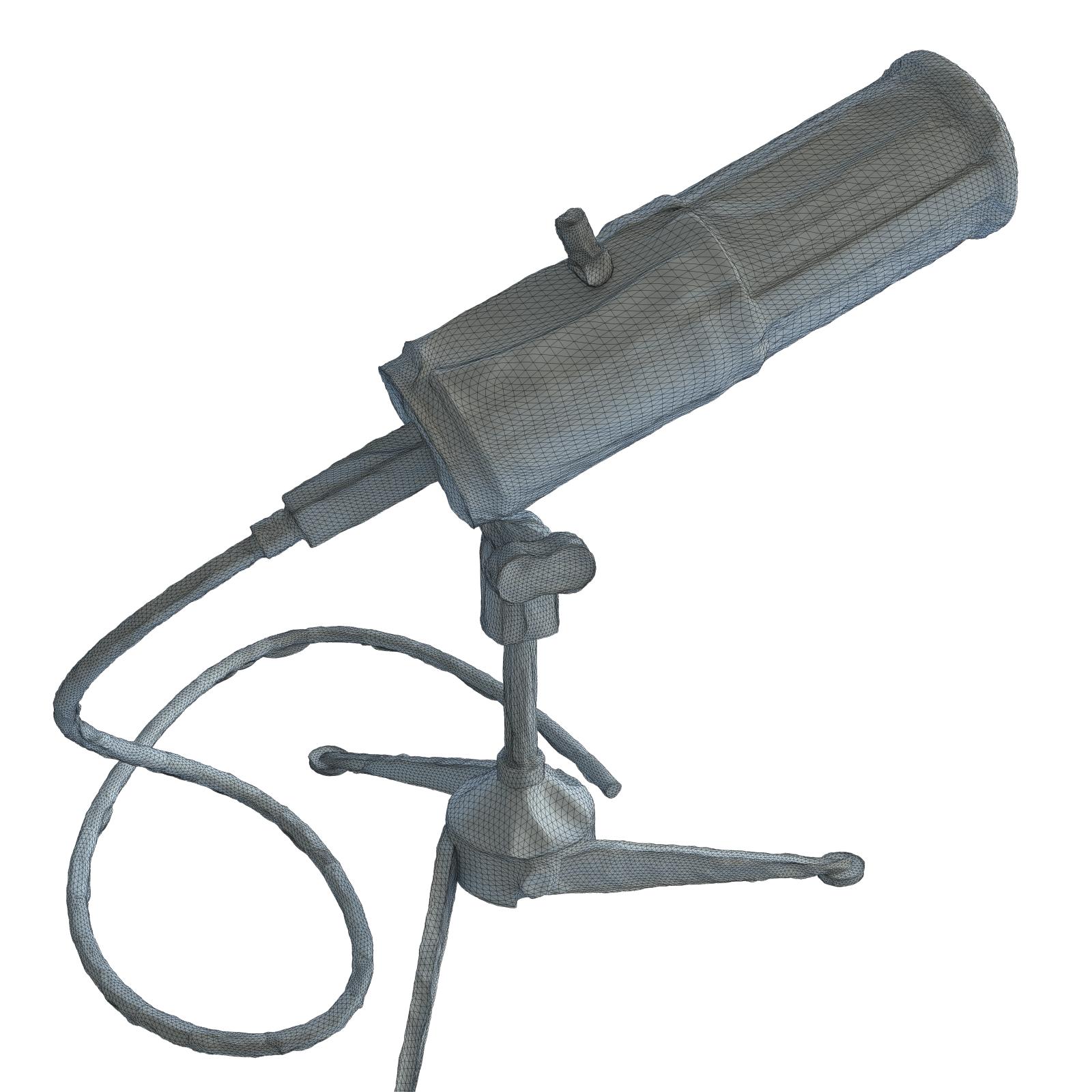} &
\includegraphics[width=0.28\textwidth]{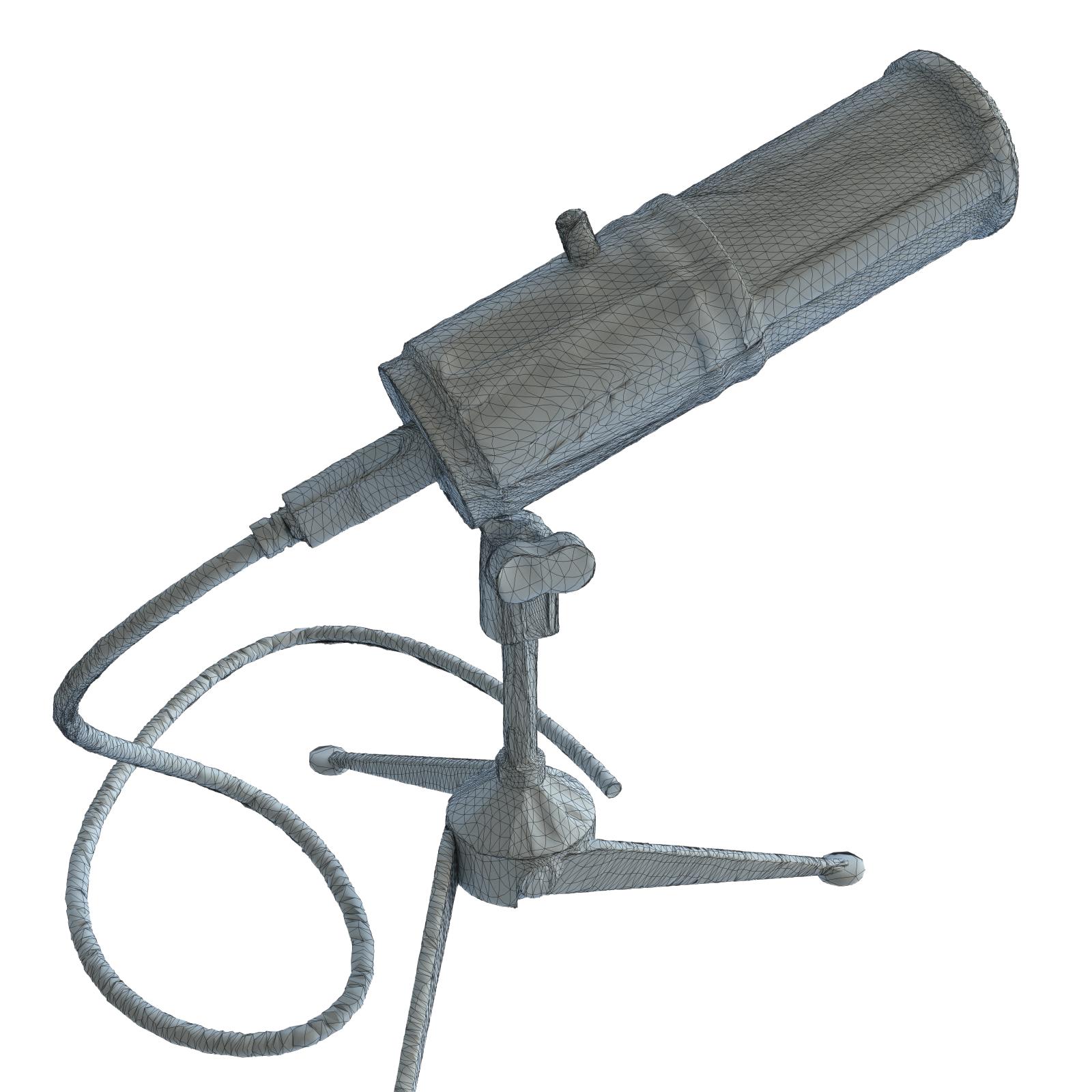} \\
\end{tabular}
\endgroup
\caption{\textbf{Blender dataset~\cite{mildenhall2020nerf}, Mesh quality}. Qualitative evaluation of reconstructed mesh on the blender dataset, test views. The first column is initialization. The second column is Nvdiffrast~\cite{laine2020modular}, third column is our method EdgeGrad.}
    \label{f-blender_mesh_1}
\end{figure*}

\begin{figure*}
\centering\footnotesize
\begingroup
\renewcommand{\arraystretch}{0.}
\setlength{\tabcolsep}{0pt}
\begin{tabular}{ccc}
Initialization mesh & Nvdiffrast~\cite{laine2020modular} & EdgeGrad (our) \\
\midrule
\includegraphics[width=0.3\linewidth]{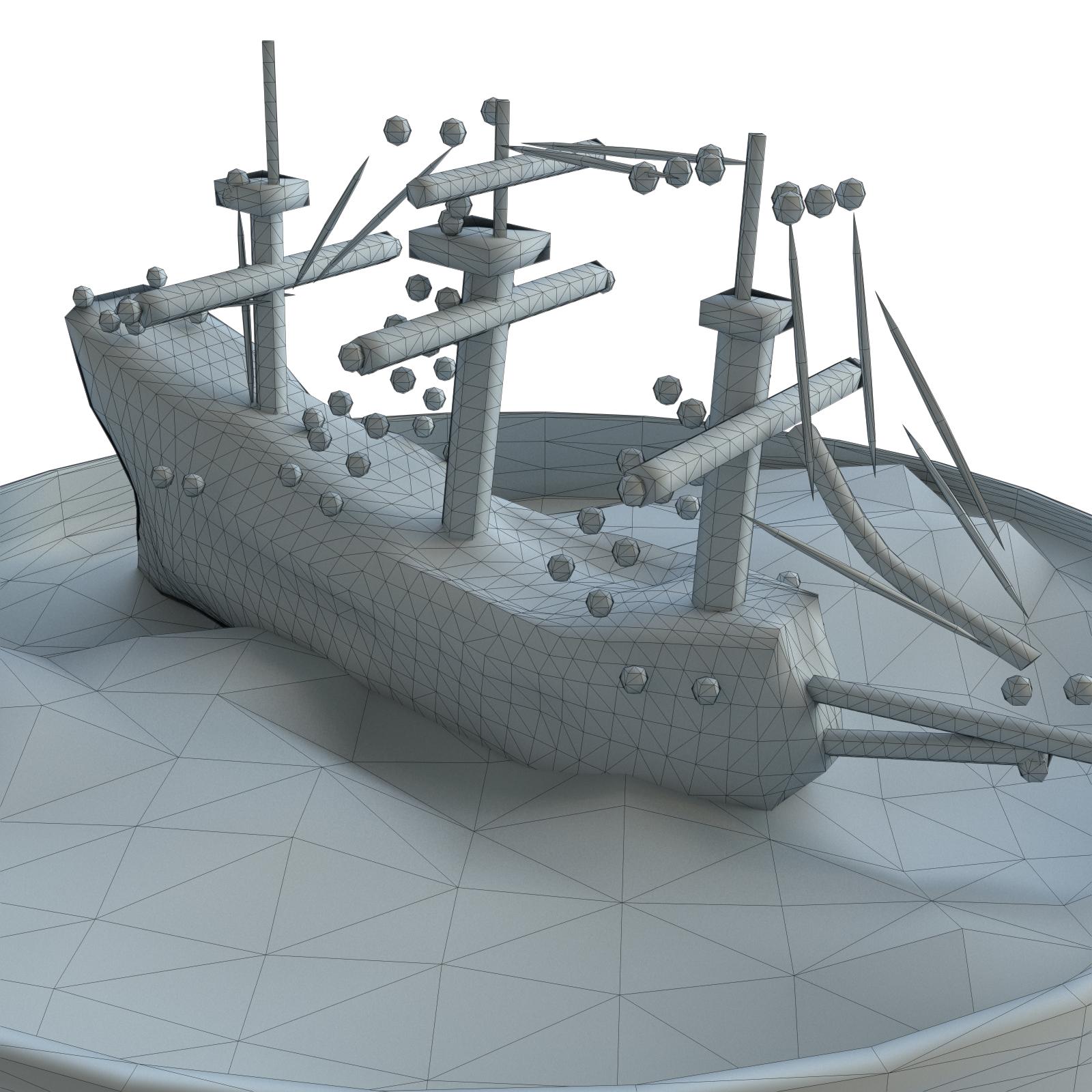} &
\includegraphics[width=0.3\textwidth]{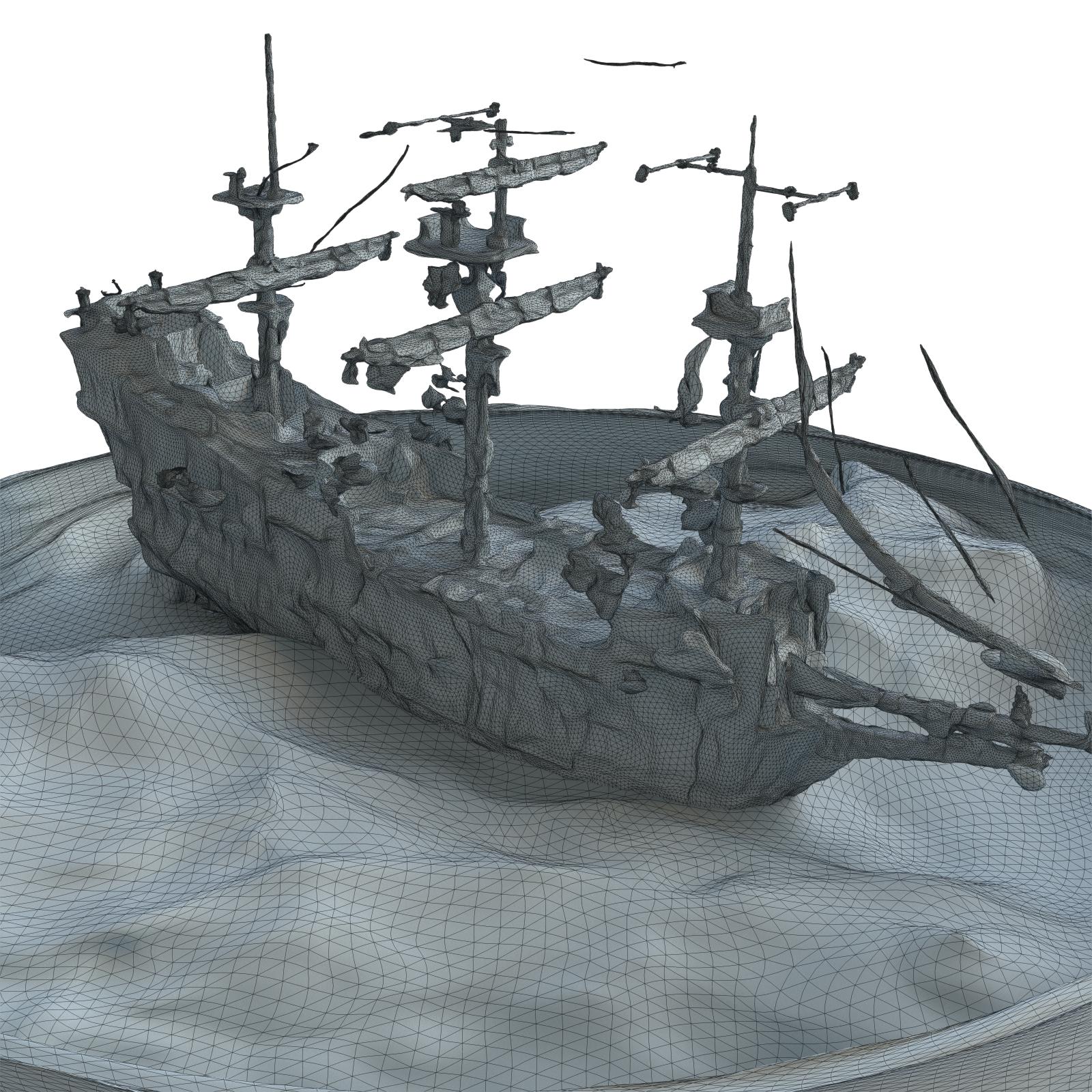} &
\includegraphics[width=0.3\textwidth]{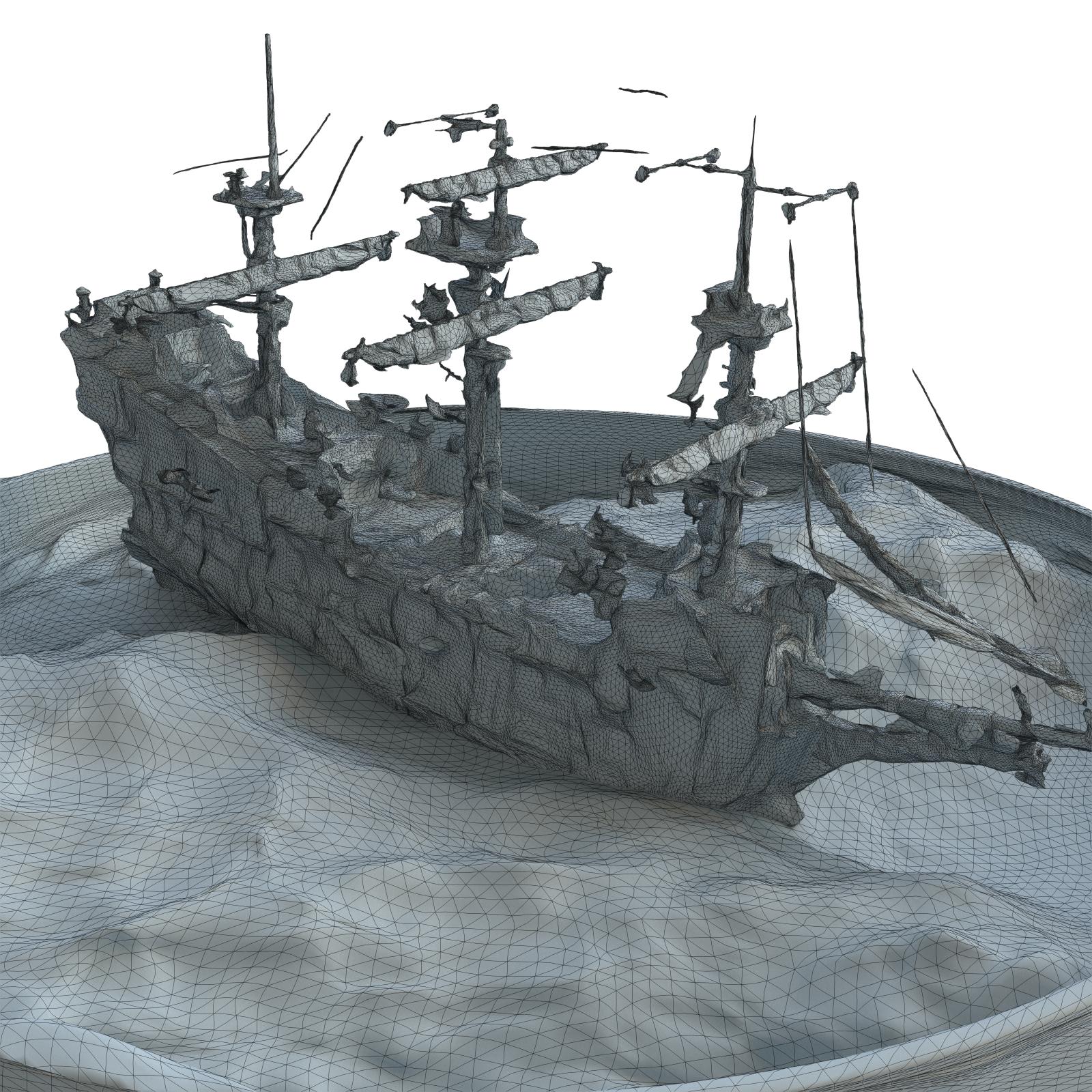} \\
\includegraphics[width=0.3\linewidth]{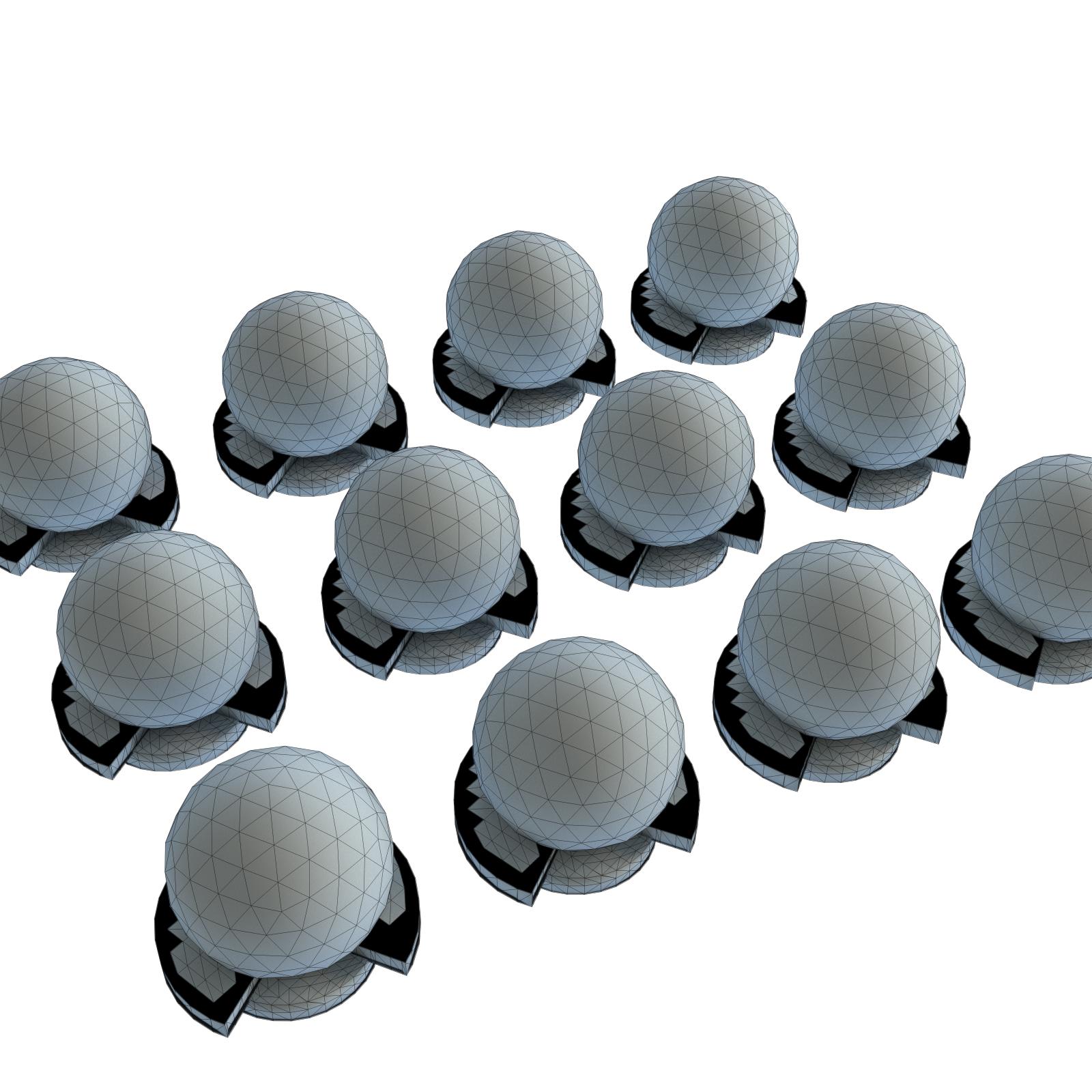} &
\includegraphics[width=0.3\textwidth]{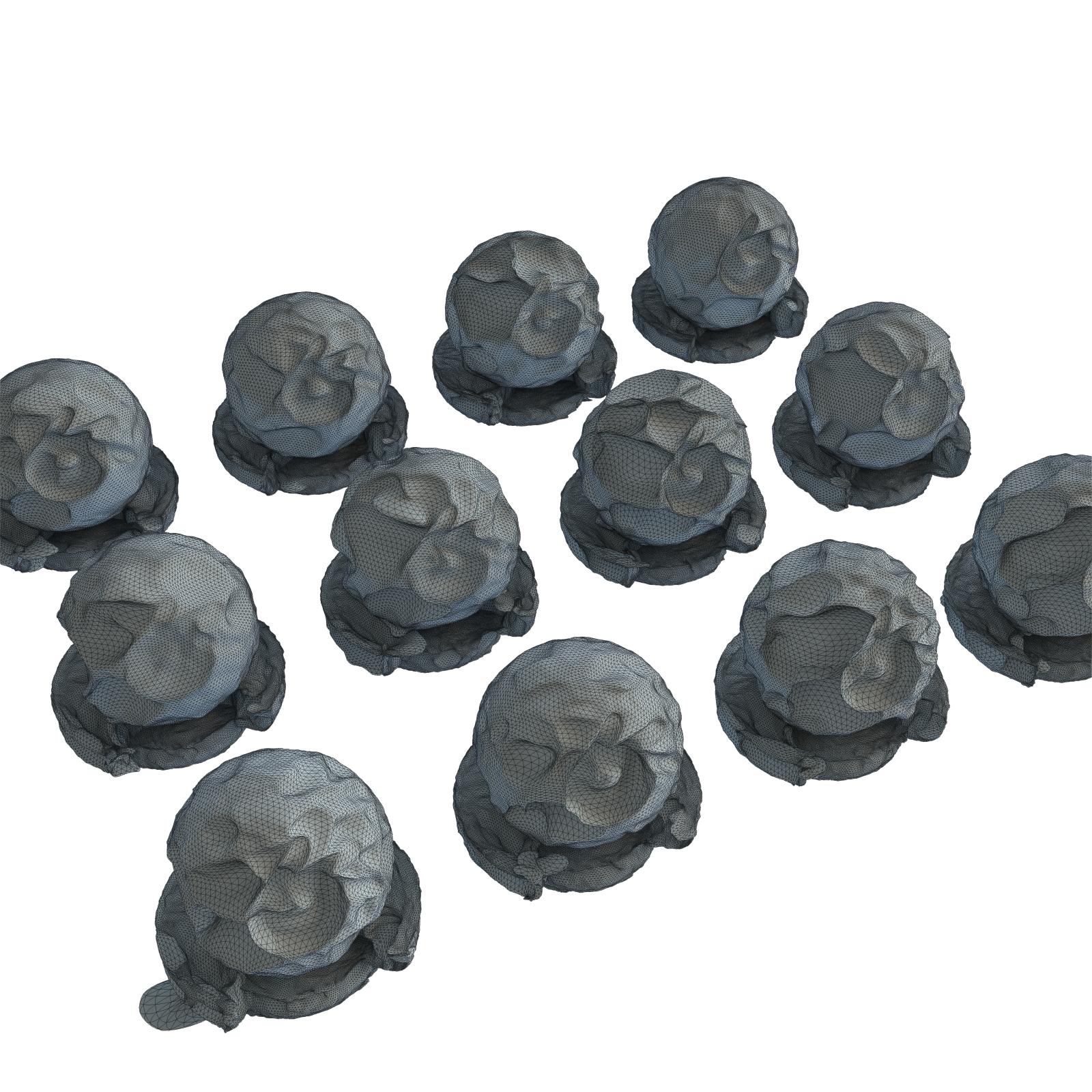} &
\includegraphics[width=0.3\textwidth]{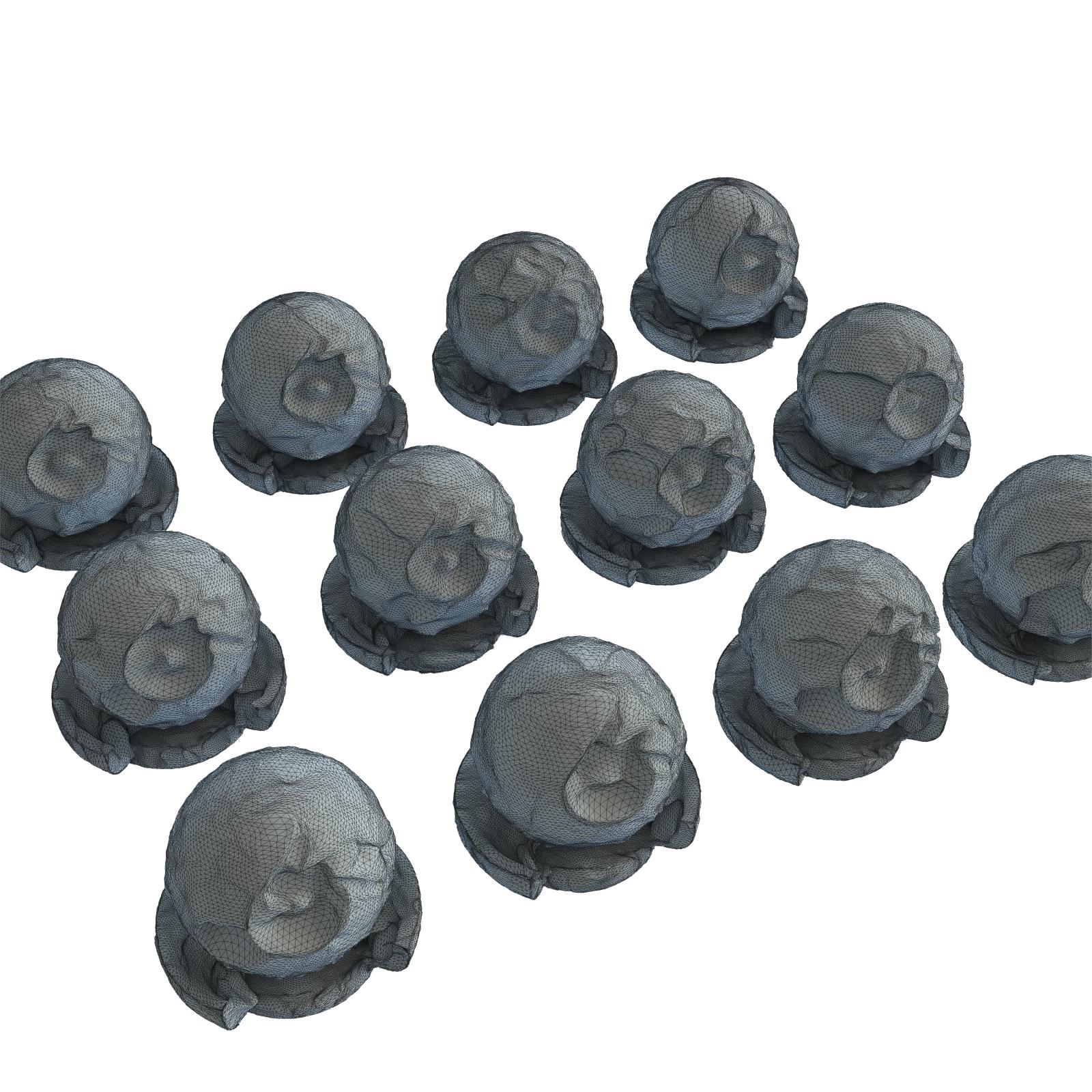} \\
\includegraphics[width=0.3\linewidth]{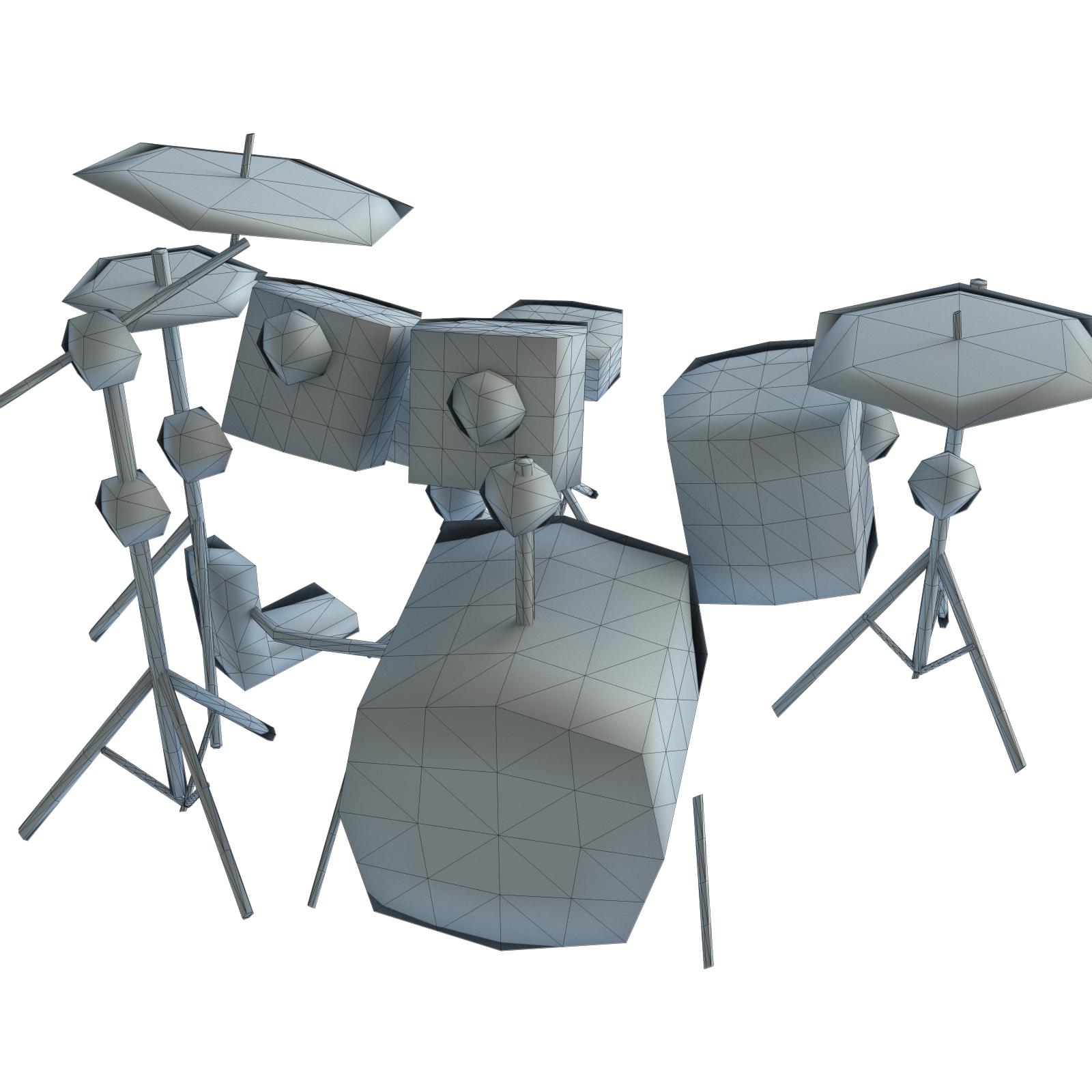} &
\includegraphics[width=0.3\textwidth]{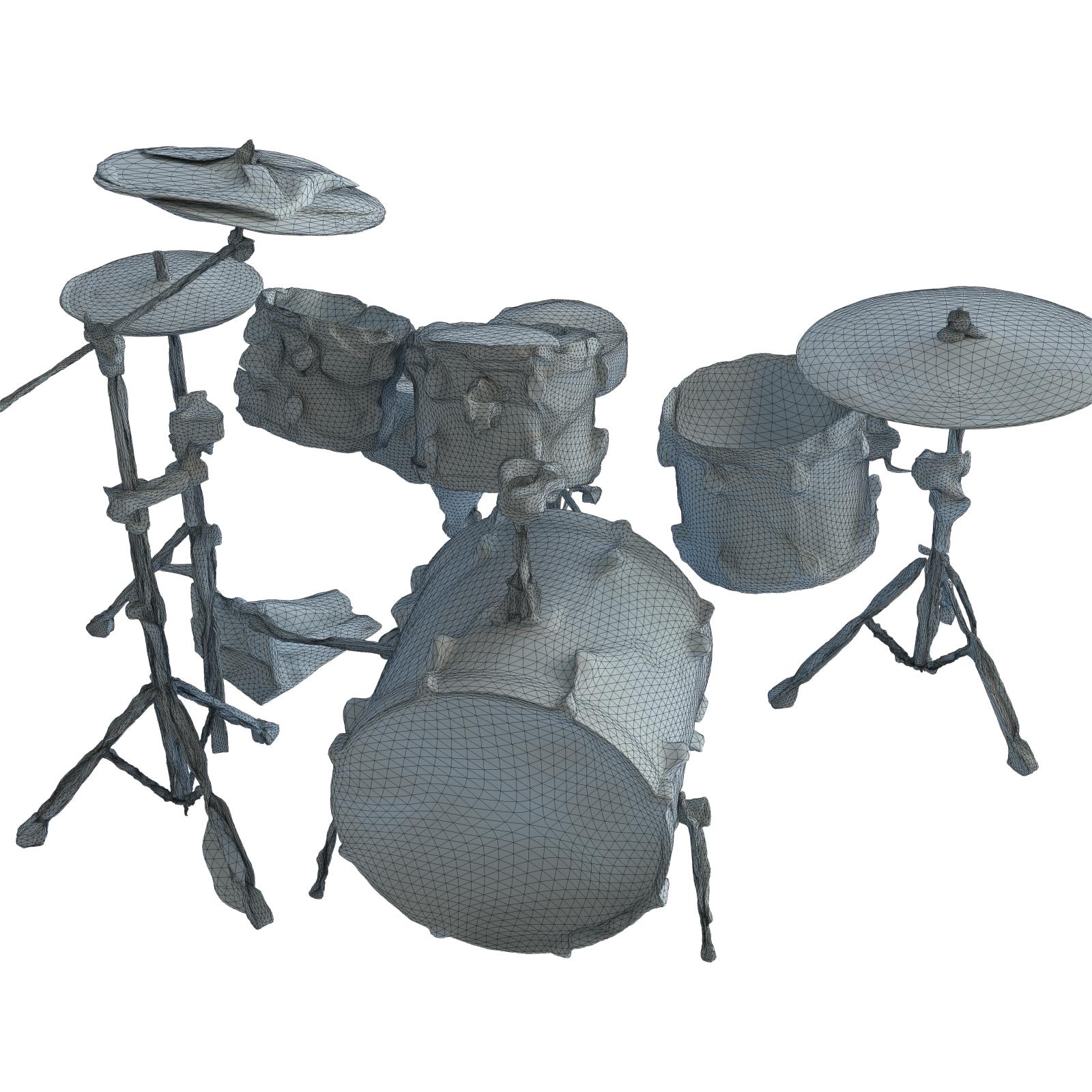} &
\includegraphics[width=0.3\textwidth]{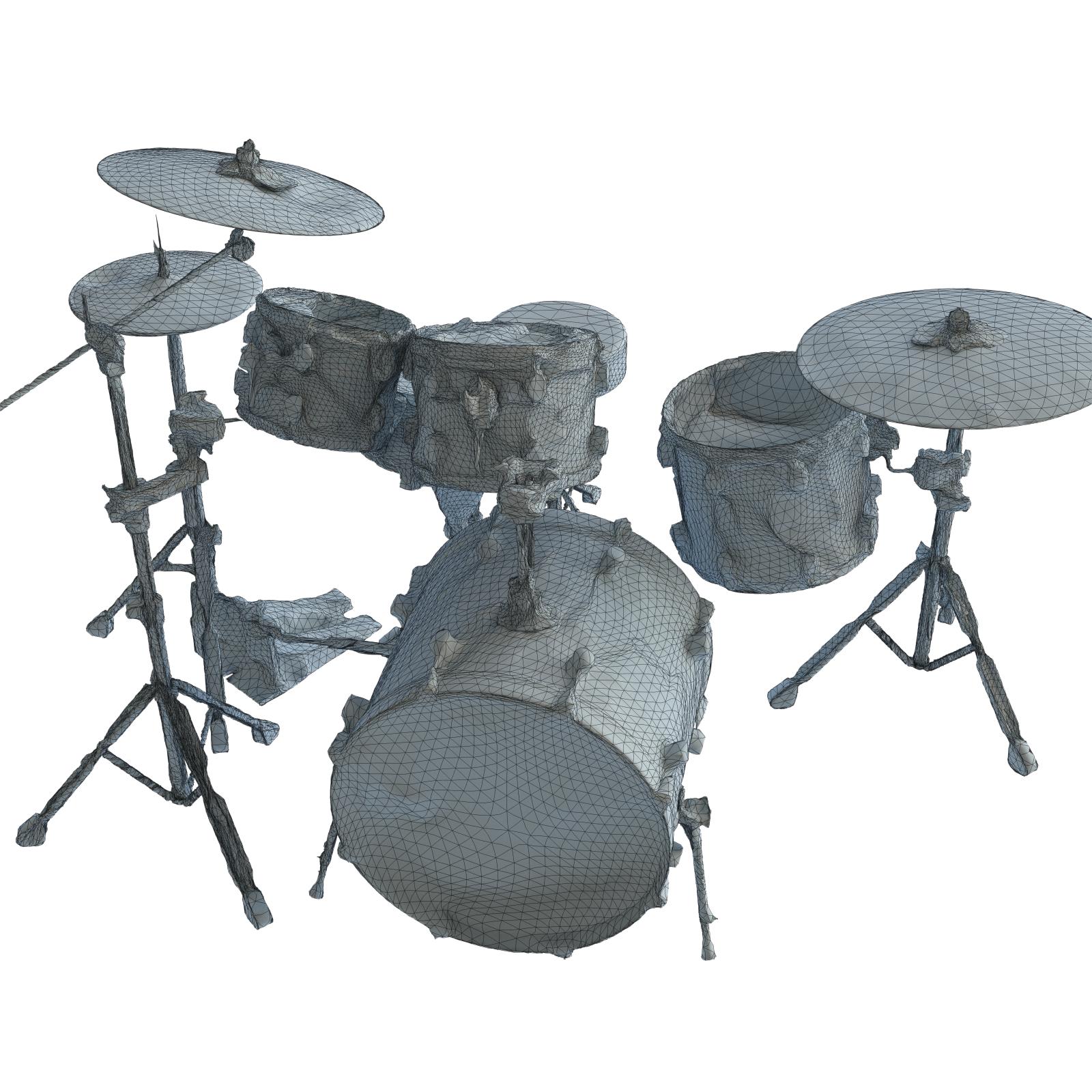} \\

\end{tabular}
\endgroup
\caption{\textbf{Blender dataset~\cite{mildenhall2020nerf}, Mesh quality (Continuation of Fig.~\ref{f-blender_mesh_1})}. Qualitative evaluation of reconstructed mesh on the blender dataset, test views. The first column is initialization. The second column is Nvdiffrast~\cite{laine2020modular}, third column is our method EdgeGrad.}
    \label{f-blender_mesh_2}
\end{figure*}

\begin{figure*}
\centering\footnotesize
\begingroup
\renewcommand{\arraystretch}{0.}
\setlength{\tabcolsep}{0pt}
\begin{tabular}{ccc}
Continuous only & EdgeGrad  -intersections & EdgeGrad (our) \\
\midrule
\includegraphics[width=0.3\textwidth]{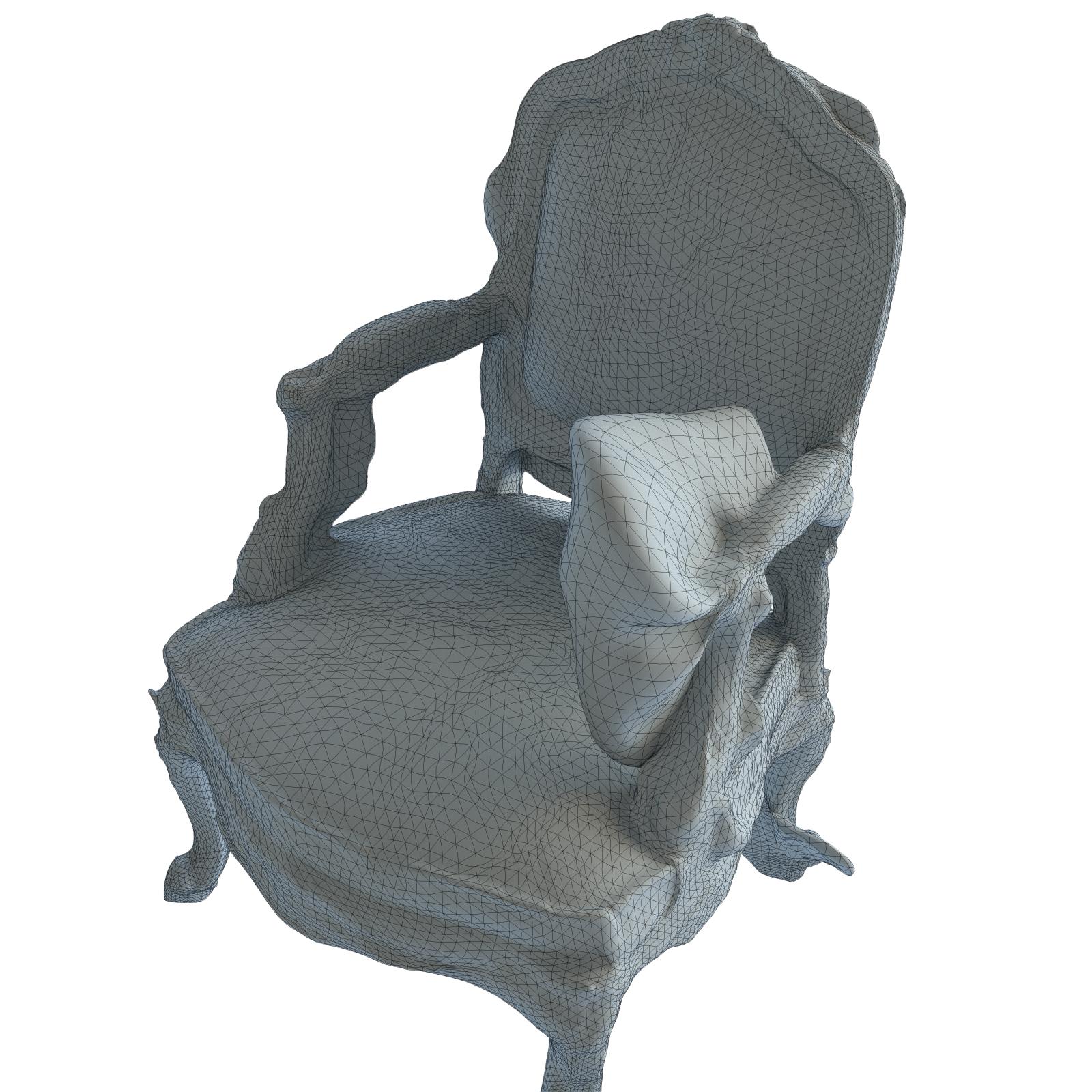} &
\includegraphics[width=0.3\textwidth]{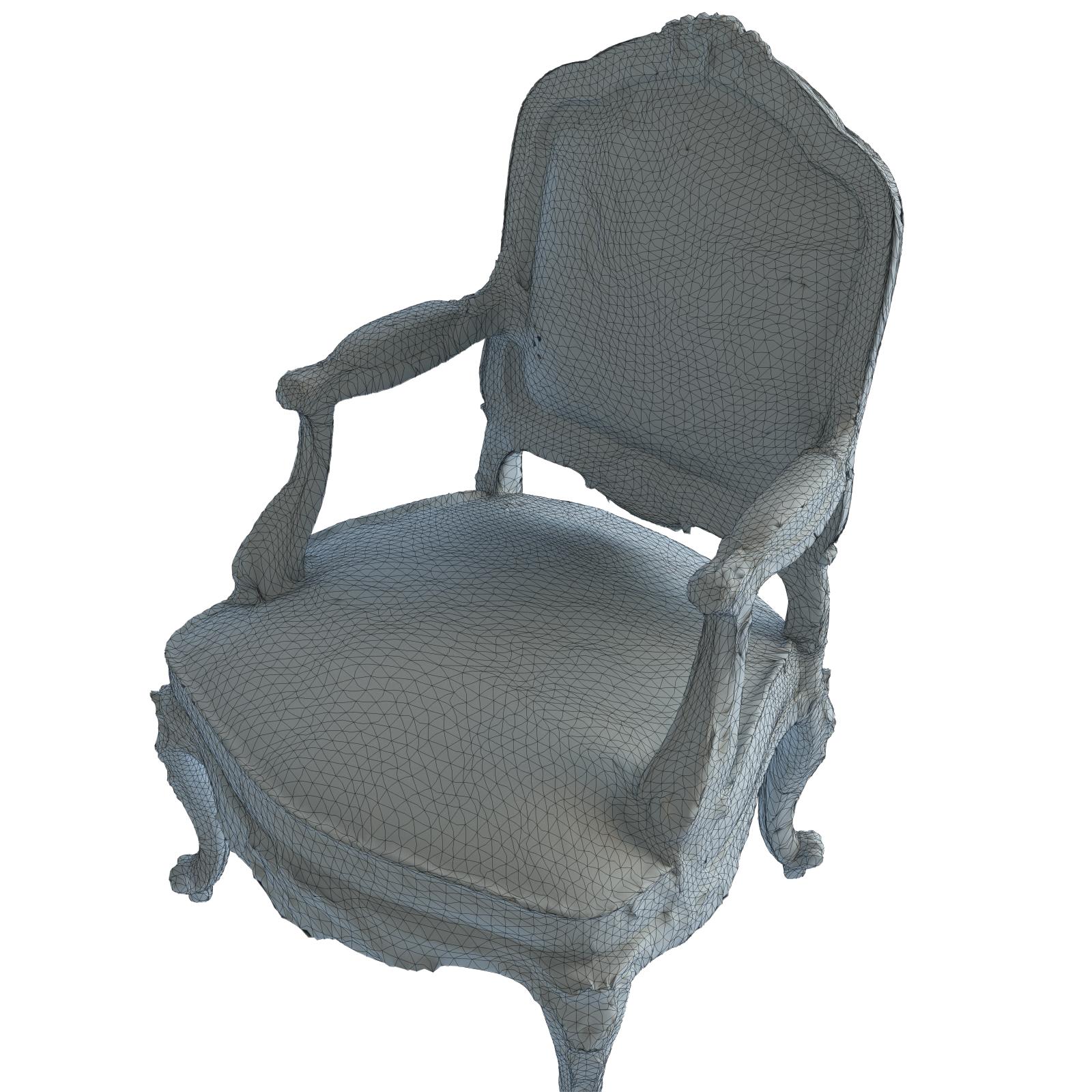} &
\includegraphics[width=0.3\textwidth]{figures_jpg/out_blender/chair000000_mi_w} \\
\includegraphics[width=0.3\linewidth]{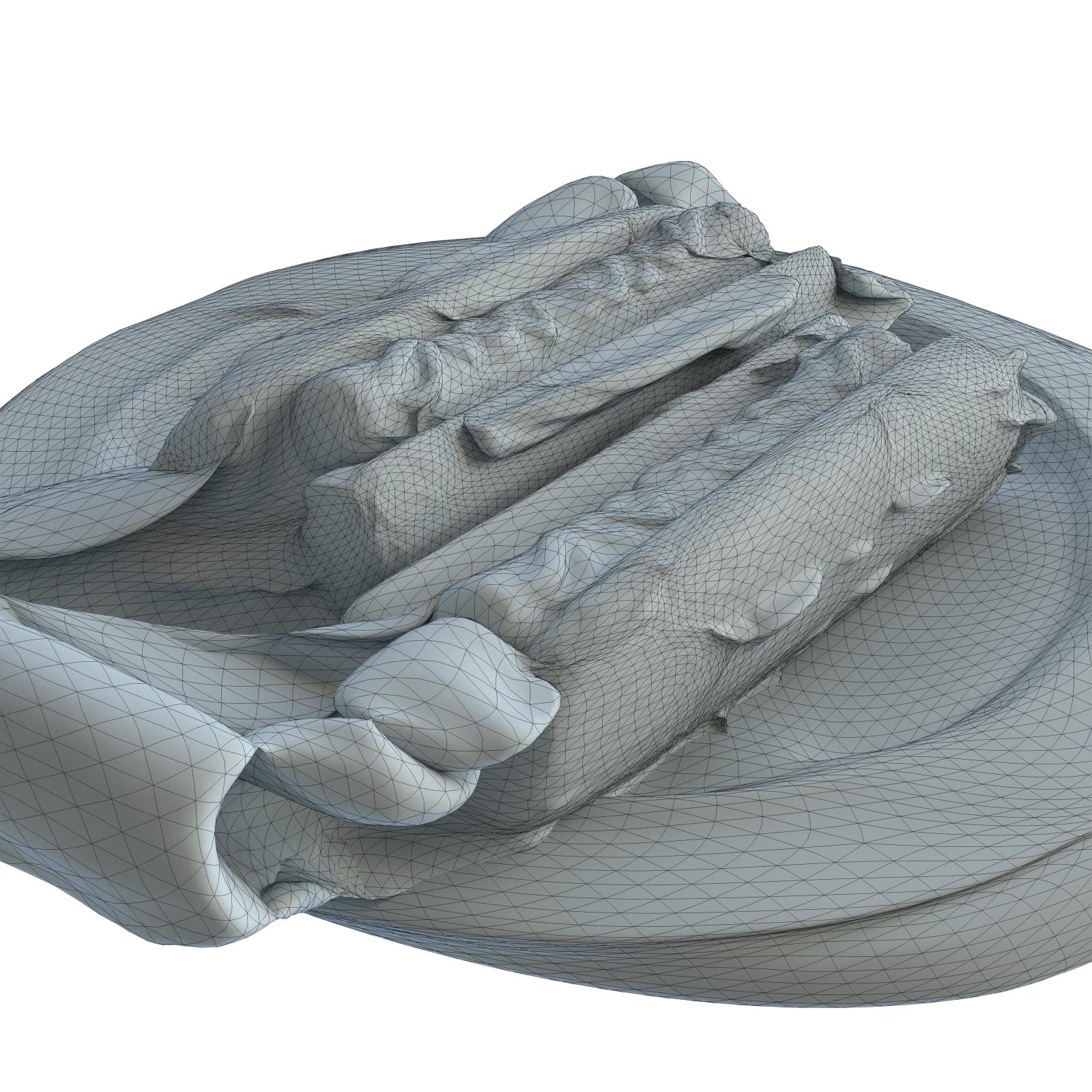} &
\includegraphics[width=0.3\linewidth]{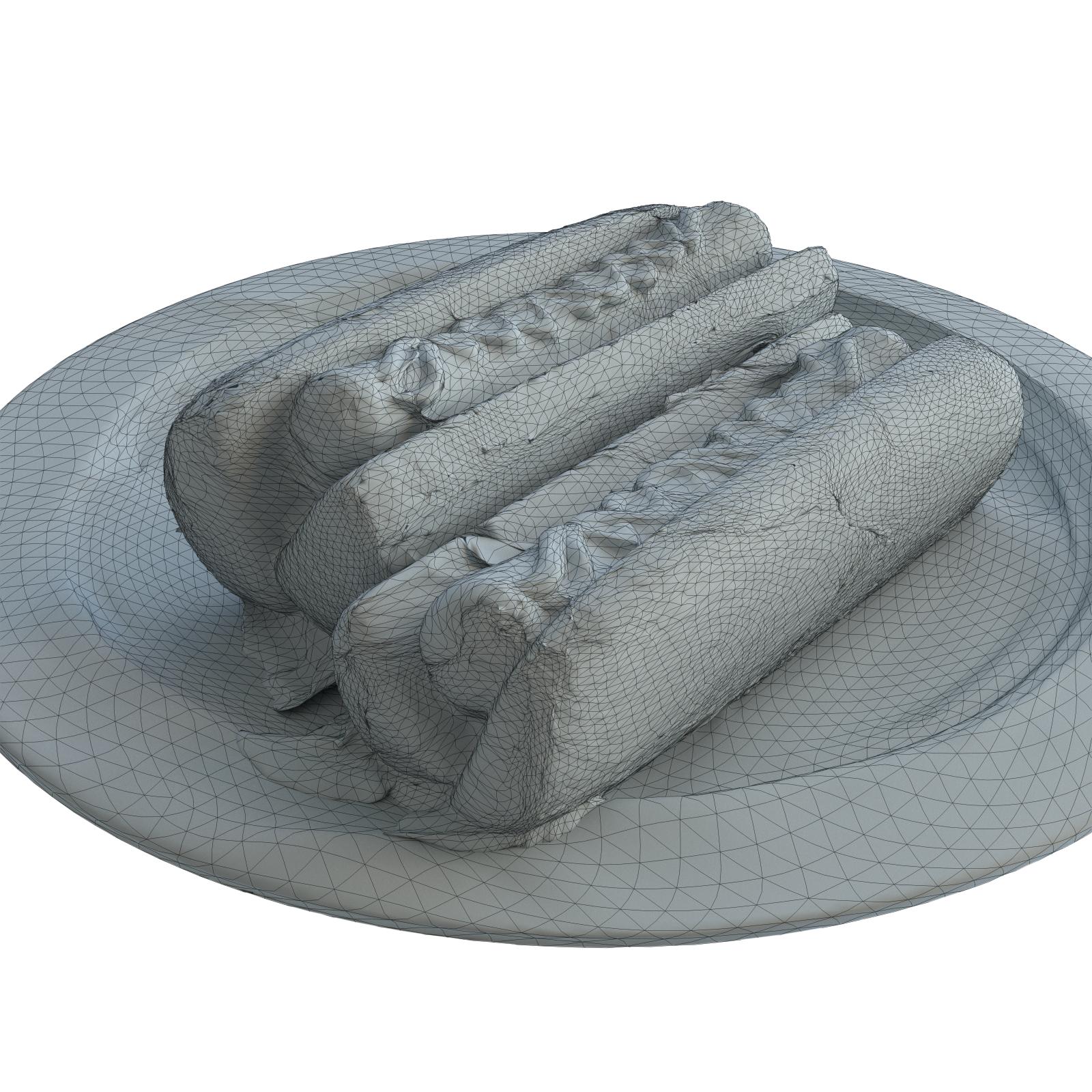} &
\includegraphics[width=0.28\textwidth]{figures_jpg/out_blender/hotdog000003_mi_w} \\
\includegraphics[width=0.28\linewidth]{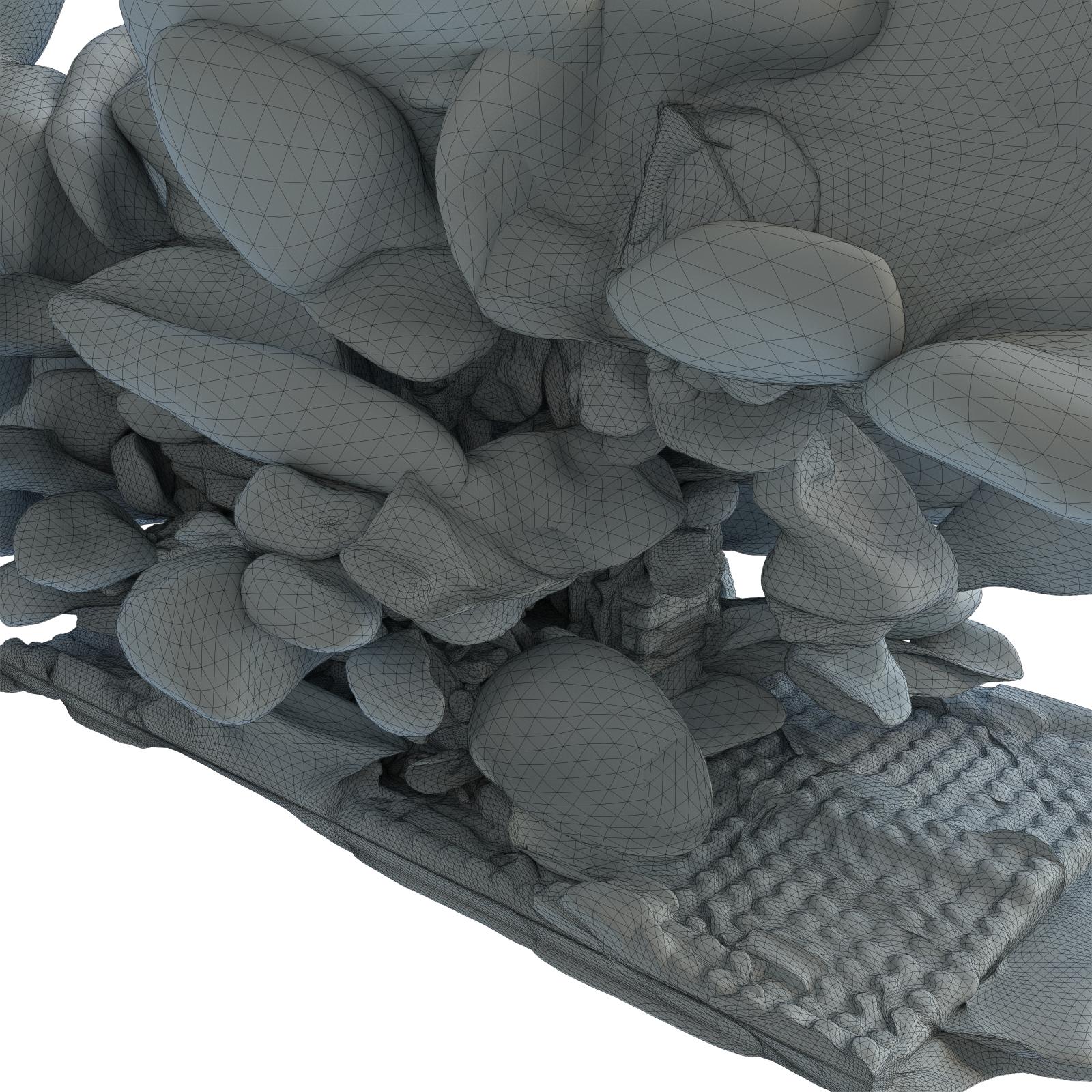} &
\includegraphics[width=0.28\textwidth]{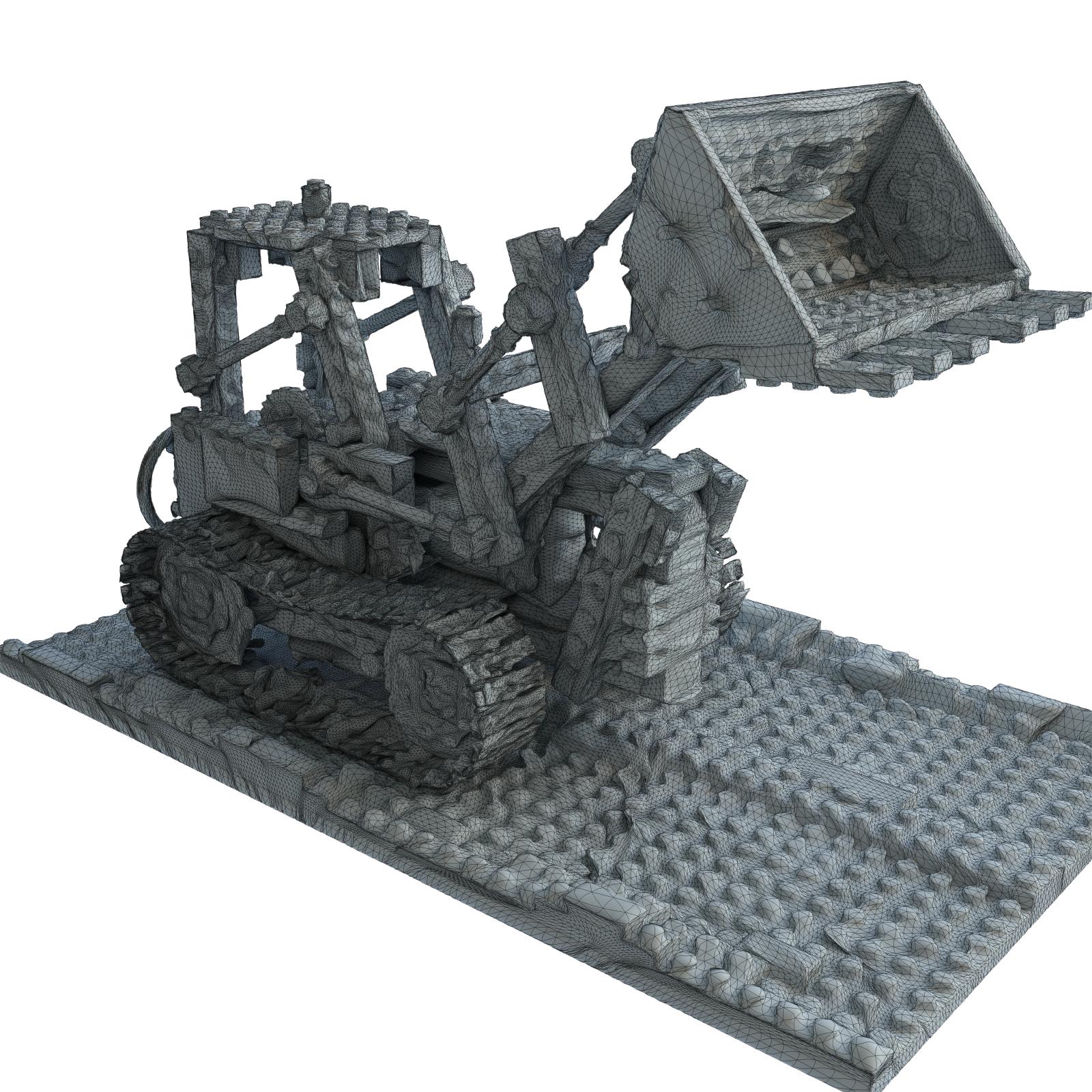} &
\includegraphics[width=0.28\textwidth]{figures_jpg/out_blender/lego000004_mi_w} \\
\includegraphics[width=0.28\linewidth]{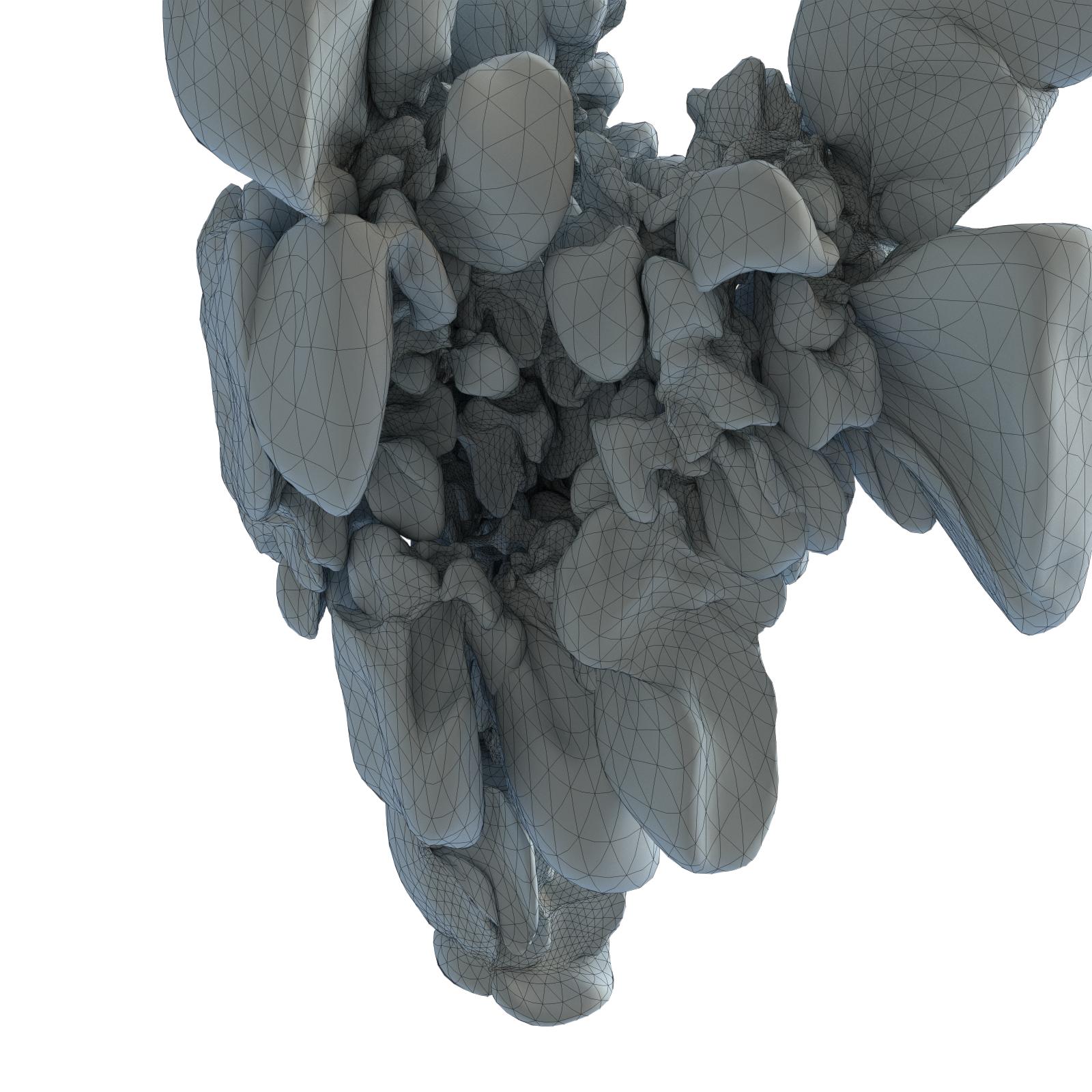} &
\includegraphics[width=0.28\textwidth]{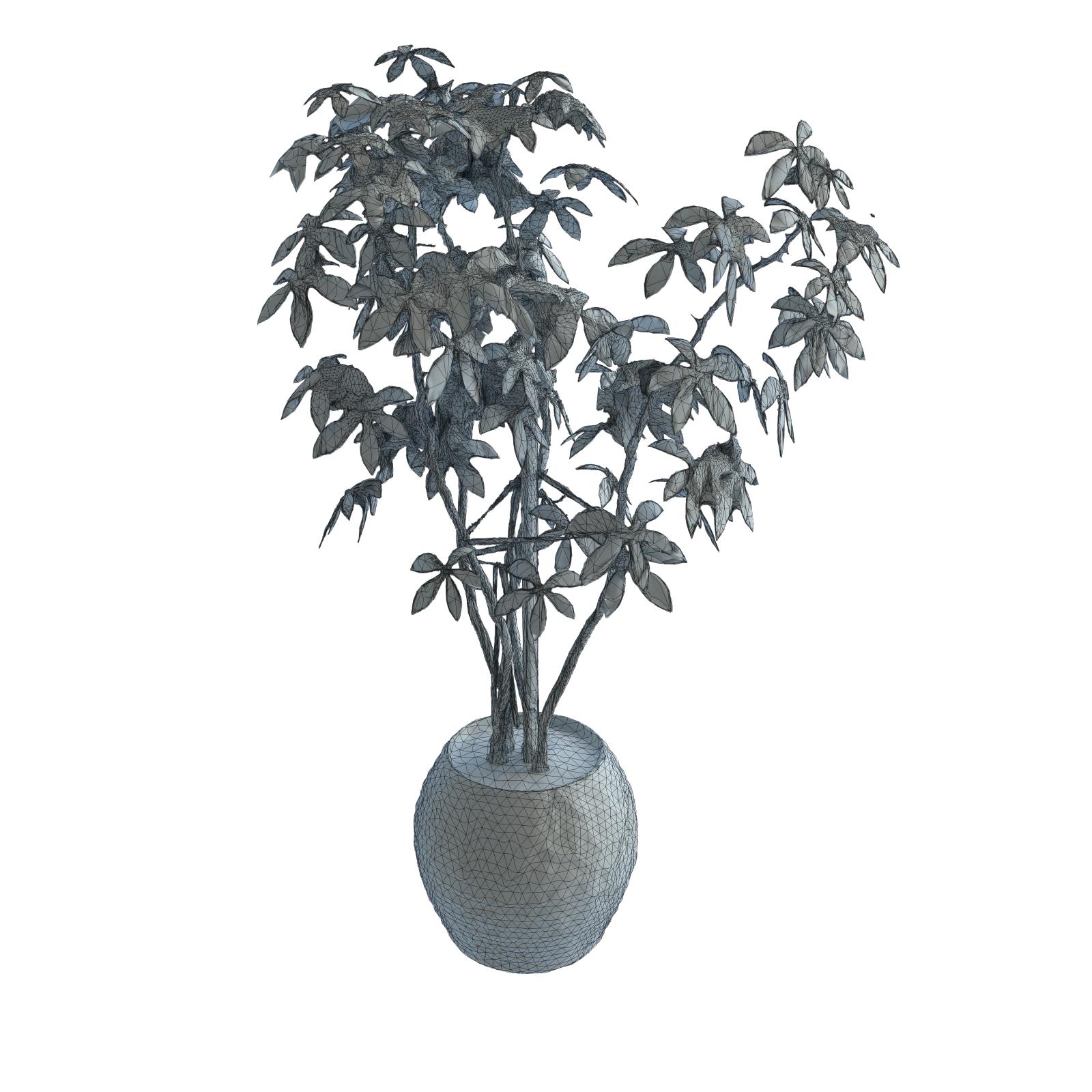} &
\includegraphics[width=0.28\textwidth]{figures_jpg/out_blender/ficus000002_mi_w} \\
\includegraphics[width=0.28\linewidth]{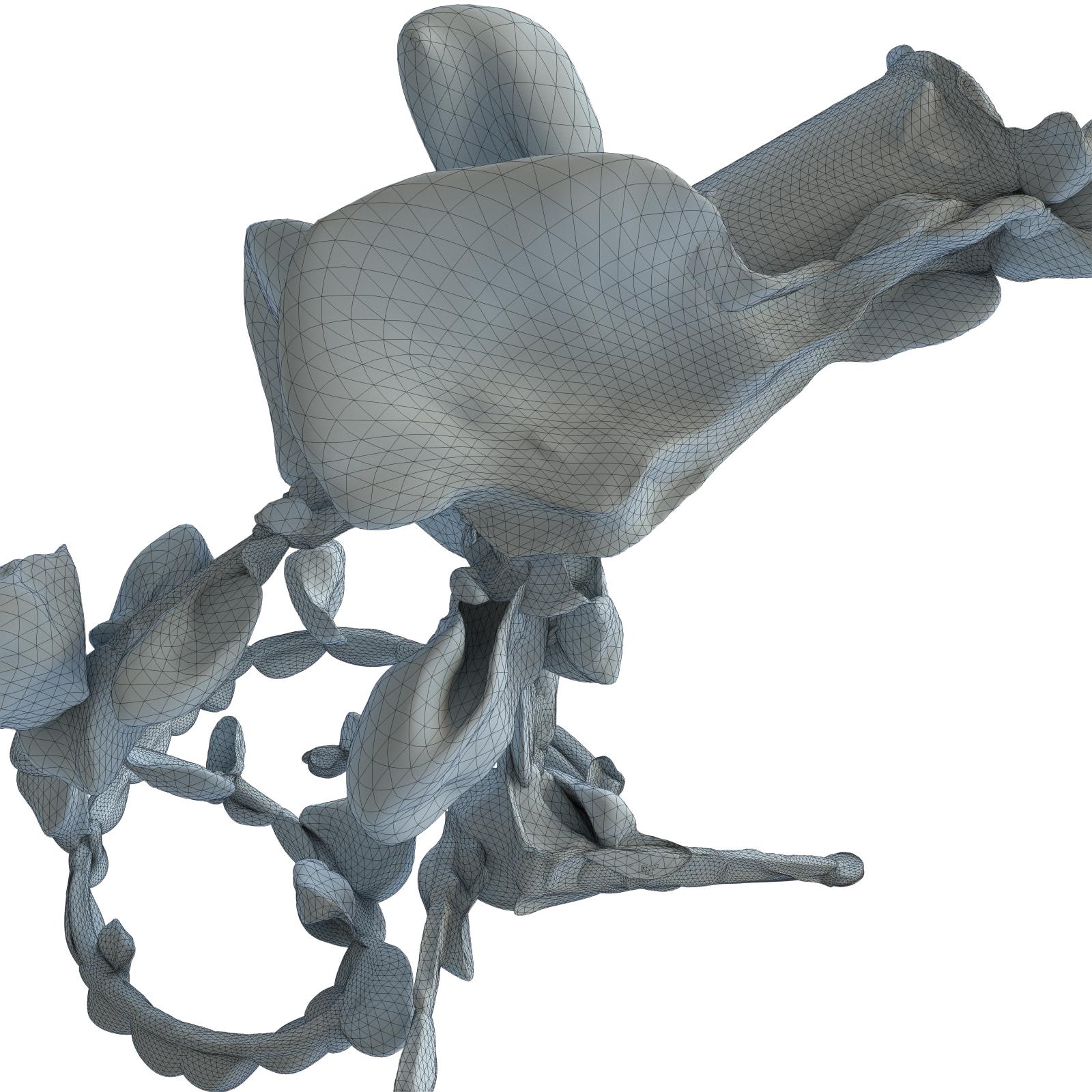} &
\includegraphics[width=0.28\textwidth]{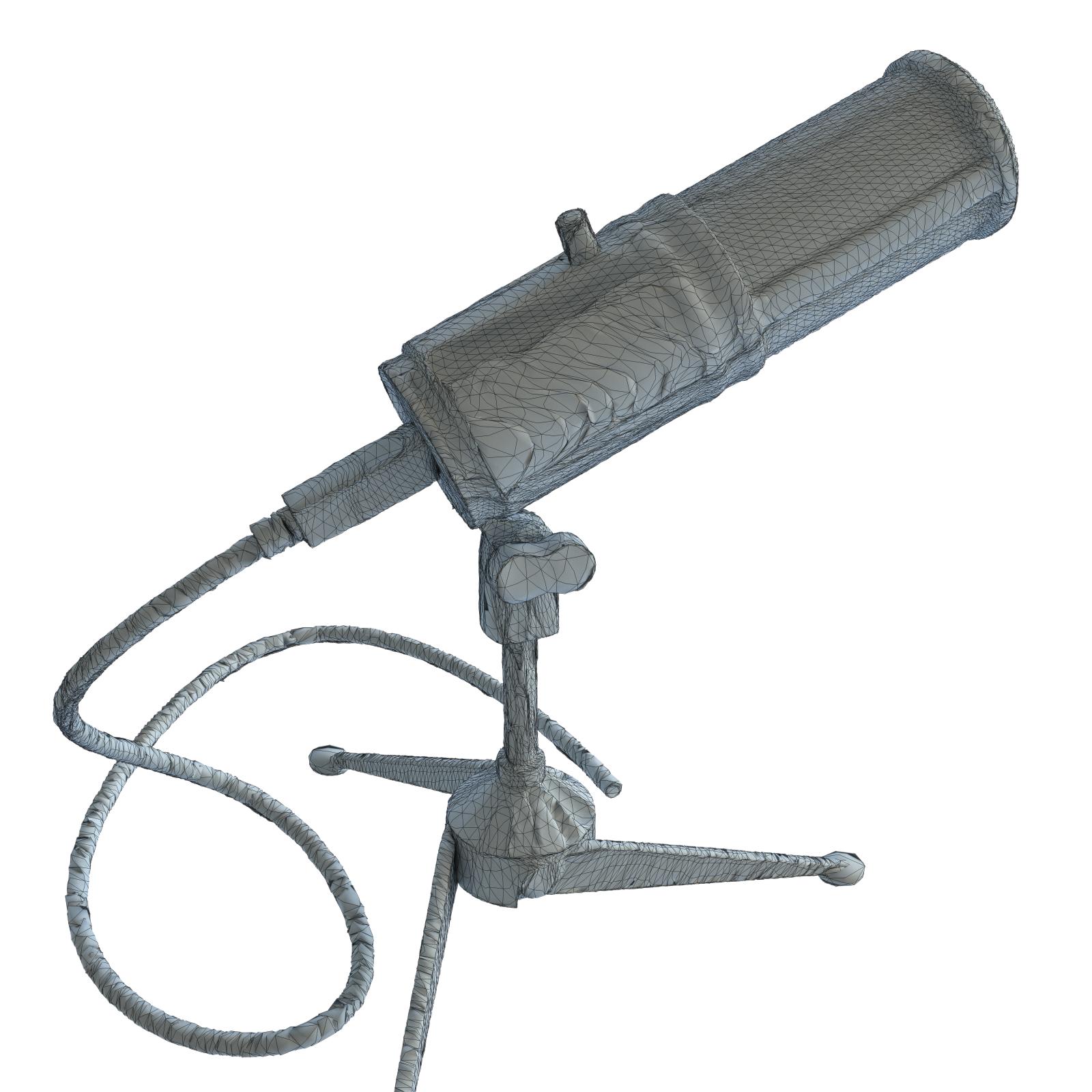} &
\includegraphics[width=0.28\textwidth]{figures_jpg/out_blender/mic000006_mi_w} \\
\end{tabular}
\endgroup
\caption{\textbf{Blender dataset~\cite{mildenhall2020nerf}, Ablation, mesh quality}. Ablation study of our method (EdgeGrad) on the blender dataset, test views. The first column is complete absence of gradients due to discontinuity, only the smooth component. Second is EdgeGrad, but without intersections handling. Third is full variant of EdgeGrad (our).}
    \label{f-blender_mesh_3}
\end{figure*}

\begin{figure*}
\centering\footnotesize
\begingroup
\renewcommand{\arraystretch}{0.}
\setlength{\tabcolsep}{0pt}
\begin{tabular}{ccc}
Ground truth image & Nvdiffrast~\cite{laine2020modular} & EdgeGrad (our) \\
\midrule
\includegraphics[width=0.29\linewidth]{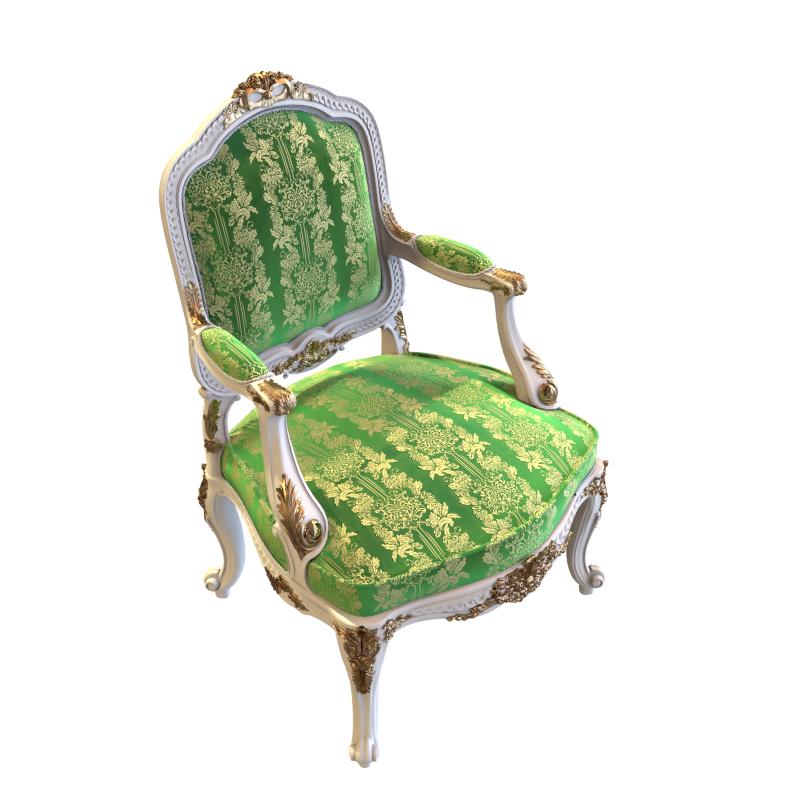} &
\includegraphics[width=0.29\textwidth]{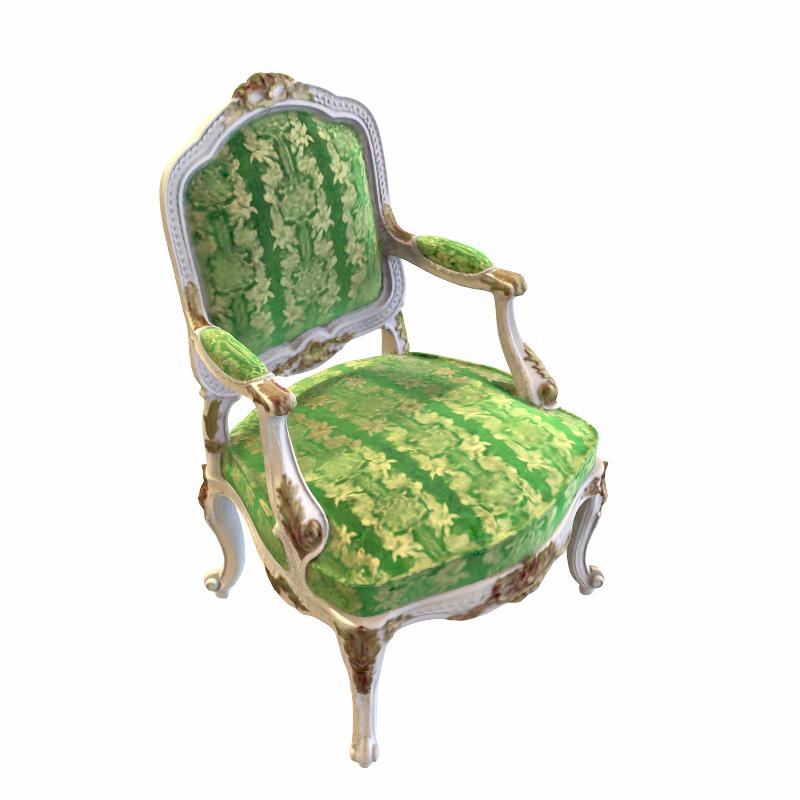} &
\includegraphics[width=0.29\textwidth]{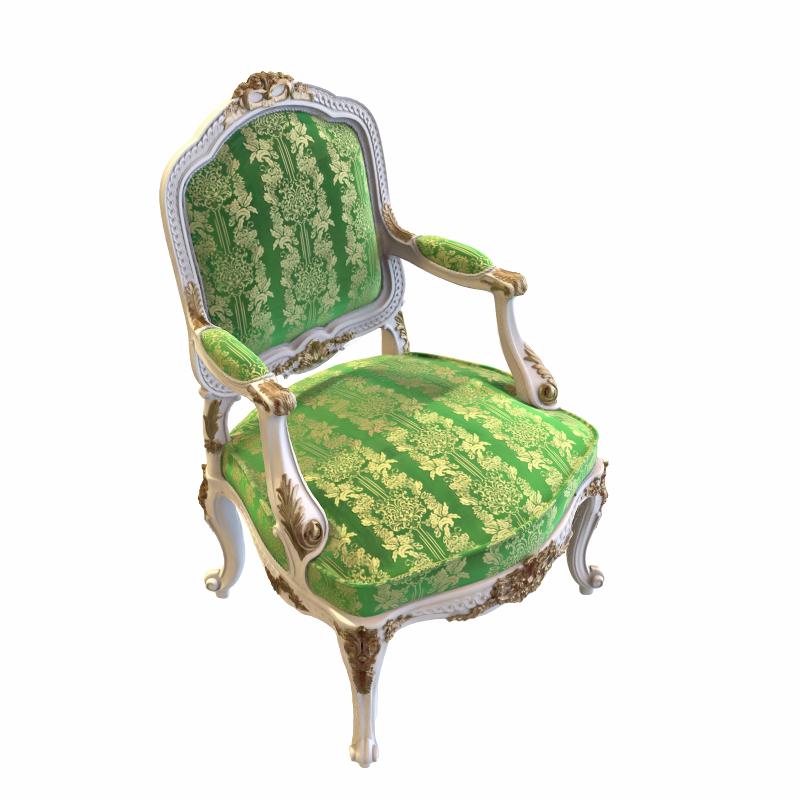} \\
\includegraphics[width=0.29\linewidth]{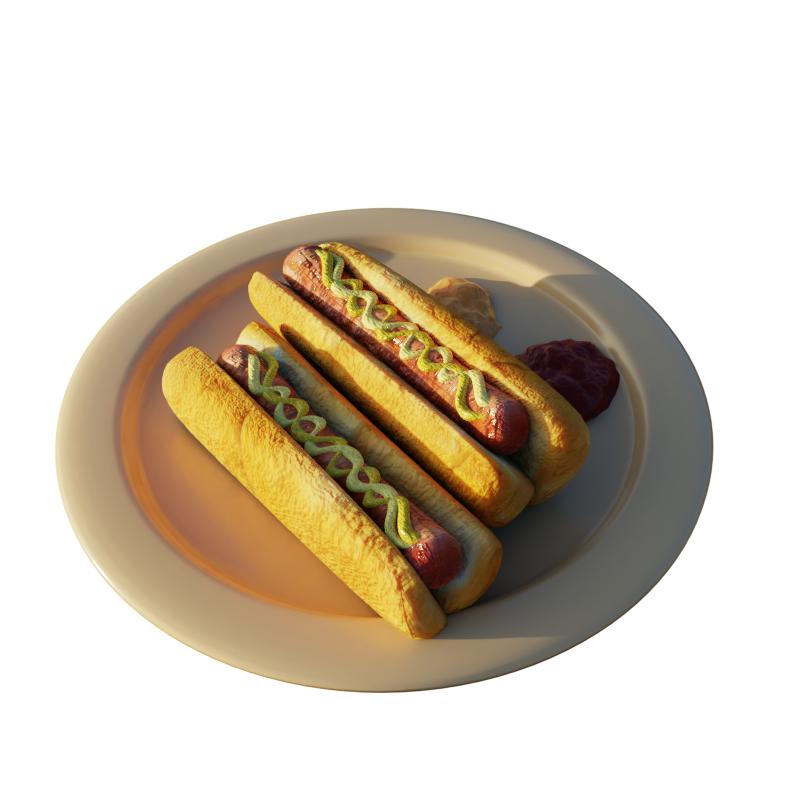} &
\includegraphics[width=0.29\textwidth]{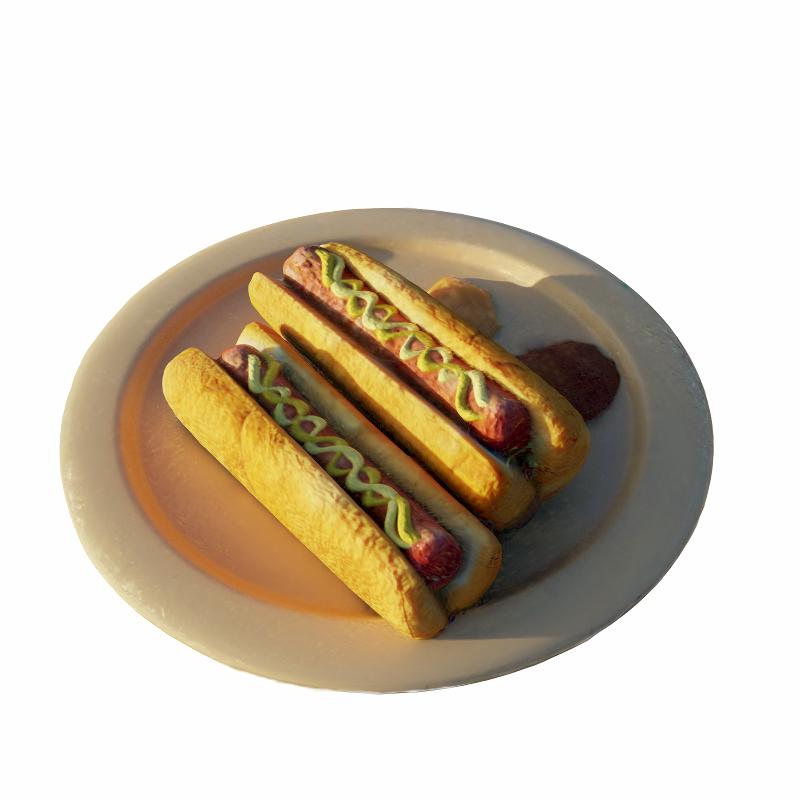} &
\includegraphics[width=0.29\textwidth]{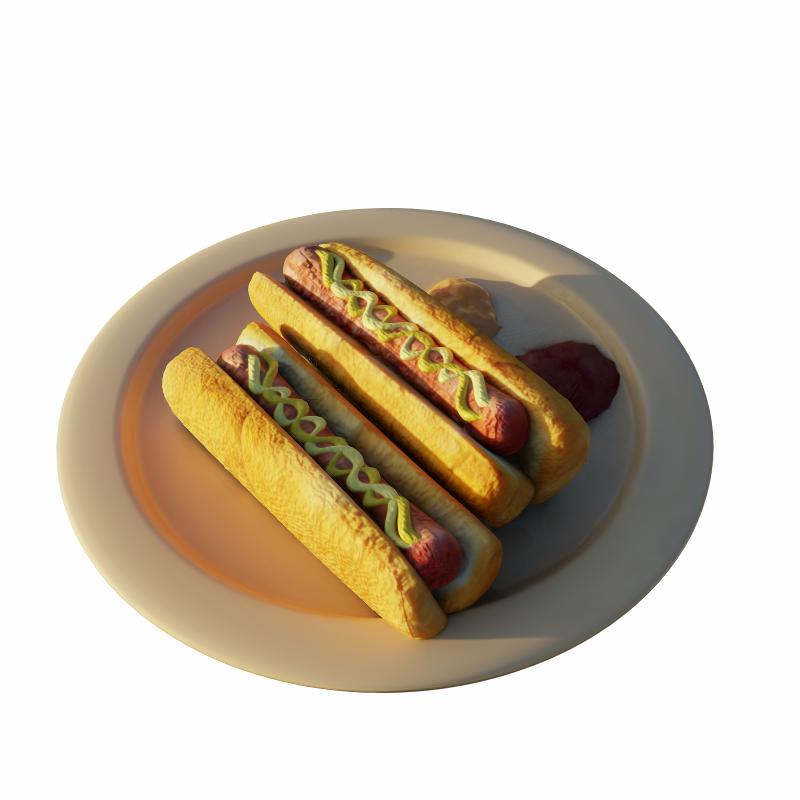} \\
\includegraphics[width=0.29\linewidth]{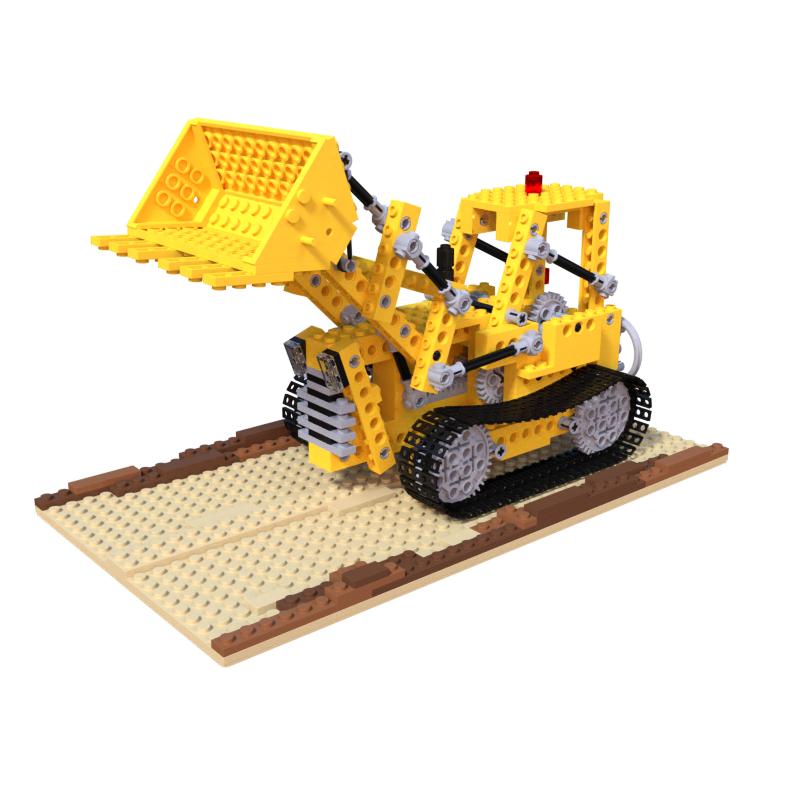} &
\includegraphics[width=0.29\textwidth]{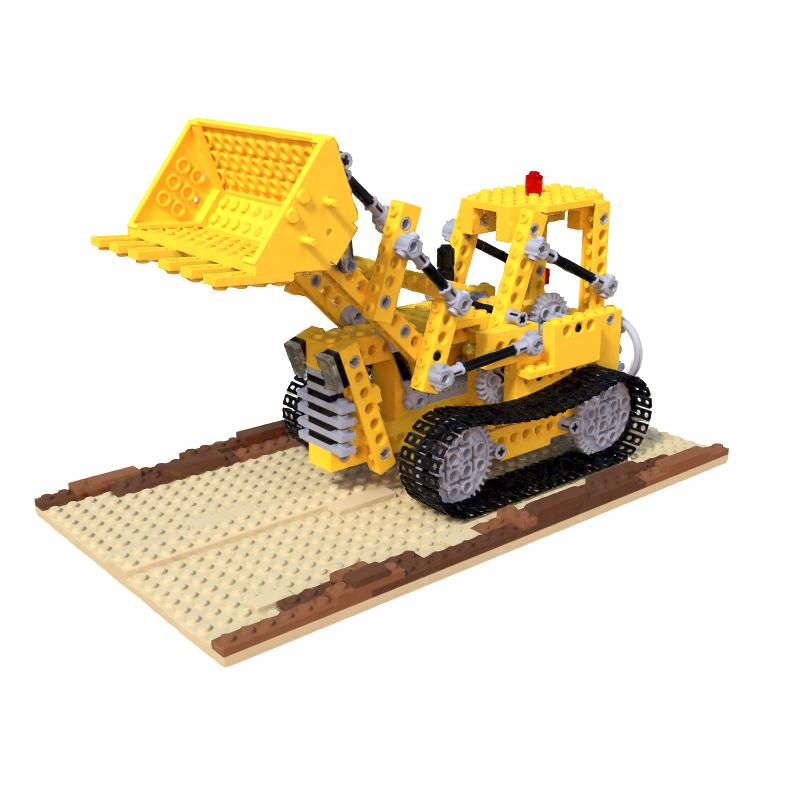} &
\includegraphics[width=0.29\textwidth]{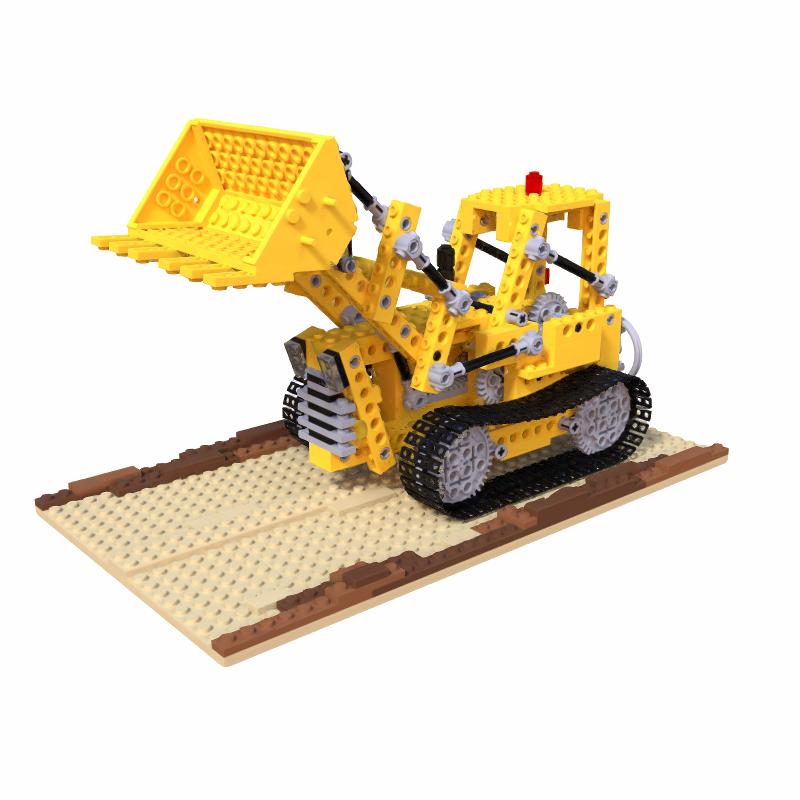} \\
\includegraphics[width=0.29\linewidth]{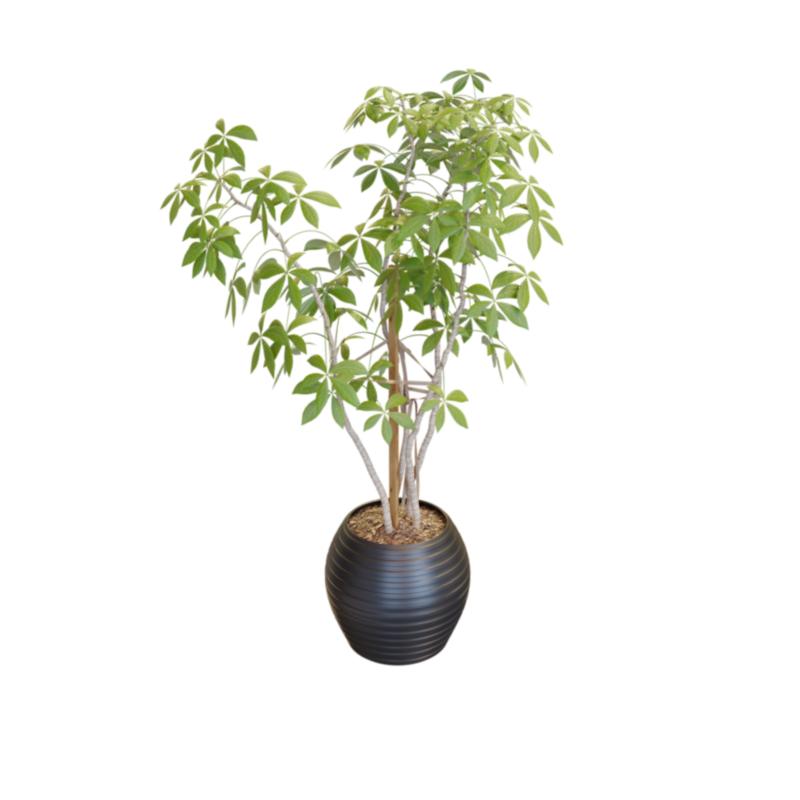} &
\includegraphics[width=0.29\textwidth]{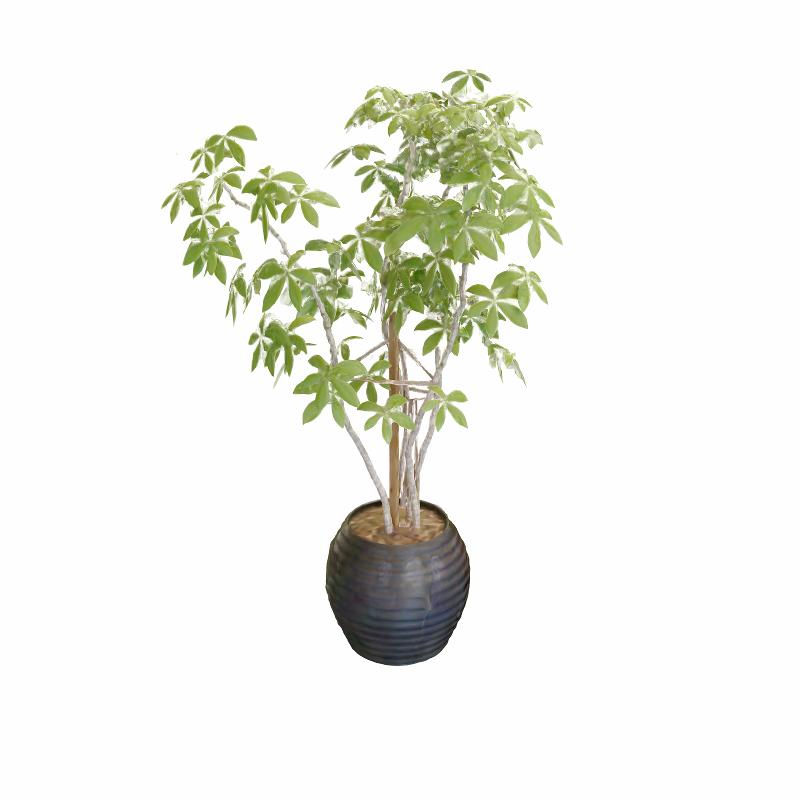} &
\includegraphics[width=0.29\textwidth]{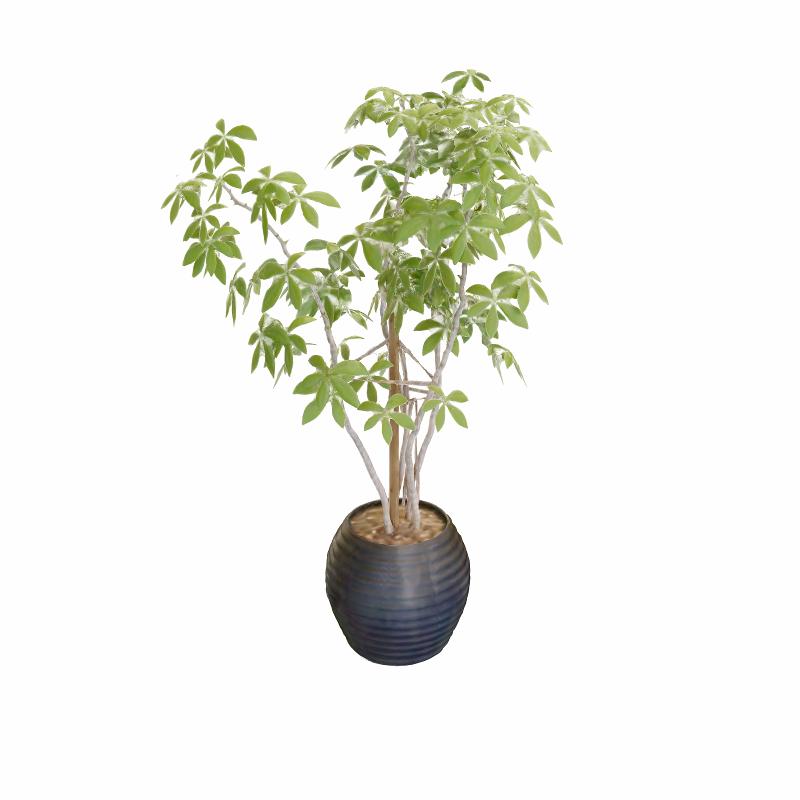} \\
\includegraphics[width=0.29\linewidth]{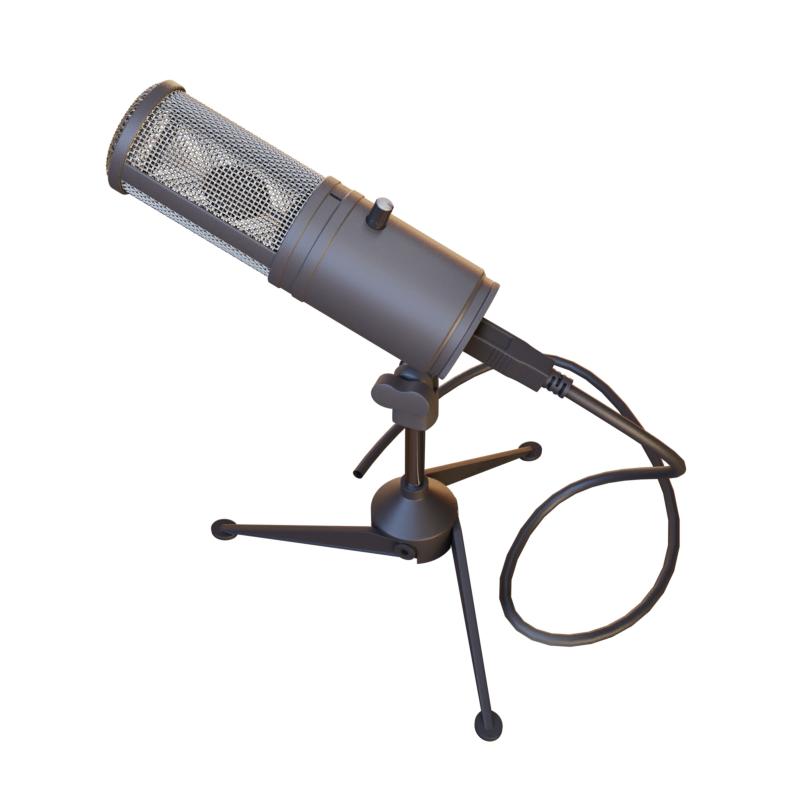} &
\includegraphics[width=0.29\textwidth]{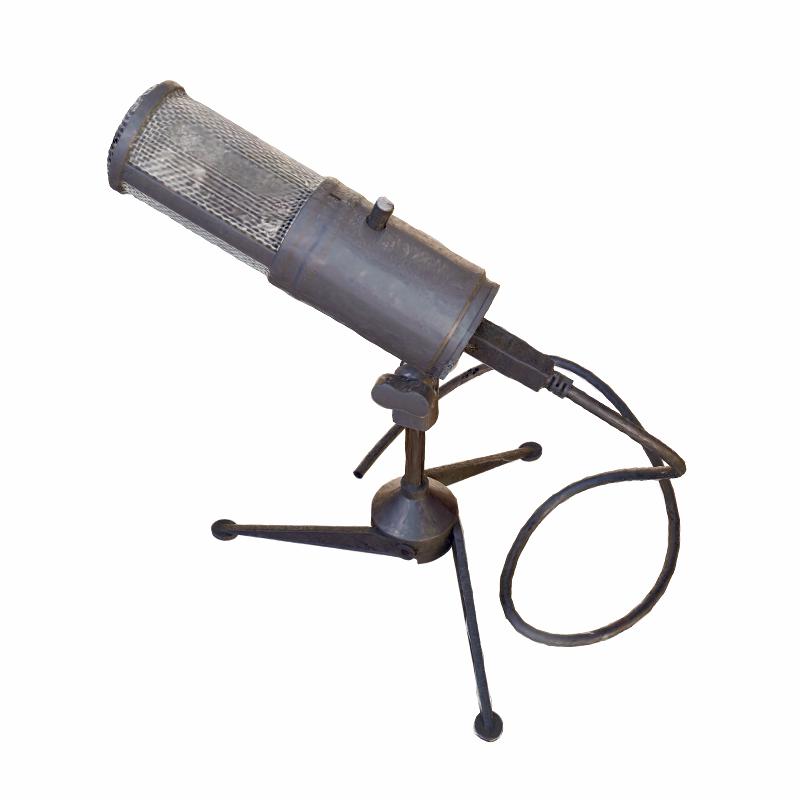} &
\includegraphics[width=0.29\textwidth]{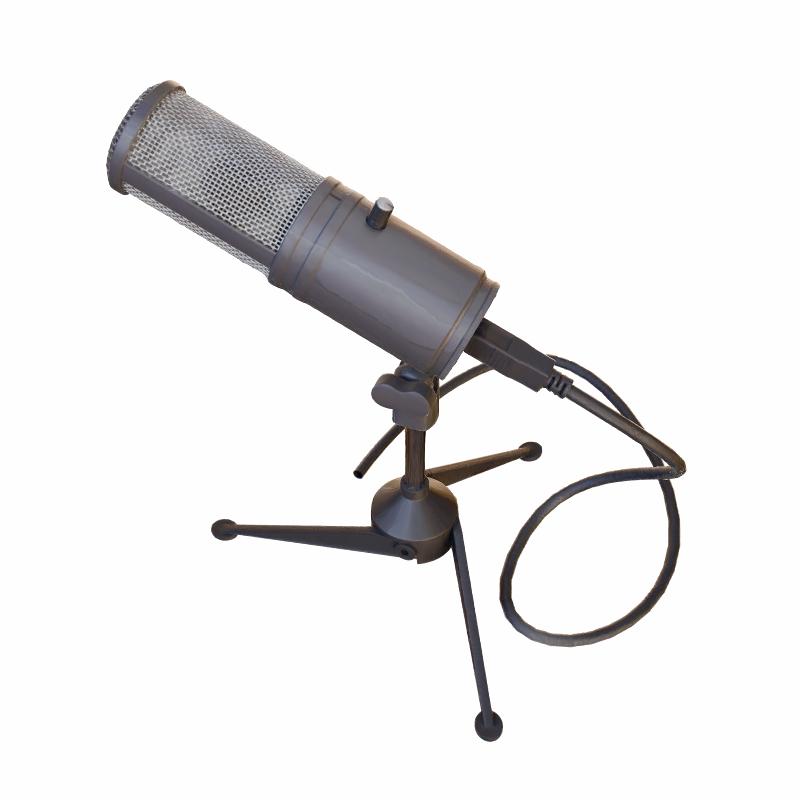} \\
\end{tabular}
\endgroup
\caption{\textbf{Blender dataset~\cite{mildenhall2020nerf}, Render image quality}. Qualitative evaluation of reconstructed scene on the blender dataset, test views. The first column is ground truth image. The second column is Nvdiffrast~\cite{laine2020modular}, third column is our method EdgeGrad.}
    \label{f-blender_mesh_4}
\end{figure*}

\begin{figure*}
\centering\footnotesize
\begingroup
\renewcommand{\arraystretch}{0.}
\setlength{\tabcolsep}{0pt}
\begin{tabular}{ccc}
Initialization mesh & Nvdiffrast~\cite{laine2020modular} & EdgeGrad (our) \\
\midrule
\includegraphics[width=0.3\linewidth]{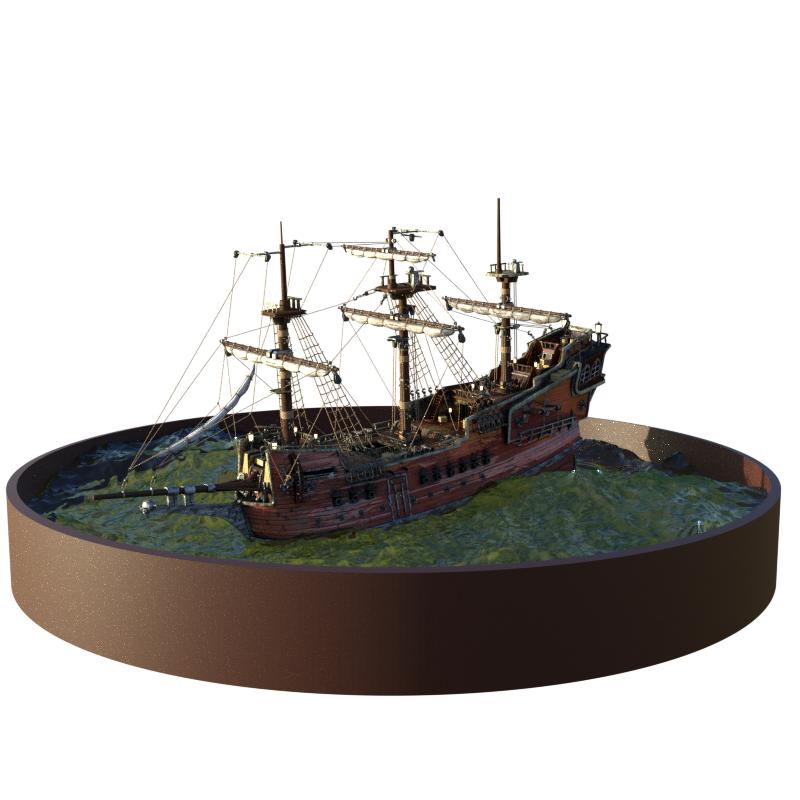} &
\includegraphics[width=0.3\textwidth]{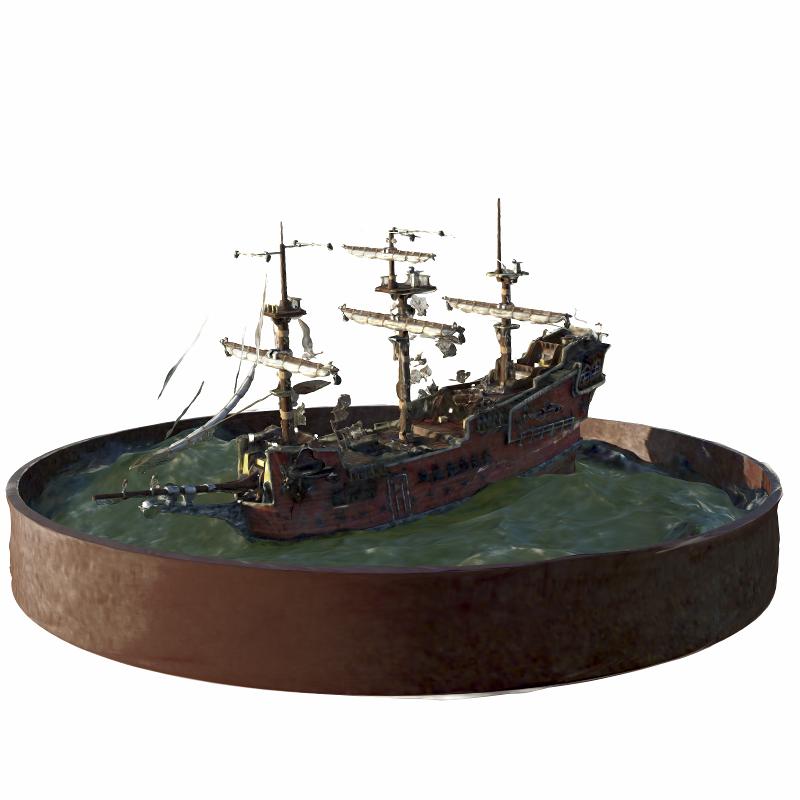} &
\includegraphics[width=0.3\textwidth]{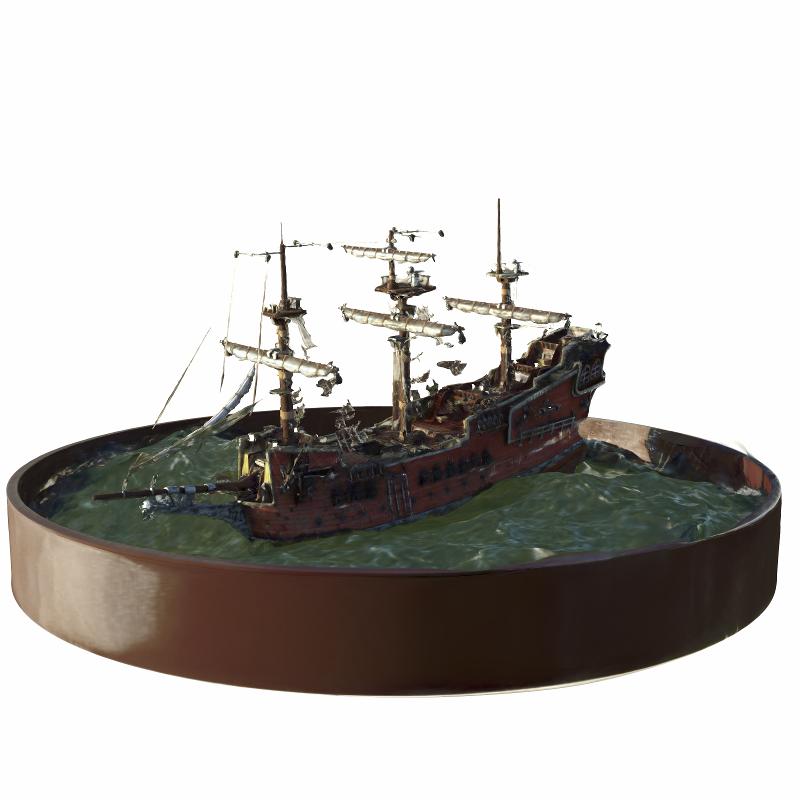} \\
\includegraphics[width=0.3\linewidth]{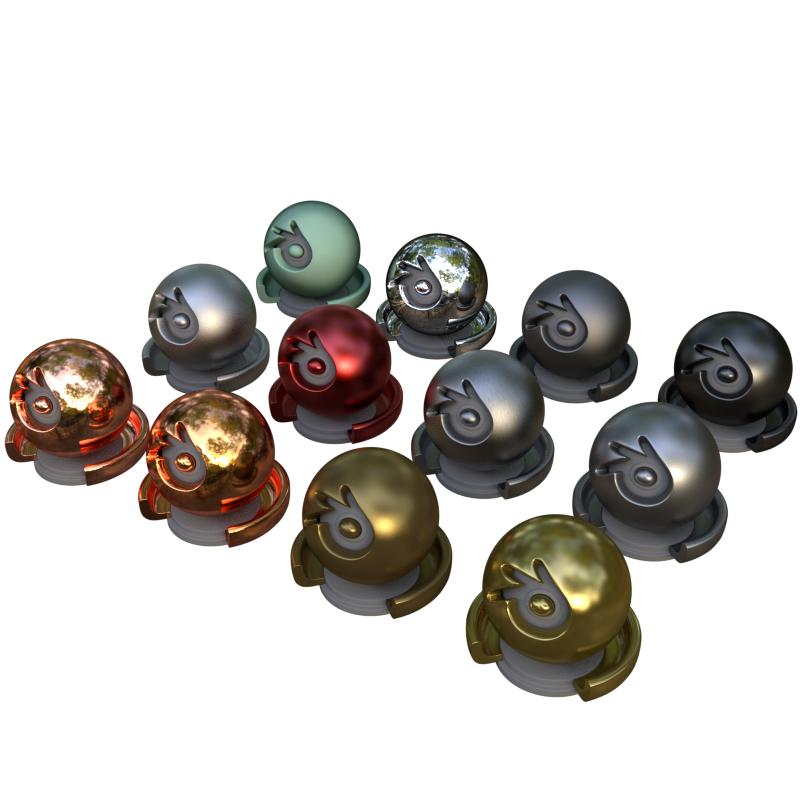} &
\includegraphics[width=0.3\textwidth]{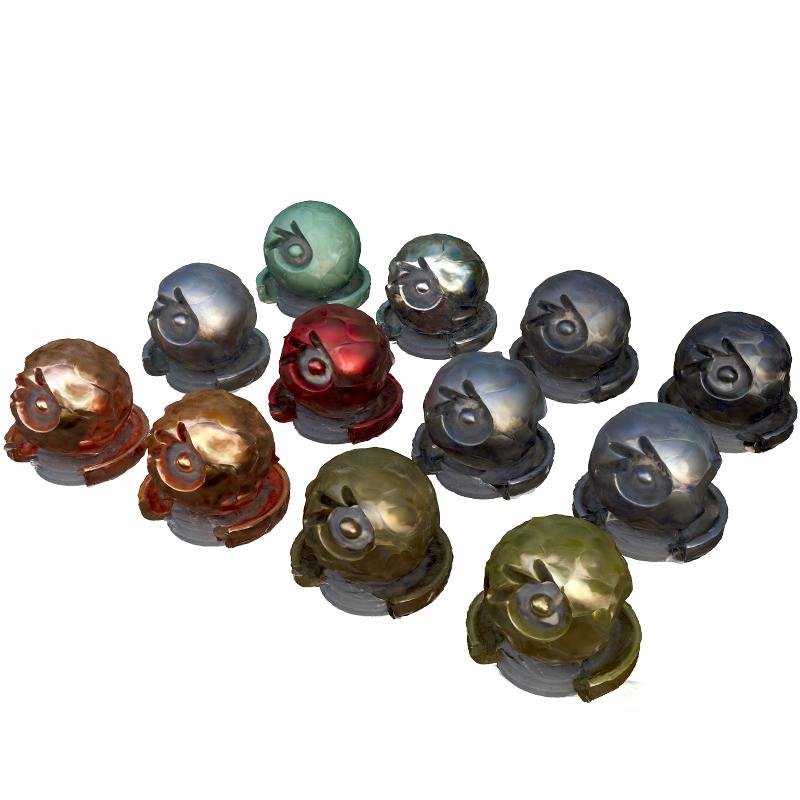} &
\includegraphics[width=0.3\textwidth]{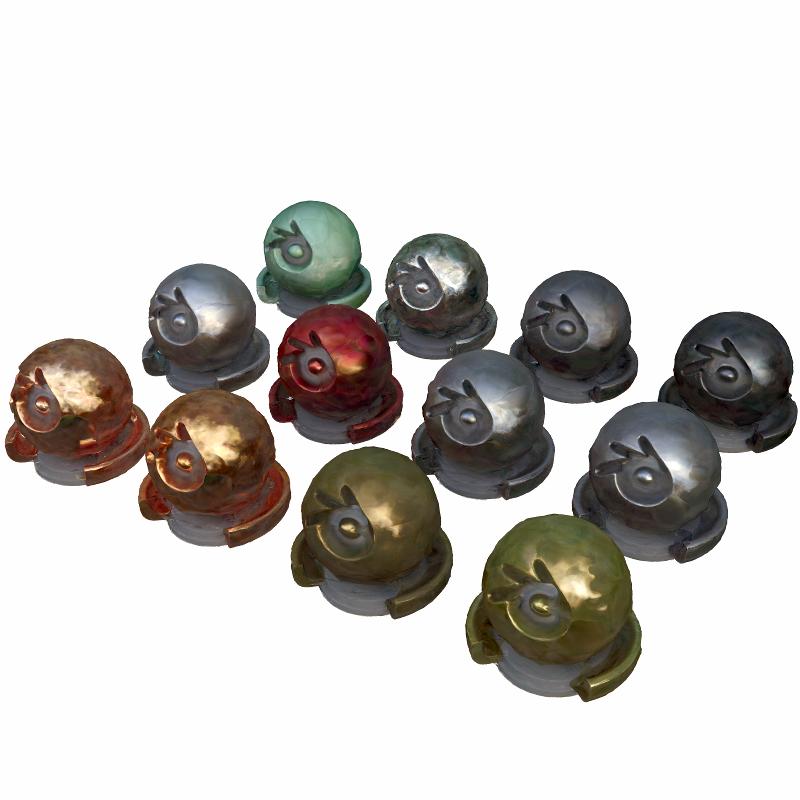} \\
\includegraphics[width=0.3\linewidth]{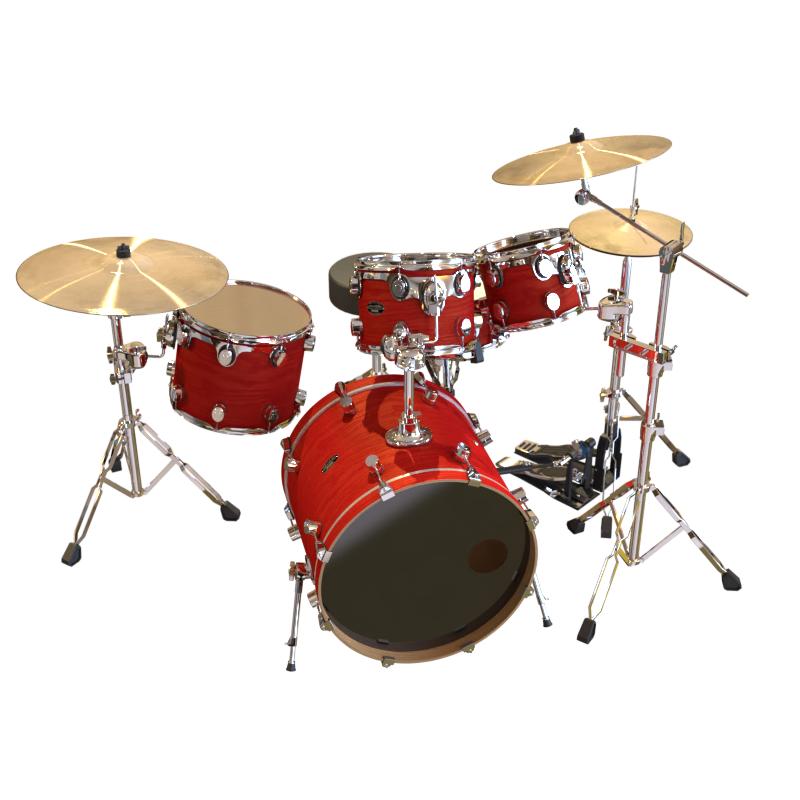} &
\includegraphics[width=0.3\textwidth]{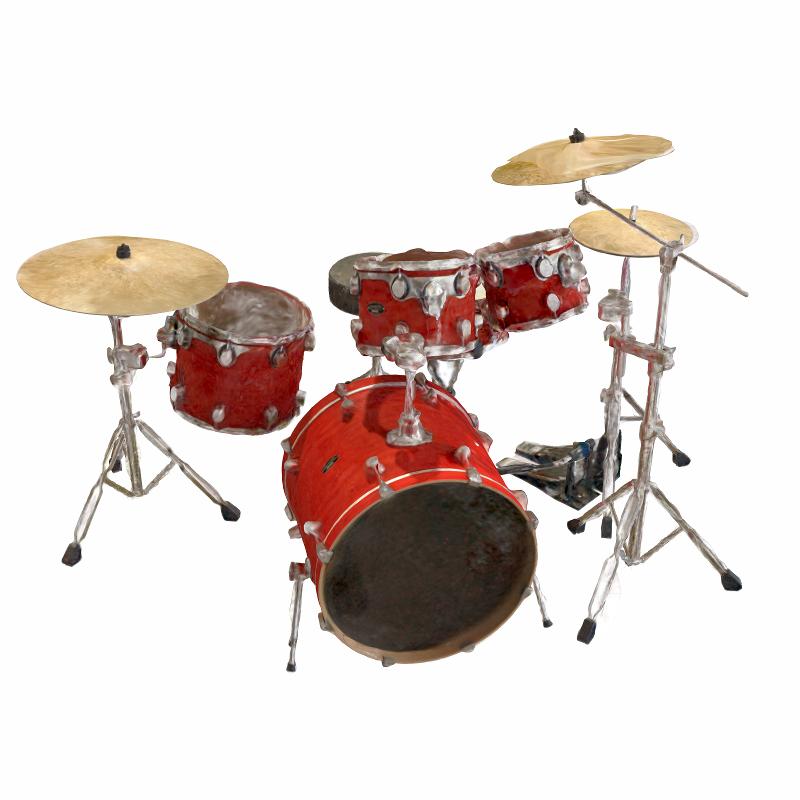} &
\includegraphics[width=0.3\textwidth]{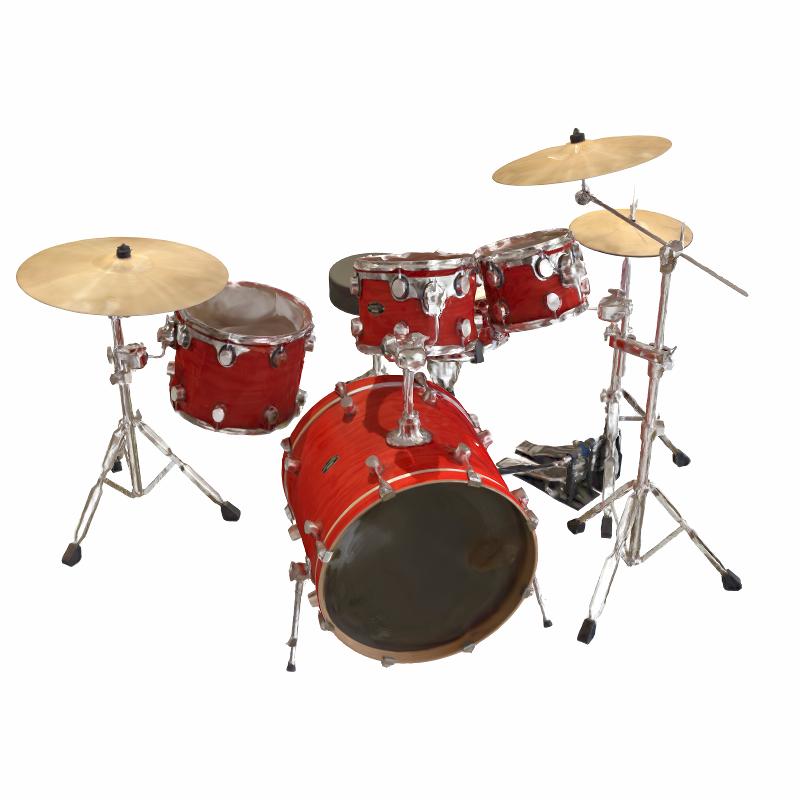} \\

\end{tabular}
\endgroup
\caption{\textbf{Blender dataset~\cite{mildenhall2020nerf}, Render quality (Continuation of Fig.~\ref{f-blender_mesh_4})}. Qualitative evaluation of reconstructed scene on the blender dataset, test views. The first column is ground truth image. The second column is Nvdiffrast~\cite{laine2020modular}, third column is our method EdgeGrad.}
    \label{f-blender_mesh_5}
\end{figure*}

Please refer to the Figures~\ref{f-blender_mesh_1}--\ref{f-blender_mesh_5} for qualitative evaluation on blender dataset~\cite{mildenhall2020nerf}.

\begin{figure*}
\centering\footnotesize
\begingroup
\renewcommand{\arraystretch}{0.}
\setlength{\tabcolsep}{0pt}
\begin{tabular}{ccc}
Ground truth image & Nvdiffrast~\cite{laine2020modular} & EdgeGrad (our) \\
\midrule
\includegraphics[width=0.34\linewidth]{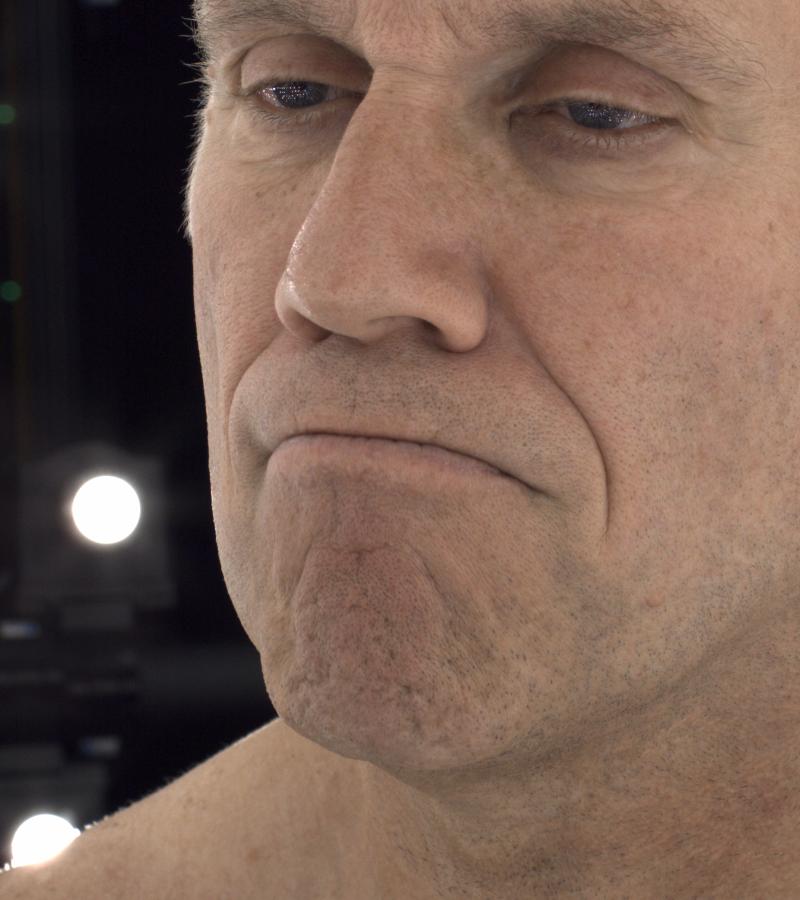} &
\includegraphics[width=0.34\textwidth]{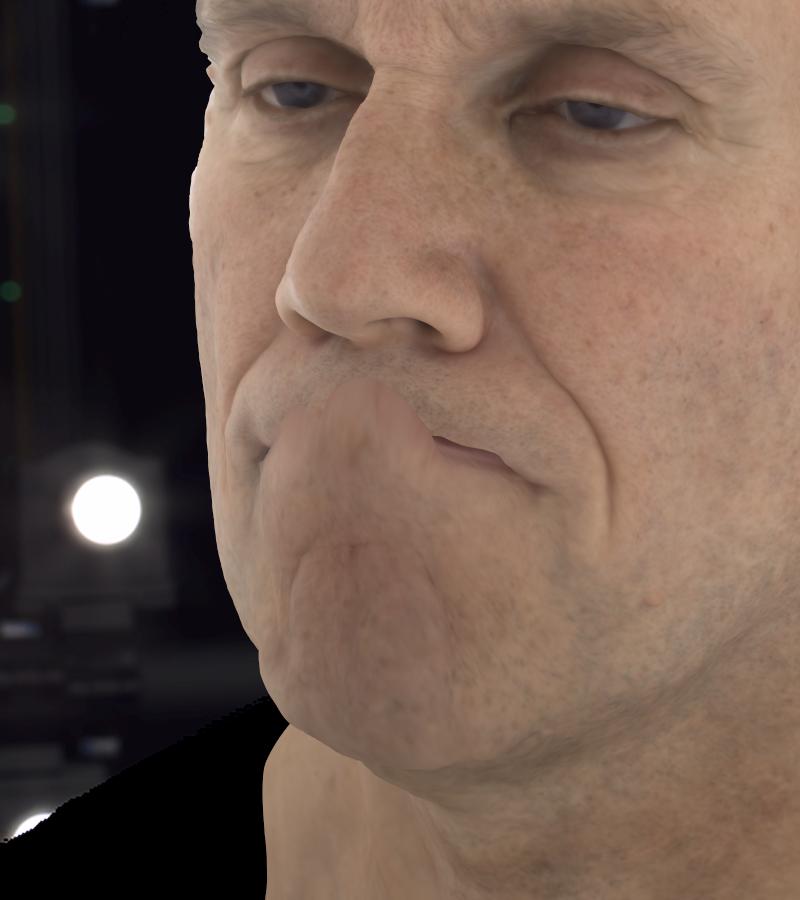} &
\includegraphics[width=0.34\textwidth]{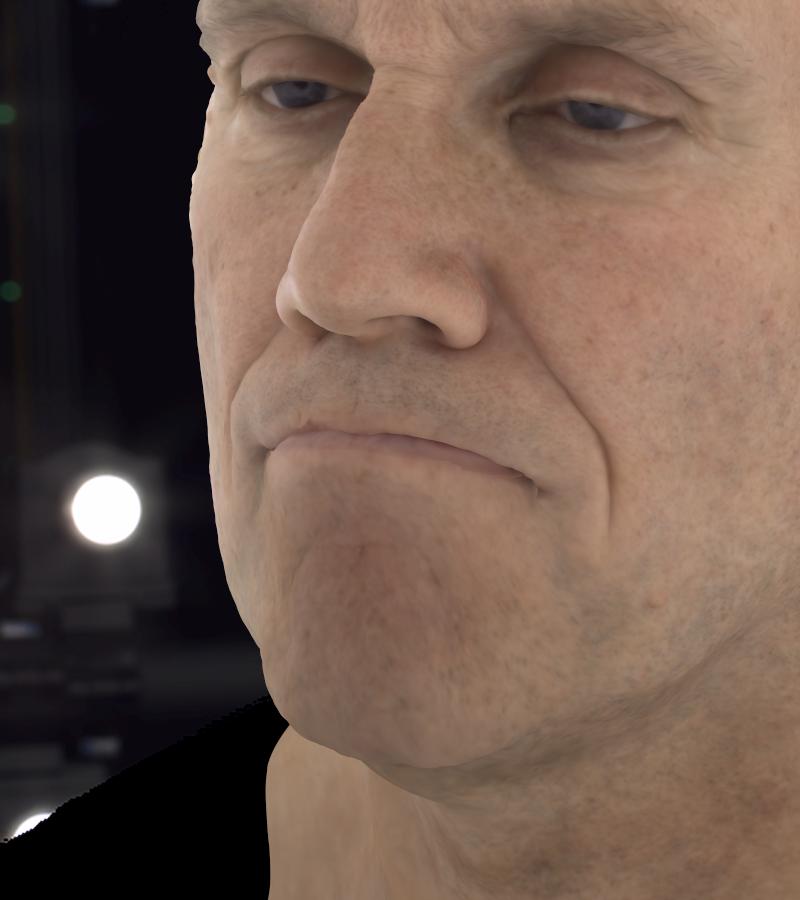} \\
\includegraphics[width=0.34\linewidth]{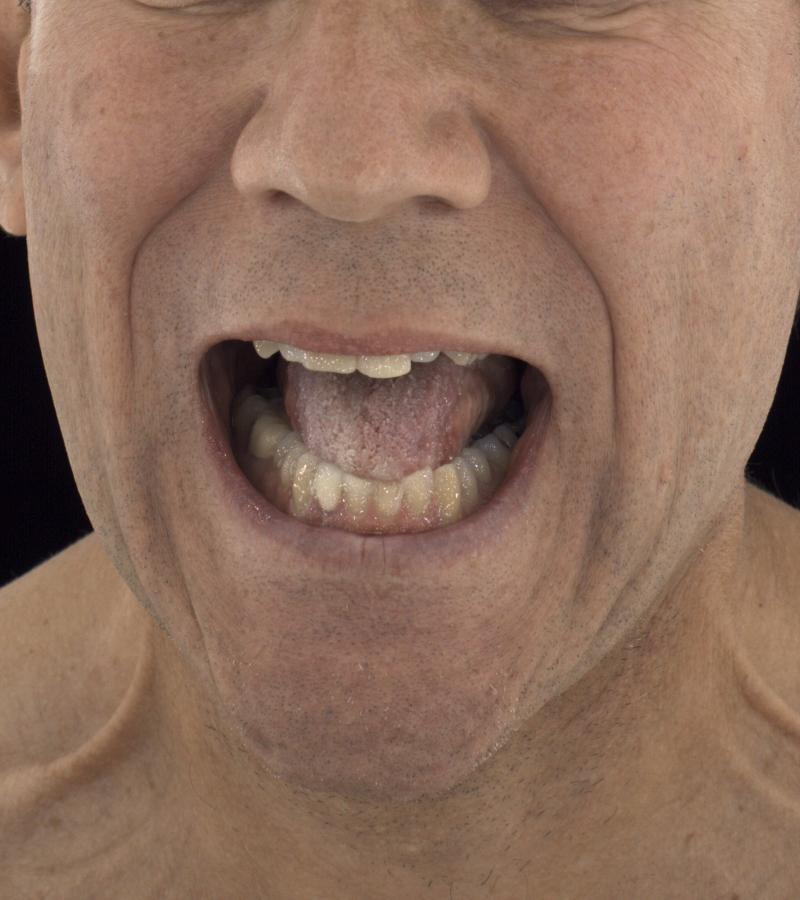} &
\includegraphics[width=0.34\textwidth]{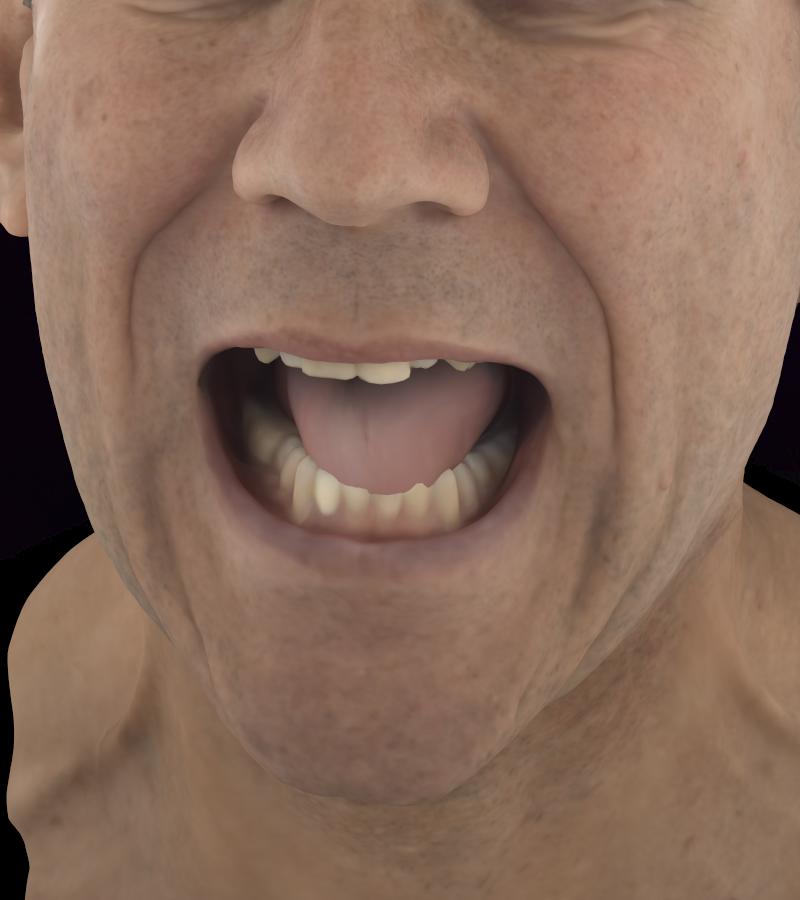} &
\includegraphics[width=0.34\textwidth]{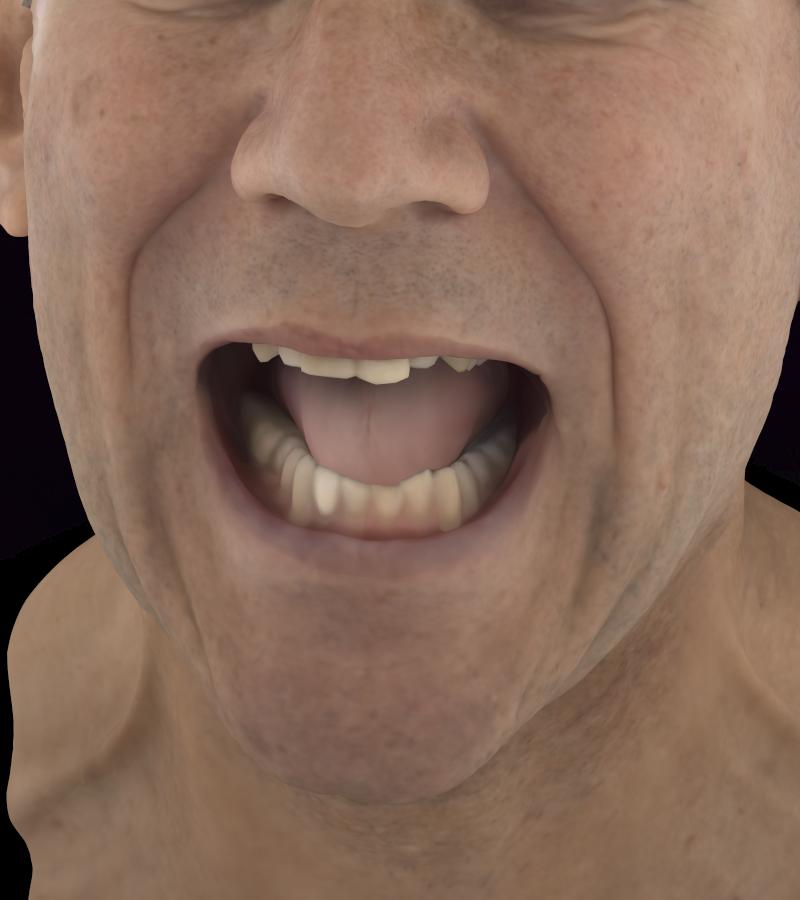} \\
\includegraphics[width=0.34\linewidth]{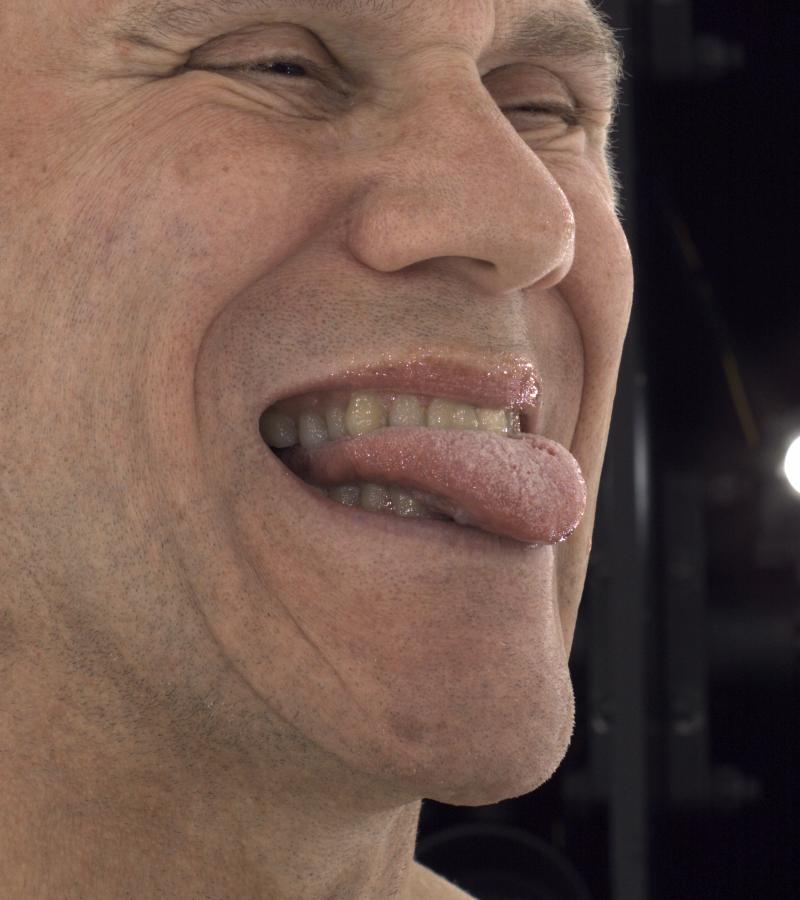} &
\includegraphics[width=0.34\textwidth]{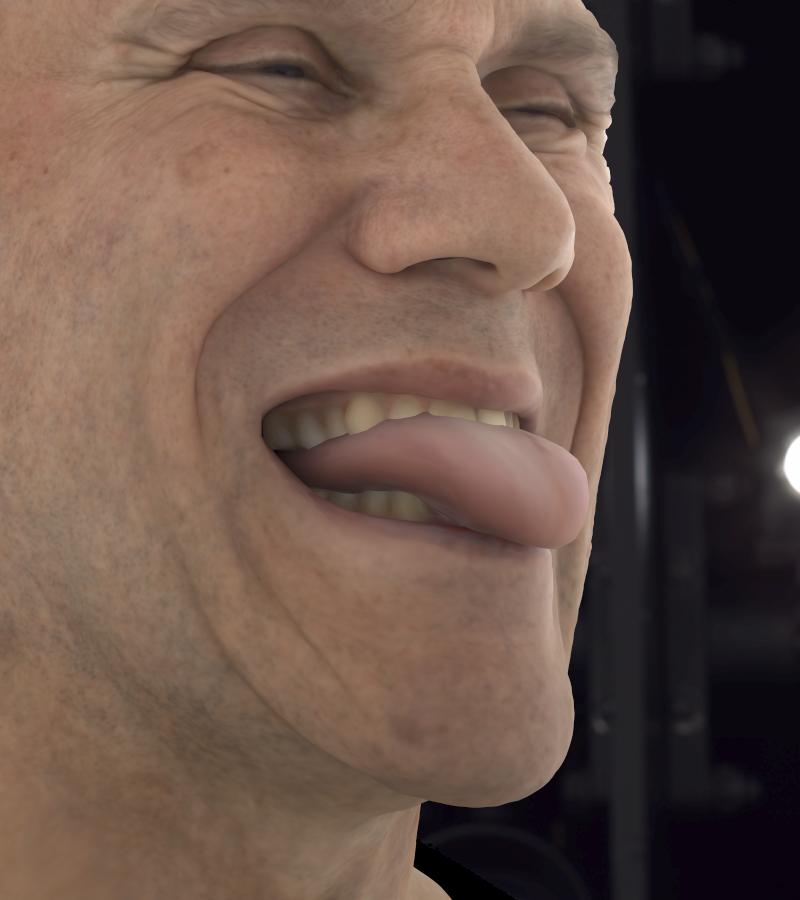} &
\includegraphics[width=0.34\textwidth]{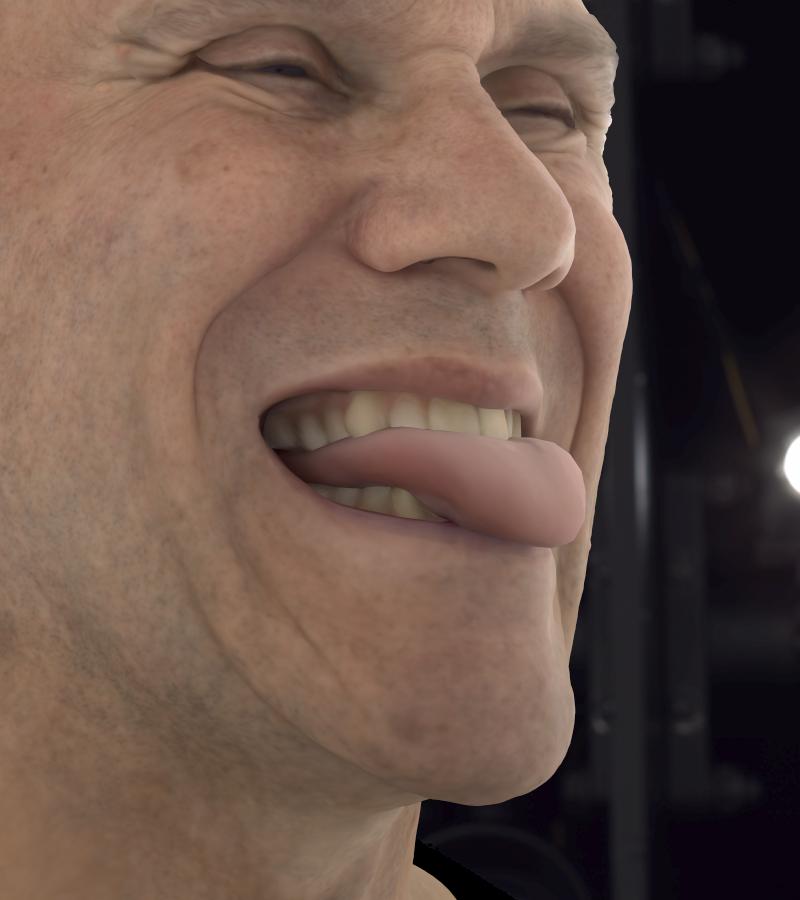}
\end{tabular}
\endgroup
\caption{\textbf{Dynamic head scene}. Qualitative evaluation of reconstructed dynamic head scene. The first column is ground truth image. The second column is Nvdiffrast~\cite{laine2020modular}, third column is our method EdgeGrad.}
    \label{f-ca1}
\end{figure*}

\begin{figure*}
\centering\footnotesize
\begingroup
\renewcommand{\arraystretch}{0.}
\setlength{\tabcolsep}{0pt}
\begin{tabular}{ccc}
Ground truth image & Nvdiffrast~\cite{laine2020modular} & EdgeGrad (our) \\
\midrule
\includegraphics[width=0.34\linewidth]{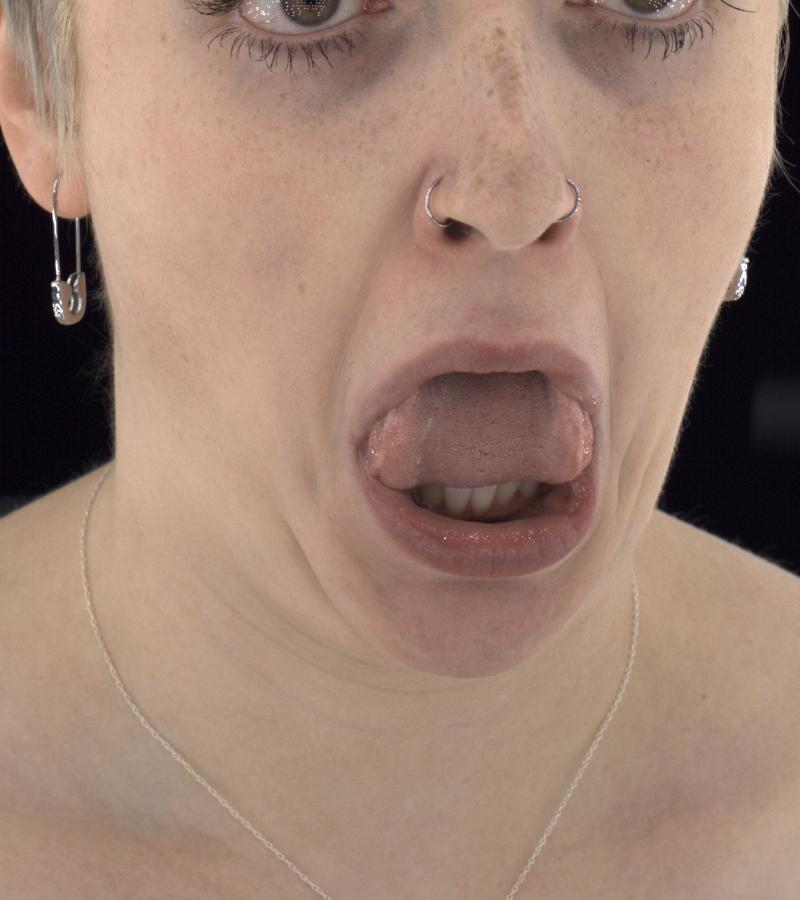} &
\includegraphics[width=0.34\textwidth]{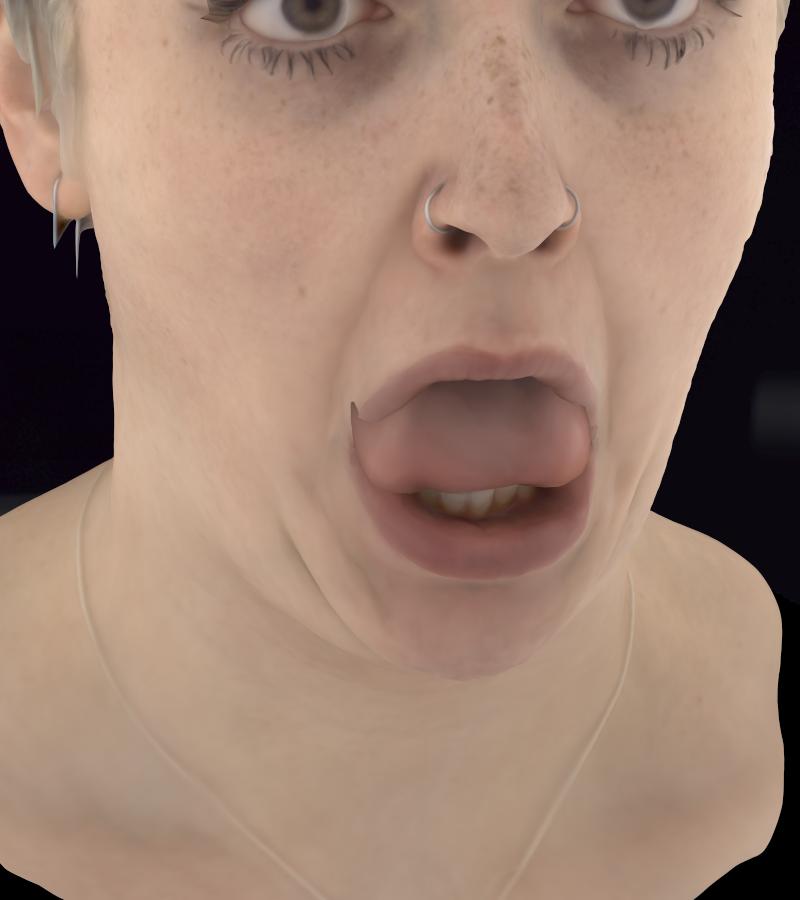} &
\includegraphics[width=0.34\textwidth]{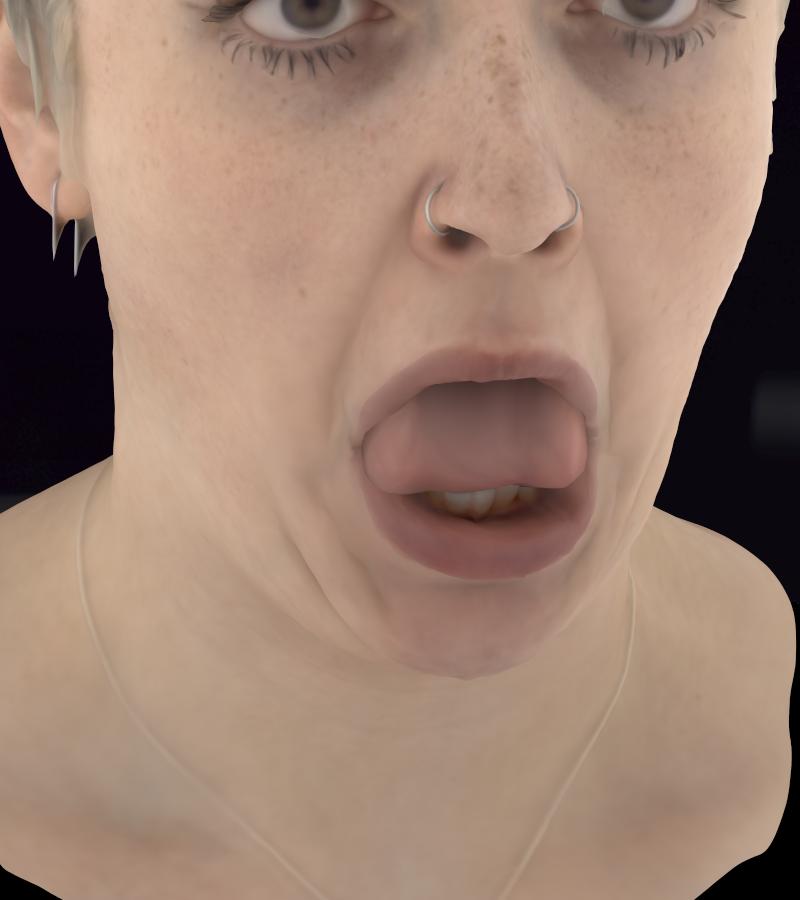} \\
\includegraphics[width=0.34\linewidth]{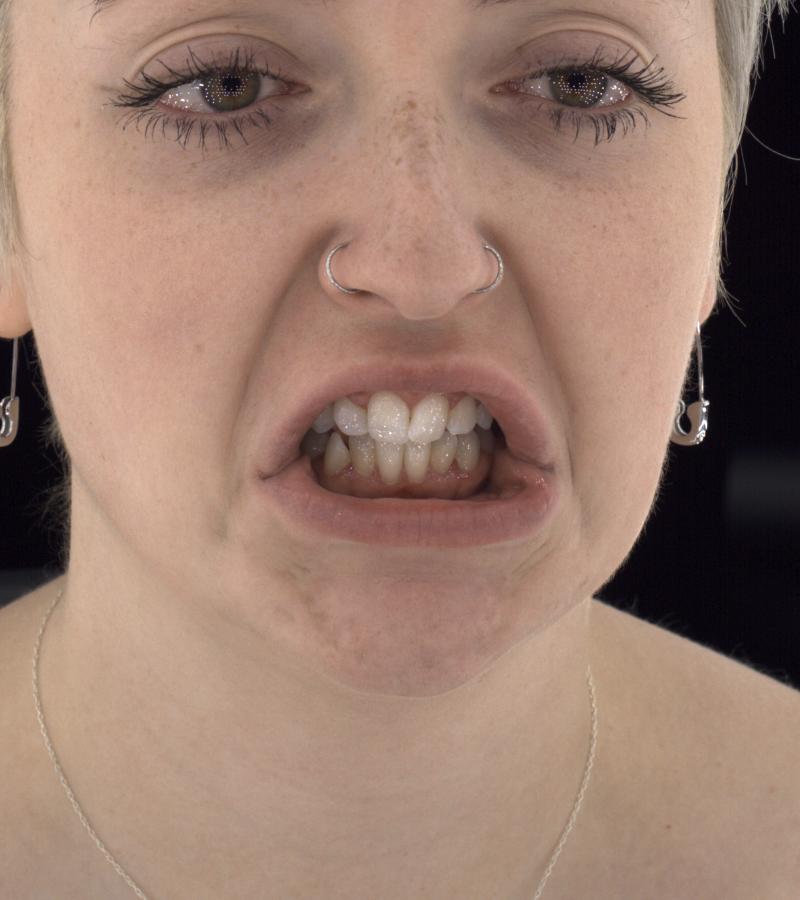} &
\includegraphics[width=0.34\textwidth]{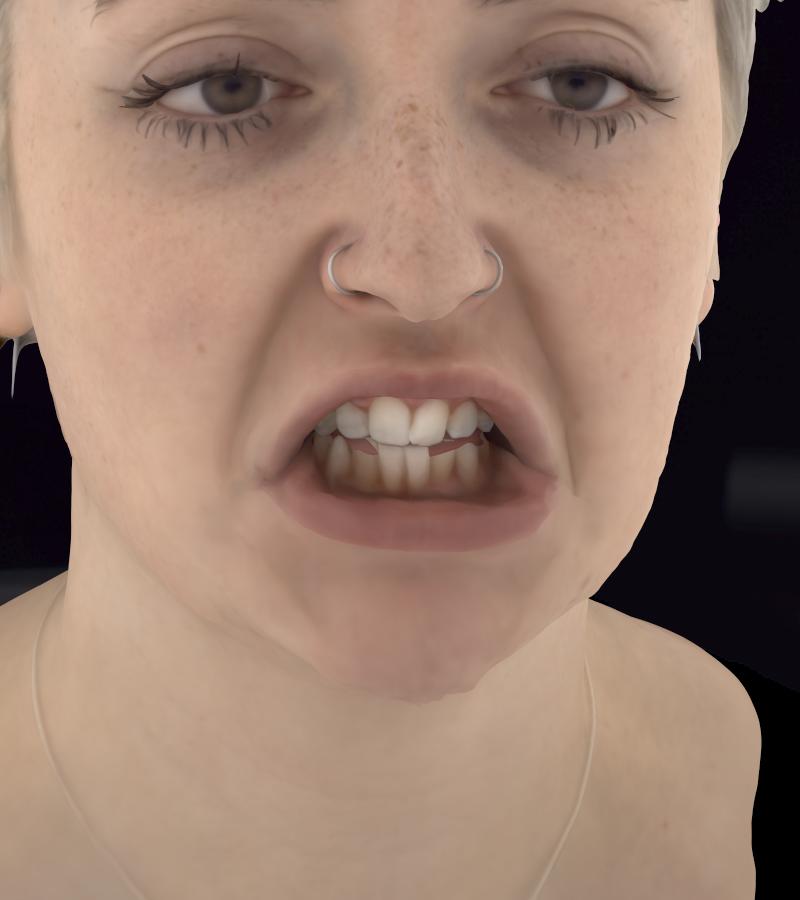} &
\includegraphics[width=0.34\textwidth]{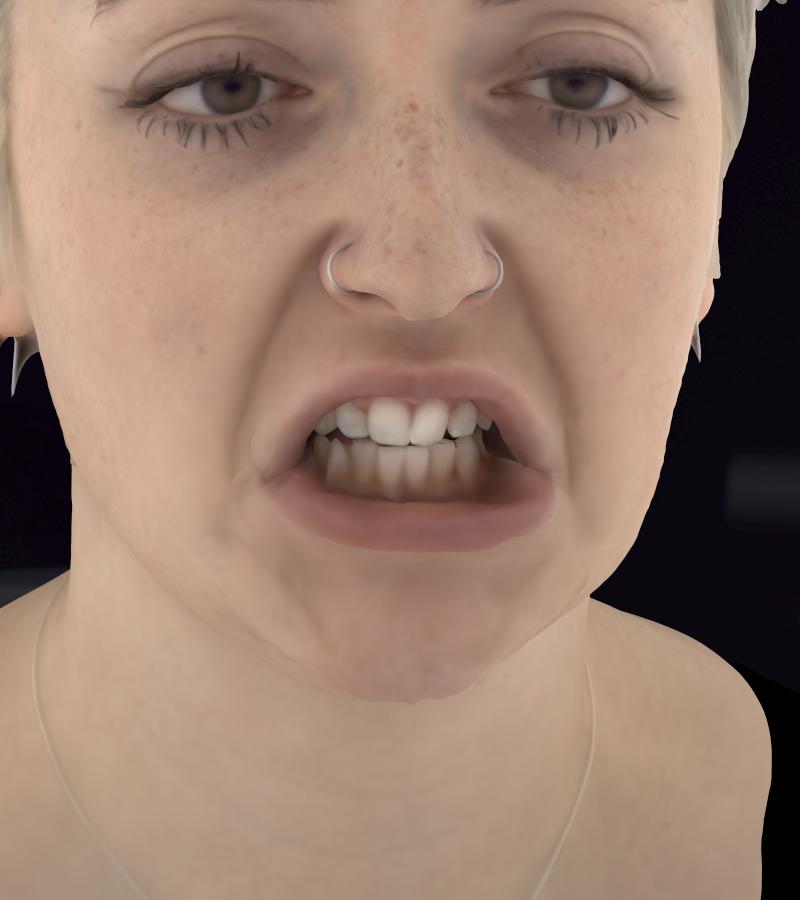} \\
\includegraphics[width=0.34\linewidth]{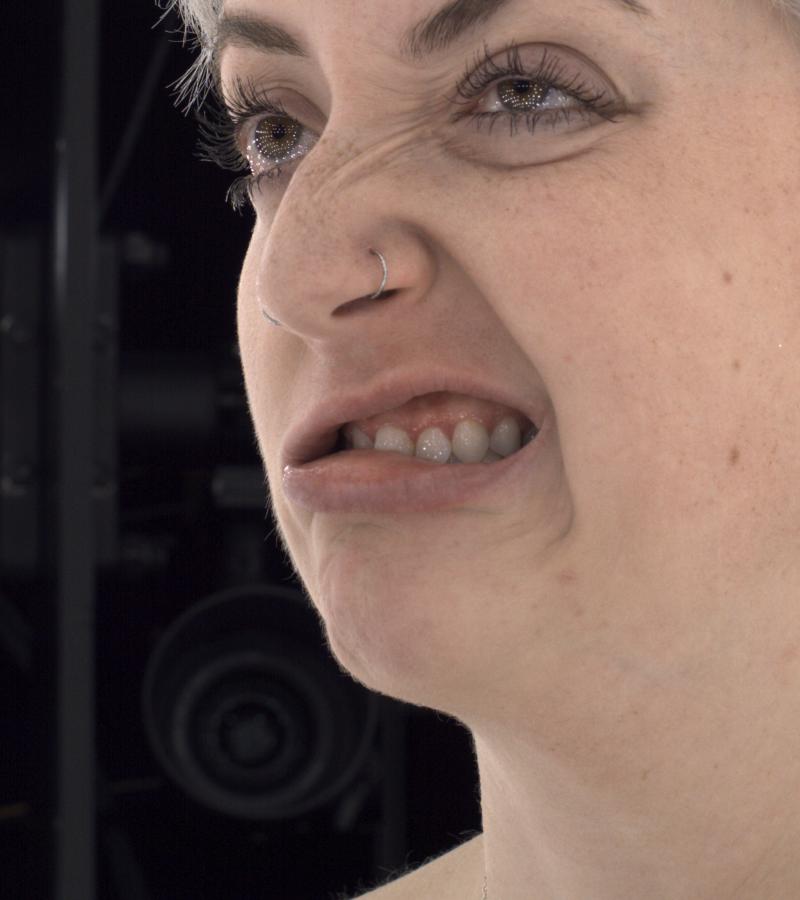} &
\includegraphics[width=0.34\textwidth]{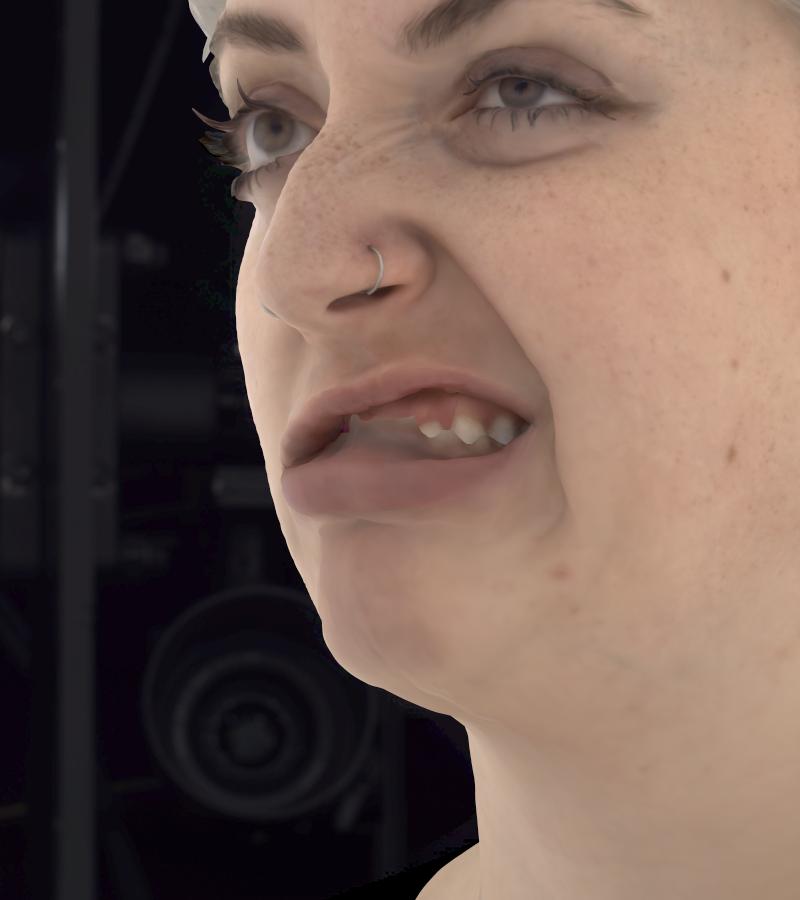} &
\includegraphics[width=0.34\textwidth]{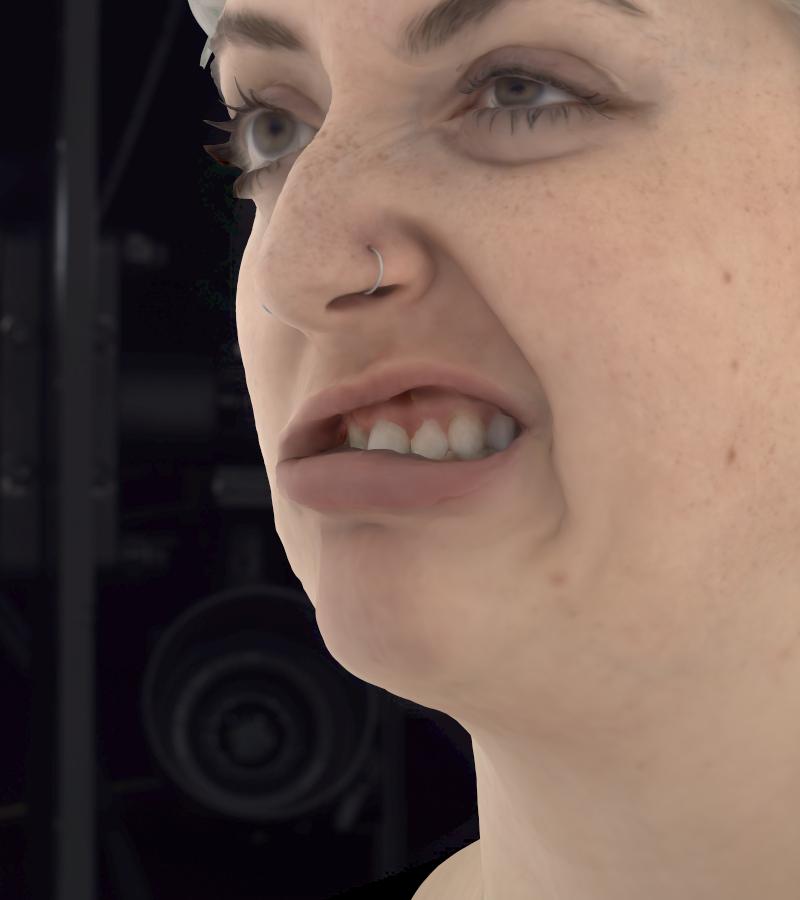} \\
\end{tabular}
\endgroup
\caption{\textbf{Dynamic head scene. (Continuation of Fig.~\ref{f-ca1})}. Qualitative evaluation of reconstructed dynamic head scene. The first column is ground truth image. The second column is Nvdiffrast~\cite{laine2020modular}, third column is our method EdgeGrad.}
    \label{f-ca2}
\end{figure*}

\begin{figure*}
\centering\footnotesize
\begingroup
\renewcommand{\arraystretch}{0.}
\setlength{\tabcolsep}{0pt}
\begin{tabular}{cc}
 Nvdiffrast~\cite{laine2020modular} & EdgeGrad (our) \\
\midrule
\includegraphics[width=0.37\textwidth]{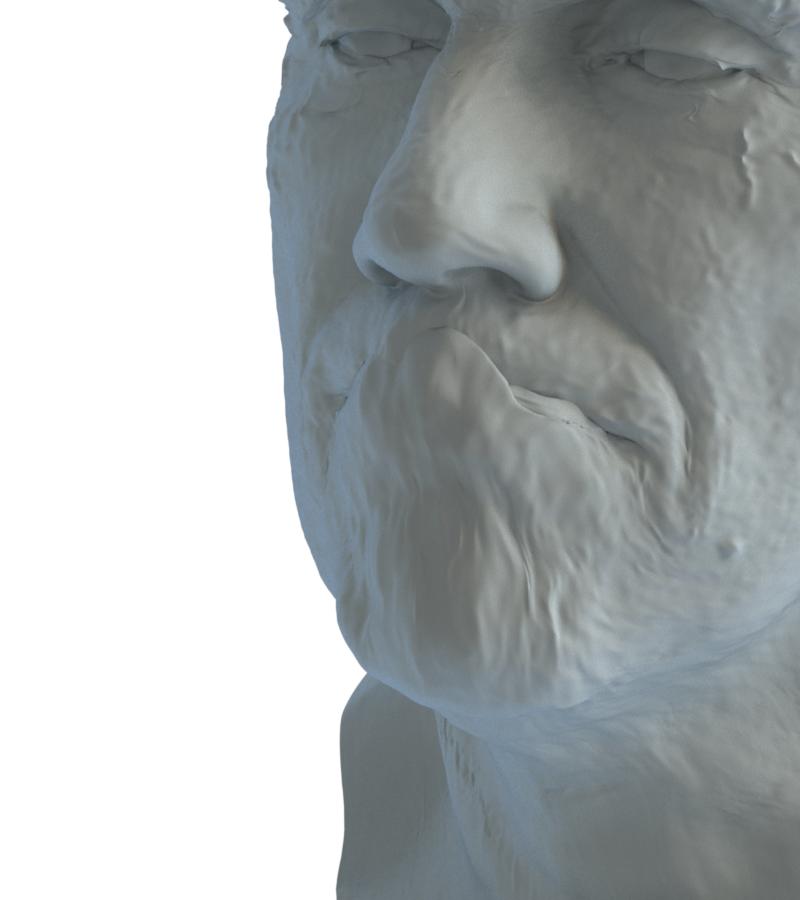} &
\includegraphics[width=0.37\textwidth]{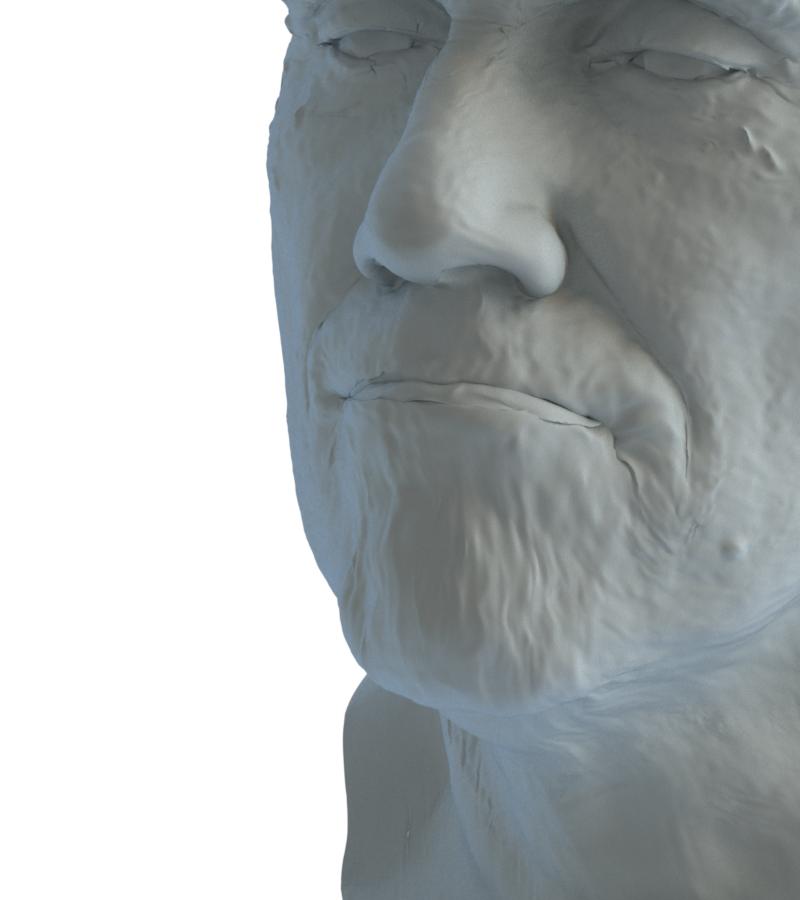} \\
\includegraphics[width=0.37\textwidth]{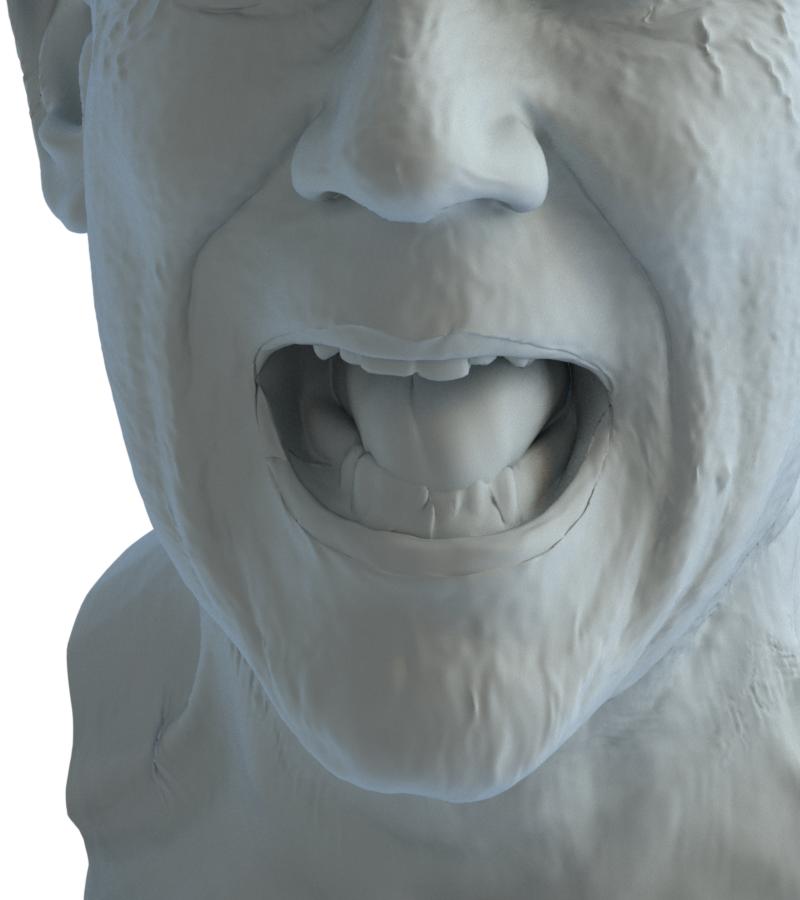} &
\includegraphics[width=0.37\textwidth]{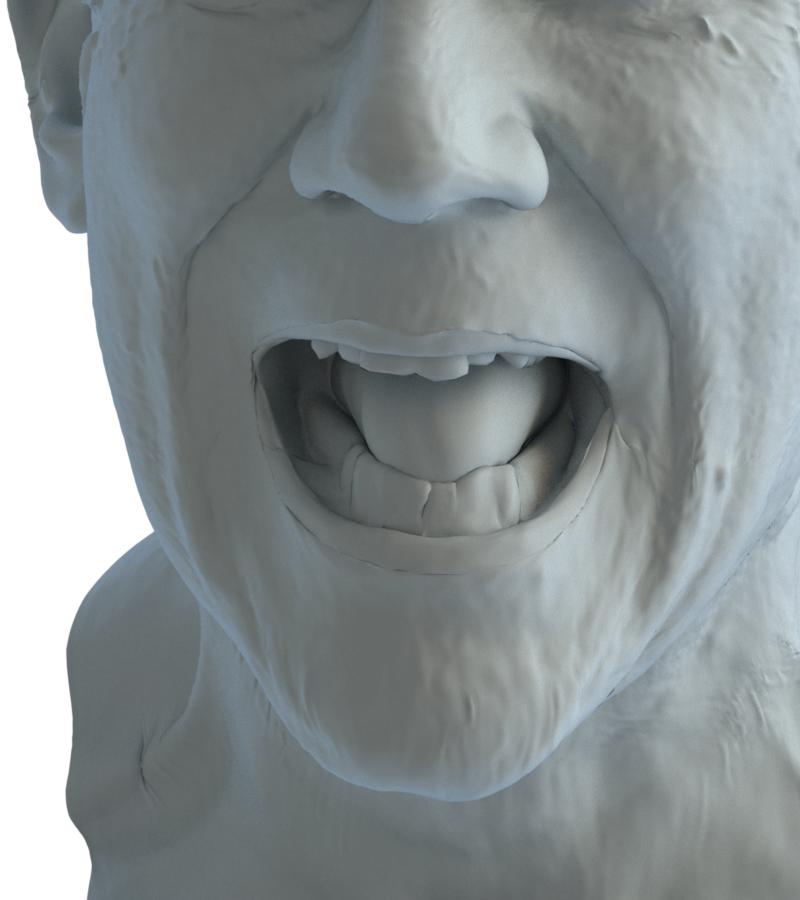} \\
\includegraphics[width=0.37\textwidth]{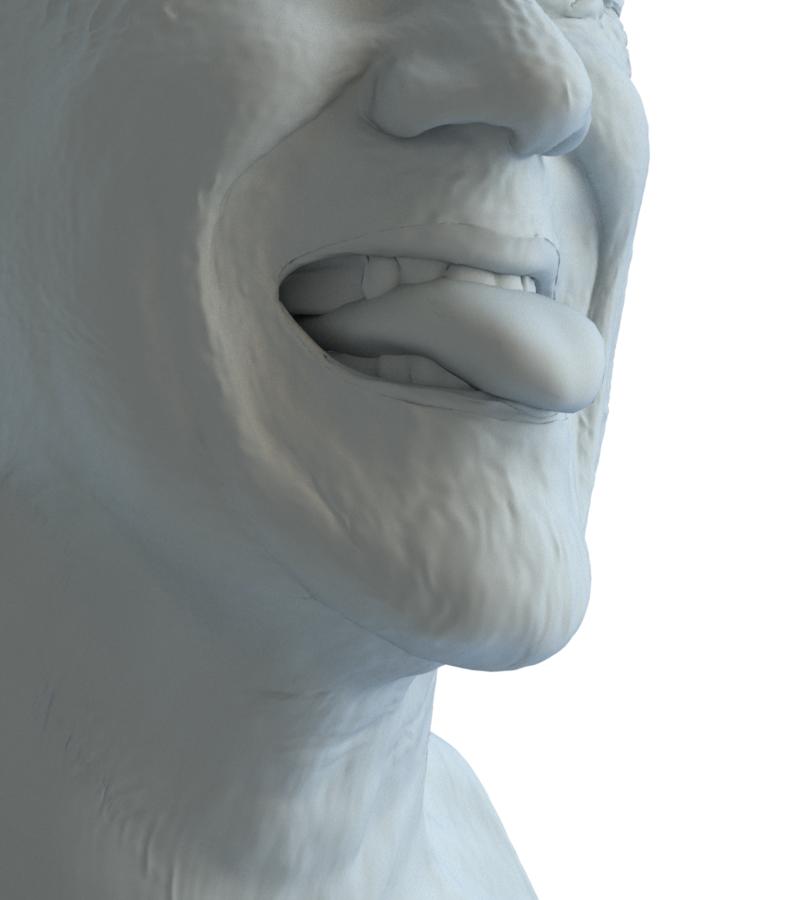} &
\includegraphics[width=0.37\textwidth]{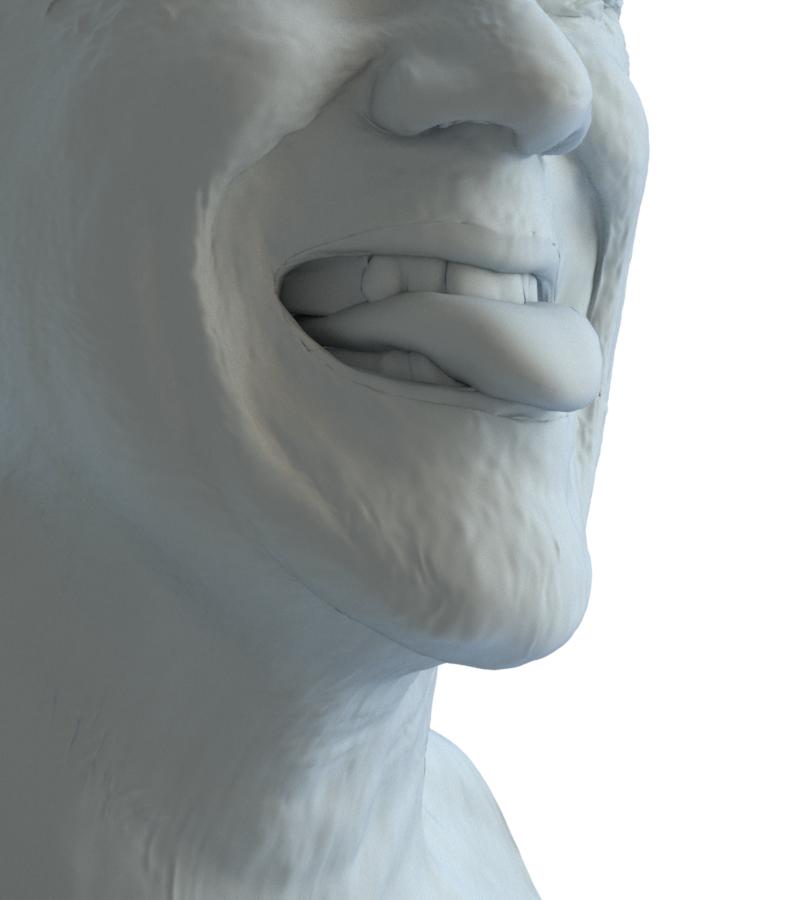}
\end{tabular}
\endgroup
\caption{\textbf{Dynamic head scene. Mesh evaluation}. Qualitative evaluation of reconstructed mesh for dynamic head scene. The first column is Nvdiffrast~\cite{laine2020modular} and the second column is our method EdgeGrad.}
    \label{f-ca3}
\end{figure*}

\begin{figure*}
\centering\footnotesize
\begingroup
\renewcommand{\arraystretch}{0.}
\setlength{\tabcolsep}{0pt}
\begin{tabular}{cc}
 Nvdiffrast~\cite{laine2020modular} & EdgeGrad (our) \\
\midrule
\includegraphics[width=0.37\textwidth]{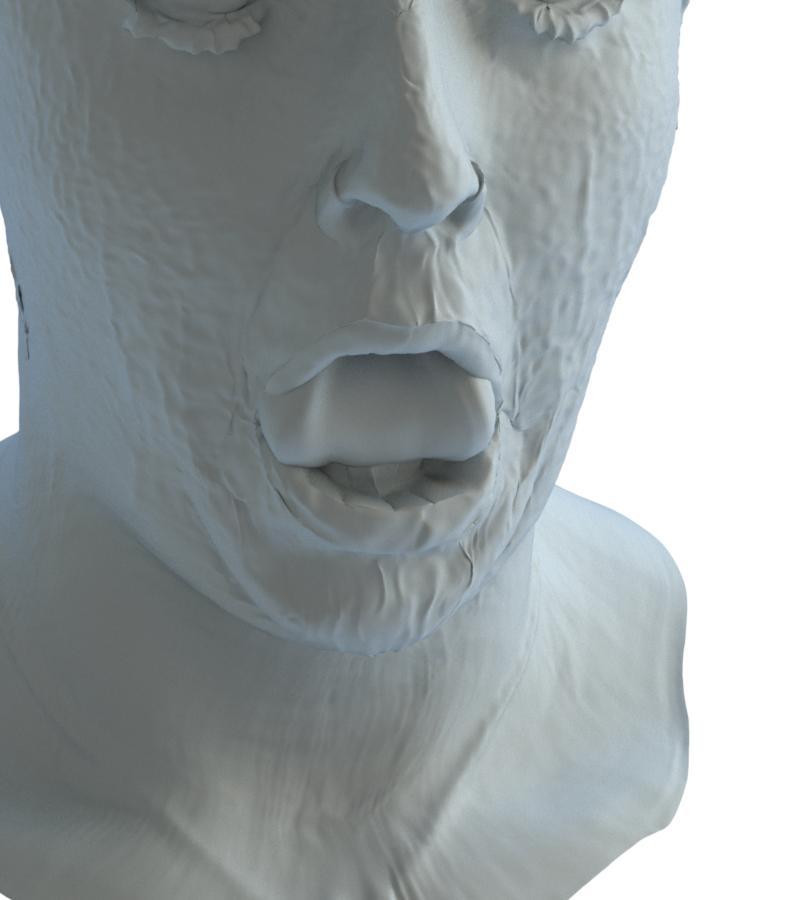} &
\includegraphics[width=0.37\textwidth]{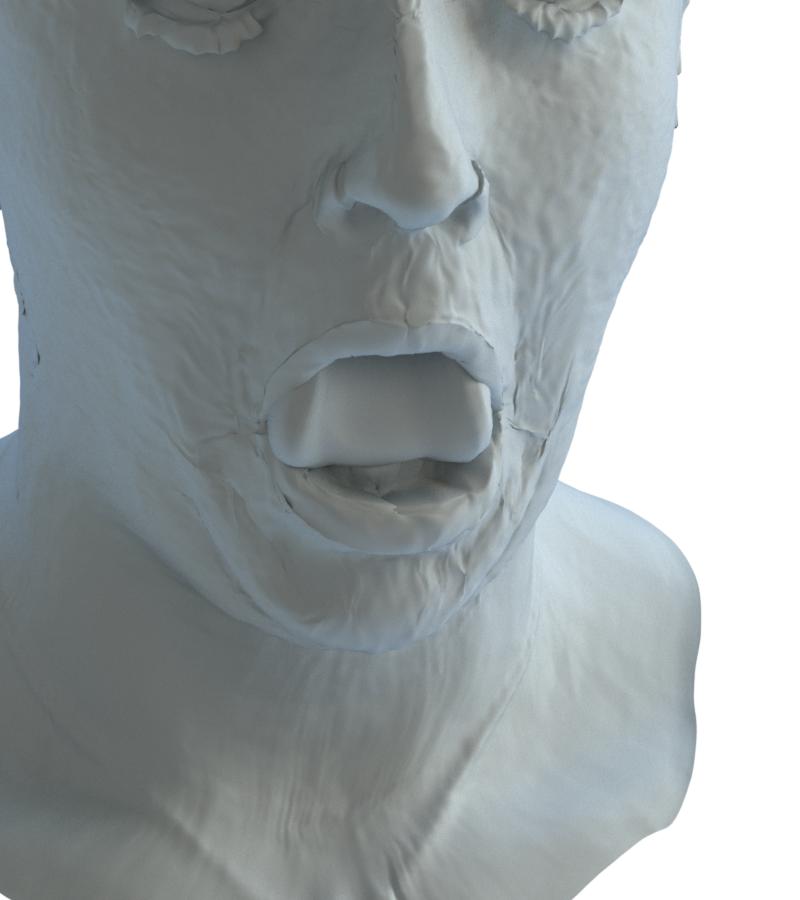} \\
\includegraphics[width=0.37\textwidth]{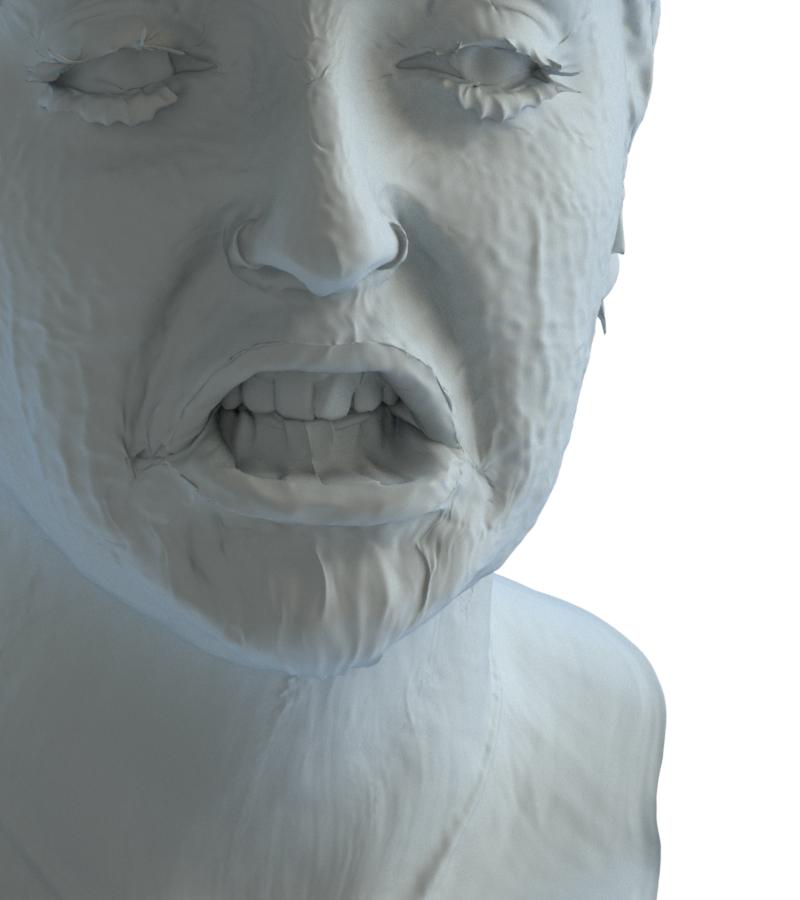} &
\includegraphics[width=0.37\textwidth]{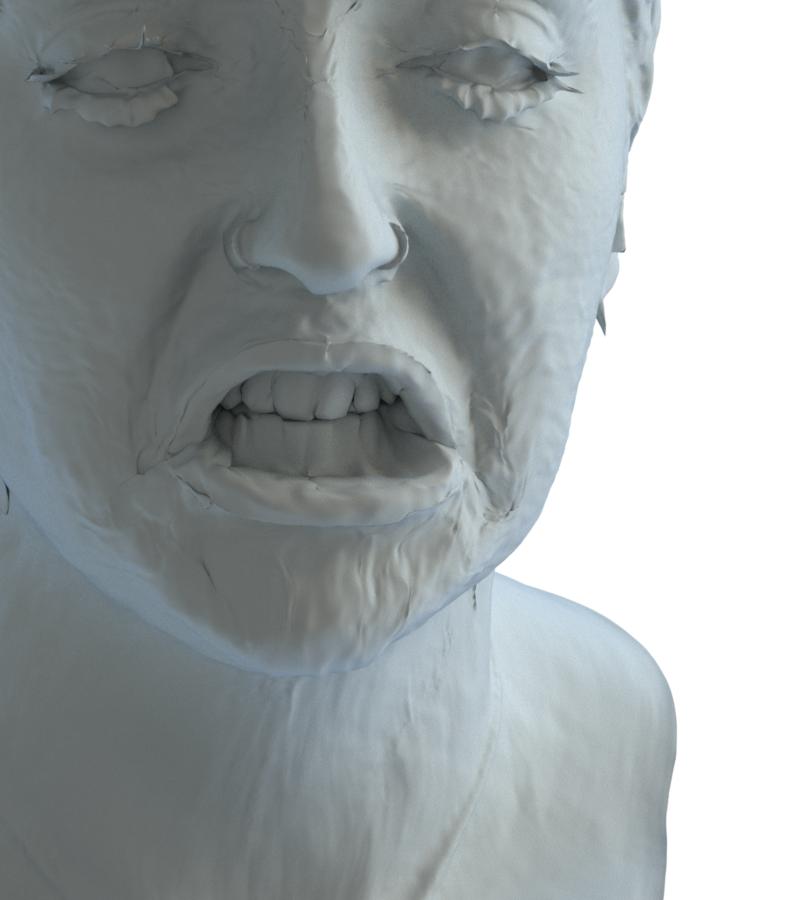} \\
\includegraphics[width=0.37\textwidth]{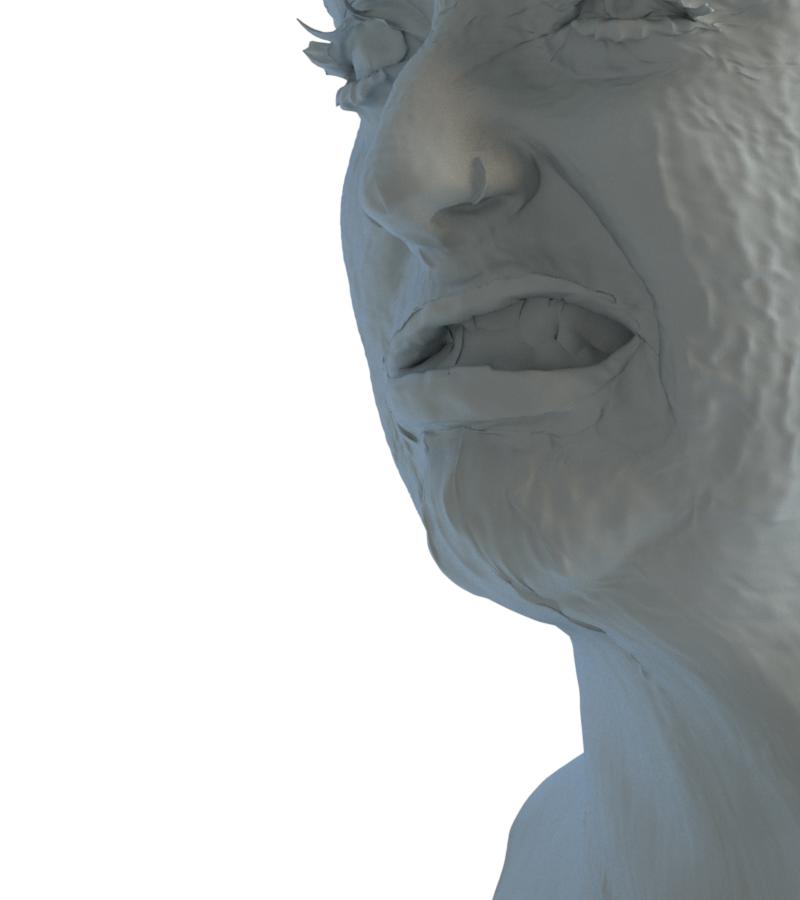} &
\includegraphics[width=0.37\textwidth]{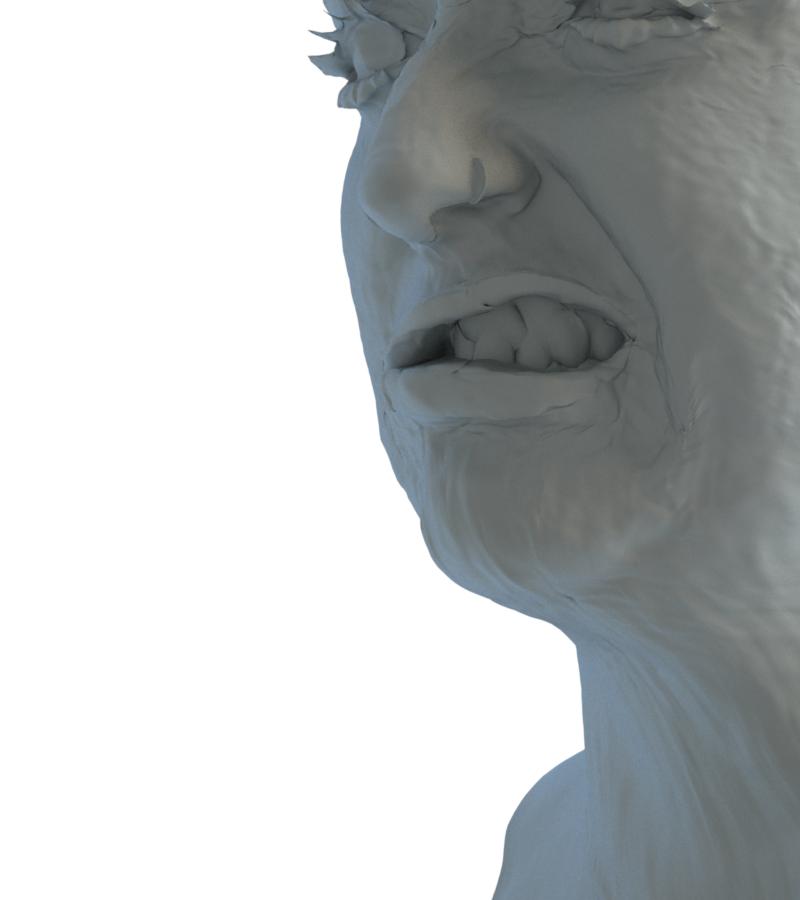} \\
\end{tabular}
\endgroup
\caption{\textbf{Dynamic head scene. Mesh evaluation. (Continuation of Fig.~\ref{f-ca3})}. Qualitative evaluation of reconstructed mesh for  dynamic head scene. The first column is Nvdiffrast~\cite{laine2020modular} and the second column is our method EdgeGrad.}
    \label{f-ca4}
\end{figure*}

Please refer to the Figures~\ref{f-ca1}--\ref{f-ca4} for qualitative evaluation of our method on fitting mesh-based avatars.

%
%
%
%

\end{document}